\documentclass{article} %
\usepackage{iclr2027_conference,times}

\usepackage{amsmath,amssymb,amsthm,mathtools}
\usepackage{bbm}              %
\usepackage{bm}
\usepackage{graphicx}

\usepackage{booktabs}
\usepackage{float}
\usepackage{placeins}
\usepackage{enumitem}
\usepackage{colortbl}  %
\definecolor{gold}{rgb}{1.0, 0.84, 0.0}
\definecolor{silver}{rgb}{0.75, 0.75, 0.75}
\definecolor{bronze}{rgb}{0.8, 0.5, 0.2}
\usepackage{rotating}  %
\usepackage{array}
\usepackage{algorithm}
\usepackage{algorithmic}

\usepackage{hyperref}
\usepackage{url}
\usepackage[capitalize,noabbrev]{cleveref}   %

\usepackage{xcolor}

\DeclareMathOperator*{\argmax}{arg\,max}
\DeclareMathOperator{\Var}{Var}
\DeclareMathOperator{\Cov}{Cov}
\DeclareMathOperator{\E}{\mathbb{E}}
\DeclareMathOperator{\diag}{diag}
\newcommand{\bigO}{\mathcal{O}}
\newcommand{\Ind}[1]{\mathbbm{1}[#1]}               %
\newcommand{\trans}{^{\top}}

\newcommand{\R}{\mathbb{R}}
\newcommand{\Rp}[1]{\mathbb{R}^{#1}}                 %
\newcommand{\Rmat}[2]{\mathbb{R}^{#1 \times #2}}     %
\newcommand{\bin}{\{0,1\}}
\newcommand{\binp}{\{0,1\}^{\np}}
\newcommand{\twop}{2^{\np}}

\newcommand{\np}{p}                                  %
\newcommand{\players}{P}                             %
\newcommand{\z}{\mathbf{z}}                          %
\newcommand{\zbar}{\bar{\z}}                         %
\newcommand{\zi}{\z^{(i)}}                           %
\newcommand{\Zall}{\mathbf{Z}}                       %
\newcommand{\cands}{\mathcal{C}}                     %

\newcommand{\vf}{\nu}                                %
\newcommand{\vfv}{\boldsymbol{\nu}}                  %
\newcommand{\sv}{\boldsymbol{\phi}}                  %
\newcommand{\A}{\mathbf{A}}                          %
\newcommand{\onevec}{\mathbf{1}}
\newcommand{\indvec}[1]{\onevec_{#1}}                %

\newcommand{\D}{\mathcal{D}}
\newcommand{\Dt}{\D_t}
\newcommand{\X}{\mathbf{X}}
\newcommand{\Xt}{\X_t}
\newcommand{\Y}{\mathbf{Y}}
\newcommand{\Yt}{\Y_t}
\newcommand{\budget}{T}
\newcommand{\Tinit}{T_0}

\newcommand{\GP}{\mathcal{GP}}
\newcommand{\N}{\mathcal{N}}
\newcommand{\hp}{\xi}                                %
\newcommand{\ls}{\ell}                               %
\newcommand{\oscale}{\sigma_{\vf}^{2}}               %
\newcommand{\noise}{\sigma_{\epsilon}^{2}}           %
\newcommand{\I}{\mathbf{I}}
\newcommand{\ez}{\mathbf{e}_{\z}}                    %
\newcommand{\kf}{k_{\hp}}                            %
\newcommand{\K}{\mathbf{K}_{\hp}}                    %
\newcommand{\KZZ}{\K(\Zall,\Zall)}
\newcommand{\KZX}{\K(\Zall,\Xt)}
\newcommand{\KXZ}{\K(\Xt,\Zall)}
\newcommand{\KXX}{\K(\Xt,\Xt)}
\newcommand{\KXXn}{\big[\KXX + \noise \I\big]}
\newcommand{\KXXninv}{\KXXn^{-1}}

\newcommand{\vfmid}{\vfv \mid \Dt}
\newcommand{\svmid}{\sv \mid \Dt}
\newcommand{\muvf}{\boldsymbol{\mu}_{\vfmid}}
\newcommand{\Sigvf}{\boldsymbol{\Sigma}_{\vfmid}}
\newcommand{\musv}{\boldsymbol{\mu}_{\svmid}}
\newcommand{\Sigsv}{\boldsymbol{\Sigma}_{\svmid}}
\newcommand{\ASig}{\A \Sigvf}
\newcommand{\ASigA}{\A \Sigvf \A\trans}

\newcommand{\EIG}{\mathrm{EIG}}
\newcommand{\EIGsv}{\EIG_{\sv}^{(t)}}
\newcommand{\EIGsvodd}{\EIG_{\sv,\odd}^{(t)}}         %
\newcommand{\MI}{I}                                  %
\newcommand{\Qz}{Q_t}                                %

\newcommand{\odd}{\mathrm{odd}}
\newcommand{\even}{\mathrm{even}}
\newcommand{\vfodd}{\vf_{\odd}}
\newcommand{\vfeven}{\vf_{\even}}
\newcommand{\ko}{k_{\hp,\odd}}                     %
\newcommand{\ke}{k_{\hp,\even}}                    %
\newcommand{\Ko}{\mathbf{K}_{\hp,\odd}}            %
\newcommand{\nodd}{t_{\odd}}                       %
\newcommand{\Dtodd}{\D_{t,\odd}}                   %
\newcommand{\reps}{\mathcal{R}_t}                    %
\newcommand{\Xtodd}{\X_{t,\odd}}                   %
\newcommand{\Ytodd}{\Y_{t,\odd}}                   %
\newcommand{\Yteven}{\Y_{t,\even}}                 %
\newcommand{\Dteven}{\D_{t,\even}}                 %
\newcommand{\Kot}{\widetilde{\mathbf{K}}_{\odd}}   %
\newcommand{\musvodd}{\boldsymbol{\mu}_{\sv \mid \Dtodd}}      %
\newcommand{\Sigsvodd}{\boldsymbol{\Sigma}_{\sv \mid \Dtodd}}   %

\newcommand{\shapleig}{\texttt{ShaplEIG}}
\newcommand{\methodname}{\texttt{ShapleyOG}}

\newcommand{\nq}{M}                                  %
\newcommand{\iv}{\tau}                               %
\newcommand{\ivb}{\tau'}                              %

\newcommand{\ncands}{c}                             %
\newcommand{\ncand}{c_{\odd}}                      %

\theoremstyle{plain}
\newtheorem{theorem}{Theorem}
\newtheorem{proposition}[theorem]{Proposition}
\newtheorem{lemma}[theorem]{Lemma}
\newtheorem{corollary}[theorem]{Corollary}
\AddToHook{env/proposition/begin}{\crefalias{theorem}{proposition}}
\AddToHook{env/lemma/begin}{\crefalias{theorem}{lemma}}
\AddToHook{env/corollary/begin}{\crefalias{theorem}{corollary}}

\title{ShapleyOG: Fast Adaptive Shapley Value Estimation with Odd Gaussian Processes}

\author{David Rundel$^{1,2}$, R. Teal Witter$^{3}$, Santo M. A. R. Thies$^{1,2}$, \\
\textbf{Matthias Feurer$^{4,5}$, Fabian Fumagalli$^{6,\dagger}$ \& Bernd Bischl$^{1,2,}$\thanks{Joint supervision.}} \\
$^{1}$LMU Munich \quad $^{2}$MCML \quad $^{3}$Claremont McKenna College \quad $^{4}$TU Dortmund University \\
$^{5}$Lamarr Institute for Machine Learning and Artificial Intelligence \quad $^{6}$Bielefeld University \\
\texttt{david.rundel@stat.uni-muenchen.de}
}

\newcommand{\runin}[1]{\par\textbf{#1}}

\iclrfinalcopy
\makeatletter
\AtBeginDocument{\let\iclr@maketitle@orig\@maketitle
  \def\@maketitle{\iclr@maketitle@orig\lhead{Preprint. Under review.}}}
\makeatother
\renewcommand{\thefootnote}{\fnsymbol{footnote}}
\begin{document}

\maketitle
\renewcommand{\thefootnote}{\arabic{footnote}}\setcounter{footnote}{0}

\begin{abstract}
   Shapley values are a central tool for attribution in machine learning. Since their exact computation requires evaluating the \emph{cooperative game} to be explained on exponentially many \emph{player coalitions}, they are typically estimated from a limited number of evaluations. Recently, \shapleig{} has shown that \emph{adaptively} selecting the coalitions to evaluate substantially improves sample efficiency over classical estimators. This, however, comes at the cost of a sequential procedure that, in every iteration, refits a Gaussian process surrogate of the game and selects the coalition with the largest expected information gain about the Shapley values.
   The resulting computational overhead limits its applicability to \emph{costly} games with low evaluation budgets.
   We present fast adaptive Shapley value estimation with Odd Gaussian Processes (\methodname{}) to overcome this limitation. \methodname{} leverages the fact that Shapley values depend only on the odd functional component of the game, and we show that under paired sampling a Gaussian process surrogate on this component alone yields the same Shapley value posterior and coalition selection criterion from half as many observations. Moreover, \methodname{} refits the surrogate only on a sparse schedule and efficiently updates the selection criterion in between, and it extends quadrature-based computation schemes to the prior covariance of the Shapley values, which allows for exact or cheaper approximate computation. %
   In extensive experiments on diverse games, \methodname{} matches the estimation accuracy of \shapleig{} while reducing its overhead by up to three orders of magnitude.
   As a result, it outperforms state-of-the-art non-adaptive estimators at equal total computation time already when a single evaluation takes a tenth of a second and a budget of a few thousand evaluations suffices, thereby widening the regime in which adaptive coalition selection is applicable and pays off.
\end{abstract}

\section{Introduction}
Shapley values (SVs; \citealp{S53}), a principled concept from cooperative game theory for fairly distributing the jointly achieved \emph{payoff} of a \emph{game} among its \emph{players}, have become a central tool for attribution in machine learning (ML). SVs are used to attribute predictions or predictive performance to features, training data points, or hyperparameters~\citep{LL17,GZ19,WMFL26}, often in critical domains such as healthcare, law, or finance~\citep{LEplus20,SHSL21}.
Computing the SVs exactly requires evaluating the game's \emph{value function} on all $\twop$ \emph{coalitions} of the $\np$ players, and each evaluation may itself incur a non-negligible cost. Exact computation is therefore only feasible for small games with cheap value functions, and SVs are typically estimated~\citep{CCLL23}. Accurate estimates, however, are essential, in particular in the critical domains mentioned above, and numerous estimators have been proposed, the vast majority of which follow the same two-stage principle: 1) evaluate the value function on a subset of coalitions and 2) estimate the SVs from these evaluations, either via Monte Carlo methods~\citep{CGT09,KBMH24} or via a surrogate model of the value function~\citep{LL17,FWM26}. While both stages have seen substantial progress in recent years, the choice of coalitions is, with few exceptions~\citep{SHSL21,BC21,RFMBF26}, \emph{non-adaptive}: 
it follows a scheme that is fixed a priori and does not adapt to the observed values. Moreover, it is \emph{symmetric} in the players, so that coalitions are distinguished at most by their size and never by the specific players involved~\citep{CL21,MW25}.

Recently, \shapleig{} \citep{RFMBF26} demonstrated that selecting coalitions \emph{adaptively}, i.e., based on previous evaluations, improves the sample efficiency of SV estimation substantially. \shapleig{} is a Bayesian experimental design (BED; \citealp{CV95,RFIB24}) approach: it models the value function with a Gaussian process (GP; \citealp{WR95,RW06}) surrogate and, in each iteration, evaluates the candidate coalition with the largest expected information gain (EIG; \citealp{L56}) about the SVs, i.e., the coalition expected to reduce the posterior uncertainty about the SVs the most. 
This adaptivity comes at a price: in every iteration, \shapleig{} refits the GP hyperparameters and evaluates the EIG over a candidate set, an overhead that grows disproportionately with the number of players and the evaluation budget. As a result, \shapleig{} is only applicable to settings with \emph{costly} value functions, in which the number of evaluations is severely limited in the first place.

Two recent developments point towards a substantially faster adaptive approach. First, SVs depend only on the \emph{odd} component of the value function, which OddSHAP~\citep{FBKRW26} exploits in polynomial regression by fitting only odd basis functions to odd observations, affording a richer surrogate while discarding no information about the SVs. Second, the integral representation of the Shapley weights~\citep{O72} enables the exact computation of SVs for product-structured games, such as those arising from the leaves of tree ensembles, via quadrature~\citep{LYL26b,WMY26,MRSMC26}, which, unlike the coefficient-based schemes employed in \shapleig{}, is embarrassingly parallel and can trade accuracy for cost by using fewer nodes. In this work, we combine both ideas into fast adaptive Shapley value estimation with Odd Gaussian Processes (\methodname{}), which retains the sample efficiency of \shapleig{} while reducing its computational overhead by a factor of up to $1634$ in our experiments, and thus substantially extends the applicability of adaptive estimation to less costly value functions, larger evaluation budgets, and games with more players. Our contributions are as follows:
\begin{itemize}[leftmargin=*,itemsep=1pt,topsep=2pt,parsep=0pt]
\item We exploit that the Hamming kernel GP decomposes into independent even and odd components~\citep{KGNN21,GRSDL16} and show that, under paired sampling~\citep{FBKRW26,CL21}, the SV posterior of the resulting \emph{odd GP} coincides with that of the full GP, and that under \emph{paired acquisition} the EIG about the SVs equals the EIG of the odd target under the odd GP. Both are thus obtained from training and candidate sets of half the size.
\item We extend quadrature-based computation schemes to the terms required for SV posterior inference and coalition selection; in particular, we express the prior covariance of the SVs as a bivariate integral, which Gauss--Legendre quadrature~\citep{GW69} evaluates exactly and embarrassingly parallel, or approximately with fewer nodes, reducing the cost of this typically dominant term of \shapleig{} from quartic to quadratic in the number of players.
\item We refit the GP hyperparameters only on a sparse schedule and, in between, efficiently update the EIG scores instead of recomputing them.
\item On a diverse benchmark, \methodname{} attains, on average, a lower estimation error than state-of-the-art non-adaptive estimators~\citep{MW25,WLM25,FBKRW26} at equal total computation time, i.e., including value function evaluations and estimation overhead, already when an evaluation takes a tenth of a second, e.g., one forward pass of a model, and a budget of several thousand evaluations suffices.

\end{itemize}

\section{Background} %

\subsection{Shapley Values} \label{sec:sv}
A cooperative game on a set $\players := \{1,\dots,\np\}$ of $\np$ players is defined by a value function $\vf: 2^{\players} \to \R$ that assigns a real-valued payoff to each coalition $S \subseteq \players$. 
SVs aggregate the value function into a single \emph{attribution} score for each player $i \in \players$,
\begin{align} \label{eq:shapley_set_def}
    \phi_i(\vf)
    &:=
    \sum_{S \subseteq \players \setminus \{ i\}}
    \frac{1}{\np \cdot \binom{\np-1}{\vert S \vert}}
    \Big( \vf(S \cup \{i\}) - \vf(S) \Big)
     = \A_{i} \, \vfv,
\end{align}
i.e., a weighted average of the marginal contributions $\vf(S \cup \{i\}) - \vf(S)$ of player $i$ over all coalitions $S$ not containing $i$, in which every coalition size receives the same total weight. Here, $\vfv := (\vf(S))_{S \subseteq \players} \in \Rp{\twop}$ collects the values of all coalitions, and $\A \in \Rmat{\np}{\twop}$ with rows $\A_i$ is the linear \emph{Shapley transformation}, so that $\sv(\vf) := (\phi_1(\vf), \dots, \phi_{\np}(\vf))\trans = \A \vfv$. Explicitly, $\A_{i,S} = 1 / \big(\np \binom{\np-1}{|S|-1}\big)$ if $i \in S$ and $\A_{i,S} = -1 / \big(\np \binom{\np-1}{|S|}\big)$ otherwise. 
By slight abuse of notation, we define $\vf$ equivalently on binary indicator vectors $\z \in \binp$ via their bijective correspondence with coalitions $S \subseteq \players$, denote the indicator vector of a coalition $S$ by $\indvec{S} \in \binp$, and write $\zbar := \onevec - \z$ for the complement coalition. We collect all $\twop$ coalitions as rows of a matrix $\Zall \in \bin^{\twop \times \np}$, index its rows and the entries of $\vfv = \vf(\Zall)$ by the coalitions themselves, and let $\ez \in \Rp{\twop}$ denote the standard basis vector of $\z$, so that $\vf(\z) = \ez\trans \vfv$. Finally, the $t$ coalition-value pairs observed up to iteration $t$ are denoted by $\Dt := \{(\zi, \vf(\zi))\}_{i=1}^{t}$, with the coalitions collected in a matrix $\Xt \in \bin^{t \times \np}$ and the values stacked in a vector $\Yt \in \Rp{t}$.

\subsection{Surrogate-Based Shapley Value Estimation} \label{sec:estimation}
Given an evaluation budget $\budget$, \emph{surrogate-based estimators} fit a surrogate $\hat{\vf}$ to the $\budget$ observed evaluations and return its SVs, $\hat{\sv} := \sv(\hat{\vf})$, which must be computable efficiently: this is immediate for the linear surrogates of Kernel SHAP~\citep{LL17} and Leverage SHAP~\citep{MW25}, whose SVs are their coefficients, tractable for polynomial surrogates~\citep{FWM26,FBKRW26}, and requires dedicated schemes for more expressive surrogates such as the tree-ensemble proxy of Regression MSR~\citep{WLM25} or the GP of \shapleig{}.
Which coalitions are evaluated is often tied to the estimator, e.g., \emph{leverage score sampling}~\citep{MW25} draws coalitions according to their leverage in the regression problem of Leverage SHAP; \emph{paired sampling}~\citep{CL21}, which evaluates every sampled coalition $\z$ together with its complement $\zbar$, and the \emph{border trick}~\citep{LL17,FMKHH23,MW25}, which favours very small and very large coalitions with the largest Shapley weights, are shared across estimators. All of these schemes are fixed before any value is observed and, although individual draws may depend on previous ones, symmetric in the players, i.e., invariant under relabeling them, so that coalitions are distinguished at most by their size.
\subsection{Adaptive Estimation with \shapleig{}} \label{sec:shapleig}
We now review \shapleig{}, the adaptive estimator that \methodname{} builds on.
\runin{Hamming Kernel GP Surrogate.}
\shapleig{} places a GP prior with a Hamming kernel~\citep{PBSWZ01,QWW08,H09} $\kf: \binp \times \binp \to \R_{>0}$ as covariance function on the value function. This product kernel quantifies the similarity of two coalitions as
\begin{align} \label{eq:hamming_kernel}
    \kf(\z, \z') = \oscale \prod_{j \in \players} \big( \alpha_j \Ind{z_j = z'_j} + \beta_j \Ind{z_j \neq z'_j} \big), \qquad \alpha_j := 1, \quad \beta_j := \exp(-1/\ls_j),
\end{align}
so that every player on whose membership the coalitions disagree reduces the covariance by its mismatch factor $\beta_j < 1$, whereas agreement leaves it unchanged (see \cref{app:gp}). The player-specific \emph{lengthscales} $\ls_j > 0$ and the output scale $\oscale$ form the hyperparameters $\hp$, which are learned from the observed data via maximum a posteriori (MAP) estimation~\citep{RW06}. In contrast to the surrogates of \cref{sec:estimation}, the GP is a \emph{probabilistic} surrogate: updating the prior with the data $\Dt$ yields the full posterior over the value function vector, $\vfmid \sim \N_{\twop}(\muvf, \Sigvf)$, whose mean $\muvf$ is used for SV estimation and whose covariance quantifies the uncertainty. Notably, since the SVs are a linear transformation of $\vfv$ (\cref{eq:shapley_set_def}), their posterior is available in closed form as a multivariate Gaussian as well~\citep{CMS23},
\begin{align} \label{eq:sv_posterior}
    \svmid \sim \N_{\np}(\A \muvf, \A \Sigvf \A\trans) = \N_{\np}(\musv, \Sigsv).
\end{align}
Denoting by $\K(\cdot,\cdot)$ the kernel matrix between two sets of coalitions, e.g., $\KZX \in \Rmat{\twop}{t}$ with entries $\kf(\z,\z')$ for all coalitions $\z$ in $\Zall$ and $\z'$ in $\Xt$, the posterior mean and covariance for a zero-mean prior are given by
\begin{align}
    \musv &= \A \KZX \KXXninv \Yt, \label{eq:sv_posterior_mean} \\
    \Sigsv &= \A \KZZ \A\trans - \A \KZX \KXXninv \KXZ \A\trans. \label{eq:sv_posterior_cov}
\end{align}
The posterior mean serves as the SV estimate, $\hat{\sv} := \musv = \sv(\muvf)$, which is consistent for a noiseless GP~\citep{RFMBF26}. Its computation involves the prior cross-covariance $\Cov(\sv, \vf(\Xt)) = \A \KZX \in \Rmat{\np}{t}$, whose $i$-th column holds the SVs of the \emph{kernel game} $\kf(\cdot, \z^{(i)})$ (\cref{app:quad_product}). Instead of enumerating all $\twop$ coalitions, it can be computed for the Hamming kernel in $\bigO(\np^2)$ per column via elementary symmetric polynomials (ESPs; \citealp{VH88,M98,C18}), as shown by \citet{MCM25}. Similarly, the prior covariance of the SVs, $\Cov(\sv, \sv) = \A \KZZ \A\trans$ in \cref{eq:sv_posterior_cov}, can be computed in $\bigO(\np^4)$~\citep{RFMBF26}. Together with the $\bigO(t^3)$ Cholesky decomposition of the training kernel matrix $\KXX \in \Rmat{t}{t}$ and the triangular solves, the SV posterior costs $\bigO(t^3 + t^2 \np + \np^4)$ for $t \geq \np$ (\cref{app:shapleig}). We refer to $\A \KZX$ and $\A \KZZ \A\trans$ as the \emph{SV prior covariances} in the following. In \cref{eq:sv_posterior_mean,eq:sv_posterior_cov}, $\noise$ is a small constant added for numerical stability, equivalent to observation noise (\cref{app:gp}).
\runin{EIG-based Coalition Selection.}
Given an initial design of coalition-value pairs, \shapleig{} at each iteration $t$ selects the next coalition $\z^{(t+1)}$ to evaluate by maximizing the EIG about the SVs over a candidate set $\cands \subseteq \binp$ of size $\ncands := |\cands|$.
This is the expected reduction in the posterior entropy $H$ of the SVs due to evaluating the candidate $\z$, or equivalently the mutual information (MI) between the SVs and the candidate value~\citep{L56,CT06}. Under the GP surrogate, it admits a closed form \citep{RFMBF26}:
\begin{align}
    \EIGsv(\z)
    &:= H\big(\sv \mid \Dt\big) - \E_{p(\vf(\z) \mid \Dt)} \Big[ H\big(\sv \mid \vf(\z), \Dt\big) \Big]
    = \MI\big(\sv; \vf(\z) \mid \Dt\big) \label{eq:eig_def} \\
    &\propto \log \Var(\vf(\z) \mid \Dt) - \log \big( \Var(\vf(\z) \mid \Dt) - \Qz(\z) \big), \label{eq:eig_closed_form}
\end{align}
where $\Var(\vf(\z) \mid \Dt)$ is the posterior predictive variance of $\vf(\z)$ and $\Qz(\z)$ is defined as
\begin{align} \label{eq:Q_def}
    \Qz(\z)
    &:= \big( \ASig \ez \big)\trans \big( \ASigA \big)^{-1} \big( \ASig \ez \big).
\end{align}
Evaluating the EIG over a candidate set $\cands$ additionally requires $\A \K(\Zall, \cands)$ in $\bigO(\ncands \np^2)$ and triangular solves for all candidates in $\bigO(\ncands t^2)$, i.e., $\bigO(t^3 + \np^4 + \ncands t^2)$ in total for $t \geq \np$ (\citealp{RFMBF26}; \cref{app:shapleig}).
The EIG depends on the observed values only through the hyperparameters $\hp$. To remain adaptive to previous evaluations, \shapleig{} therefore refits $\hp$ in every iteration, which requires repeated $\bigO(t^3)$ decompositions of the training kernel matrix and solutions of the associated triangular systems. Together with the EIG evaluation over potentially large candidate sets, this overhead accumulates over the iterations and, being polynomial in $t$ and $\np$, grows disproportionately for large evaluation budgets and numbers of players, so that the estimation can take hours or even days \citep{RFMBF26}.

\section{\methodname{}} \label{sec:method}
We transfer two recent developments to the Hamming kernel GP surrogate of \shapleig{}, the restriction to the odd component (\cref{sec:oddgp}) and quadrature-based computation for product games (\cref{sec:quad}), and combine them with a sparse refit schedule and efficient EIG updates into the \methodname{} algorithm (\cref{sec:algorithm}).
\subsection{Restricting the Surrogate to the Odd Component} \label{sec:oddgp}

Our starting point is that every (value) function decomposes uniquely into an \emph{odd} and an \emph{even} component, $\vf = \vfodd + \vfeven$, where for all $\z \in \binp$
\begin{align} \label{eq:odd_even}
    \vfodd(\z) := \tfrac{1}{2}\big(\vf(\z) - \vf(\zbar)\big) = -\vfodd(\zbar)
\end{align}
and $\vfeven := \vf - \vfodd$ satisfies $\vfeven(\z) = \vfeven(\zbar)$. Moreover, the SVs depend on the odd component only, i.e., $\sv(\vf) = \sv(\vfodd)$, since the SVs of any even function vanish \citep{FBKRW26}. Surrogate-based estimators can exploit this by isolating the odd component and thereby avoid spending resources on the even component, which is irrelevant for the SVs. This requires a surrogate $\hat{\vf}_{\odd}$ that is odd by construction, i.e., $\hat{\vf}_{\odd}(\z) = -\hat{\vf}_{\odd}(\zbar)$ for all $\z \in \binp$, and fitting it to the odd component only. The latter is achieved by paired sampling, since a pair $(\vf(\z), \vf(\zbar))$ determines the odd target $\vfodd(\z)$ exactly, whereas the former is less straightforward.
OddSHAP \citep{FBKRW26} obtains an odd polynomial surrogate by restricting the Fourier basis to odd basis functions, which halves both the candidate basis and, since only one representative per pair is fitted, the design matrix. For a fixed budget, this affords a more expressive surrogate and better SV estimates.

\runin{Posterior Inference.}
The Hamming kernel GP surrogate of \cref{sec:shapleig} admits the same restriction. The Hamming kernel is \emph{complement-invariant}, i.e., argumentwise invariant in the sense of \citet{GBRC12} under complementation, $\kf(\zbar, \zbar') = \kf(\z, \z')$ for all $\z, \z' \in \binp$ (\cref{app:complement}). The Hamming GP prior thus decomposes into independent even and odd components.
\begin{lemma}[Even-odd decomposition of the Hamming kernel GP] \label{lem:decomp}
Let $\vf \sim \GP(m, \kf)$ be the Hamming kernel GP with constant mean $m \in \R$. Then $\vf = \vfodd + \vfeven$ with independent GPs $\vfodd$ and $\vfeven$, where the odd component follows the \emph{odd GP}
\begin{align} \label{eq:odd_kernel}
    \vfodd \sim \GP(0, \ko), \qquad \ko(\z, \z') := \tfrac{1}{2} \big( \kf(\z, \z') - \kf(\z, \zbar') \big).
\end{align}
\end{lemma}
The lemma is a special case of the decomposition of GPs by (anti-)symmetrization~\citep{HB07,KGNN21} into independent components~\citep[Cor.~2(b)]{GRSDL16}; we include a direct proof in our notation in \cref{app:decomp_proof}. Crucially, the odd GP is precisely the odd surrogate required above: its posterior mean is odd for any training data~\citep{GRD16} (\cref{lem:odd_gp_odd} in \cref{app:decomp_proof}).
It remains to fit the odd GP to the odd component only. To this end, let the data $\Dt$ be \emph{paired}, i.e., contain the complement $\zbar$ of every observed coalition $\z$, so that it consists of $\nodd := t/2$ complement pairs. We collect one representative of each pair together with its odd target in the \emph{odd dataset} $\Dtodd := \{(\z, \vfodd(\z)) : \z \in \reps\}$, where $\reps \subset \{\zi\}_{i=1}^{t}$ contains exactly one coalition of each pair, with $\Xtodd \in \bin^{\nodd \times \np}$ and $\Ytodd \in \Rp{\nodd}$ defined analogously to $\Xt$ and $\Yt$.
\begin{theorem}[SV posterior equivalence under paired sampling] \label{thm:odd_posterior}
Let $\vf \sim \GP(m, \kf)$ be the Hamming kernel GP with fixed hyperparameters $\hp$ and constant mean $m \in \R$, and let $\Dt$ be paired with odd dataset $\Dtodd$. Then $p(\sv \mid \Dt) = p(\sv \mid \Dtodd)$, where the left-hand side denotes the SV posterior of the full GP $\vf$ conditioned on $\Dt$, and the right-hand side the SV posterior of the odd GP $\vfodd \sim \GP(0, \ko)$ from \cref{eq:odd_kernel} conditioned on the odd dataset $\Dtodd$.
\end{theorem}
The proof is given in \cref{app:posterior_proof}, where we also extend the statement to i.i.d.\ Gaussian observation noise. Thus, for given hyperparameters, both the posterior mean and the posterior covariance of the SVs can be obtained from the odd GP conditioned on half as many observations.
\runin{Adaptive Coalition Selection.}
Given the odd GP, the EIG-based coalition selection of \shapleig{} (\cref{sec:shapleig}) naturally becomes \emph{paired acquisition}: since only pairs yield odd targets, in each iteration a pair $\{\z, \zbar\}$ is selected and evaluated instead of a single coalition, and the natural criterion is the EIG of the odd target under the odd GP.
\begin{theorem}[EIG equivalence under paired acquisition] \label{thm:odd_eig}
Under the assumptions of \cref{thm:odd_posterior}, for every candidate coalition $\z \in \binp$, $\MI\big(\sv; \vf(\z), \vf(\zbar) \mid \Dt\big) = \MI\big(\sv; \vfodd(\z) \mid \Dtodd\big) =: \EIGsvodd(\z)$, i.e., the joint EIG about the SVs of evaluating the pair $\{\z, \zbar\}$ under the full GP equals the EIG of the odd target $\vfodd(\z)$ under the odd GP conditioned on the odd dataset.
\end{theorem}
The proof is given in \cref{app:odd_eig}, again including observation noise. Hence, under paired sampling, coalition selection can be performed entirely on the odd GP as well. Moreover, the odd EIG is invariant under complements, $\EIGsvodd(\z) = \EIGsvodd(\zbar)$ (\cref{app:odd_eig}), so that one representative per candidate pair suffices.
\runin{Computational Consequences.}
By \cref{thm:odd_posterior,thm:odd_eig}, the SV posterior and the EIG under the odd GP take the same form as \cref{eq:sv_posterior_mean,eq:sv_posterior_cov,eq:eig_closed_form,eq:Q_def}, but with the odd dataset $\Dtodd$ of half as many rows in place of $\Dt$: the training kernel matrix is four times smaller and its Cholesky decomposition up to $8$ times faster. Moreover, the EIG has to be scored for only one representative per candidate pair, which halves the number of scored candidates to $\ncand := \ncands/2$, and paired acquisition halves the number of iterations for a given budget. Finally, the prior covariances involving the Shapley transformation coincide with those of the full kernel (\cref{prop:odd_identities} in \cref{app:identities}), so they need not be computed for the odd kernel, which would double their cost. See \cref{app:odd_consequences} for details.

\subsection{Quadrature-Based Computation of the SV Prior Covariances} \label{sec:quad}

The SV posterior (\cref{eq:sv_posterior_mean,eq:sv_posterior_cov}) and the EIG (\cref{eq:eig_closed_form,eq:Q_def}) require the SV prior covariances $\A \K(\Zall, \z)$ and $\A \K(\Zall, \Zall) \A\trans$. By \cref{prop:odd_identities}, the same terms based on the full kernel serve the odd GP. We now show that both terms can be computed via quadrature.
To this end, we first recall \emph{product games}~\citep{MRSMC26}, which arise, for instance, from the leaves of tree ensembles, and whose value function factorizes as $\vf(S) = \prod_{j \in S} u_j$ with player-specific factors $u_j \in \R$ (and $\vf(\emptyset) = 1$). Moreover, the Shapley weights in \cref{eq:shapley_set_def} admit a classical integral representation~\citep{AAR99,O72}, $1 / \big( \np \binom{\np-1}{s} \big) = \int_0^1 \iv^{s} (1-\iv)^{\np-1-s} \, d\iv$ for $s \in \{0, \dots, \np-1\}$ (\cref{lem:beta} in \cref{app:quad_beta}).
Plugging this representation into the definition of the SVs in \cref{eq:shapley_set_def} for a product game collapses its SVs into one-dimensional integrals over polynomials of degree $\np - 1$, without enumerating the exponentially many coalitions~\citep{LYL26b,WMY26,MRSMC26}.

\runin{Linear Term.}
We now show that the Hamming kernel falls into this class: for a fixed coalition $\z \in \binp$, the kernel game $\vf_{\z}(S) := \kf(\indvec{S}, \z)$ is a product game up to the multiplicative constant $\kf(\boldsymbol{0}, \z)$ (see \cref{app:quad_product}), i.e., $\vf_{\z}(S) = \kf(\boldsymbol{0}, \z) \prod_{j \in S} \beta_j^{1-2z_j}$. Consequently, the $i$-th entry of $\A \K(\Zall, \z)$ is the SV of player $i$ in $\vf_{\z}$, and the product game integral becomes
\begin{align} \label{eq:quad_linear}
    \big( \A \K(\Zall, \z) \big)_i = \phi_i(\vf_{\z}) = \kf(\boldsymbol{0}, \z) \, \big( \beta_i^{1-2z_i} - 1 \big) \int_0^1 \prod_{j \neq i} \big( (1-\iv) + \iv \, \beta_j^{1-2z_j} \big) \, d\iv.
\end{align}
This merely specializes the product game result to the Hamming kernel. In \cref{app:quad_linear}, we derive it directly from the factorized Hamming kernel and show that the result coincides with \cref{eq:quad_linear}.
\runin{Quadratic Term.}
The prior covariance of the SVs $\A \KZZ \A\trans$, in contrast, involves the Shapley weights twice: its $(i,j)$-th entry is the bilinear form induced by the kernel matrix, evaluated at the $i$-th and $j$-th row of $\A$, i.e., $(\A \KZZ \A\trans)_{i,j} = \A_i \KZZ \A_j\trans$. We show that this term similarly collapses into bivariate integrals over polynomials, again without enumerating coalitions.
\begin{theorem}[Quadratic term] \label{thm:quad_quadratic}
For $r \in \players$ and $\iv, \ivb \in [0,1]$, let $d_r := 1 - \beta_r$ and $h_r(\iv,\ivb) := 1 - d_r (\iv + \ivb - 2\iv\ivb)$. Then, for all players $i \neq j$,
\begin{align}
    \big( \A \K(\Zall, \Zall) \A\trans \big)_{i,j}
    &= \oscale \, d_i \, d_j \iint_{[0,1]^2} (1-2\iv)(1-2\ivb) \prod_{r \notin \{i,j\}} h_r(\iv,\ivb) \, d\iv \, d\ivb, \label{eq:quad_offdiag} \\
    \big( \A \K(\Zall, \Zall) \A\trans \big)_{i,i}
    &= 2 \, \oscale \, d_i \iint_{[0,1]^2} \prod_{r \neq i} h_r(\iv,\ivb) \, d\iv \, d\ivb. \label{eq:quad_diag}
\end{align}
\end{theorem}
This bivariate integral representation has, to our knowledge, not been considered before; the proof is given in \cref{app:quad_quadratic}.
\runin{Computational Cost.}
The integrands of \cref{eq:quad_linear,eq:quad_offdiag,eq:quad_diag} are polynomials of degree $\np - 1$ in each integration variable, amenable to Gauss--Legendre (GL) quadrature~\citep{GW69}.
\begin{corollary}[Cost and exactness of GL quadrature] \label{cor:quad_gl}
Let $(\tau_q, \omega_q)_{q=1}^{\nq}$ be the GL nodes and weights on $[0,1]$. Replacing the integral in \cref{eq:quad_linear} by $\sum_{q} \omega_q \, (\cdot)(\tau_q)$ and the integrals in \cref{eq:quad_offdiag,eq:quad_diag} by $\sum_{q, q'} \omega_q \omega_{q'} \, (\cdot)(\tau_q, \tau_{q'})$ costs $\bigO(\np \nq)$ for all $\np$ entries of $\A \K(\Zall, \z)$ per coalition $\z$ and $\bigO(\np^2 \nq^2)$ for $\A \K(\Zall, \Zall) \A\trans$, and is exact for $\nq \geq \lceil \np/2 \rceil$, i.e., at costs $\bigO(\np^2)$ and $\bigO(\np^4)$.
\end{corollary}
This follows from the fact that GL quadrature with $\nq$ nodes integrates polynomials of degree up to $2\nq - 1$ exactly (\cref{app:quad_gl}; \citealp{GW69}), and from an implementation based on leave-one-out products (\cref{app:quad_impl}). For the linear term, this computation scheme was proposed in prior work~\citep{LYL26b,WMY26,MRSMC26}. For exact quadrature, this route matches the asymptotic cost of the ESP schemes of \shapleig{} (\cref{app:quad_vs_esp}). However, it is inherently parallelizable and admits approximate computations with $\nq < \lceil \np/2 \rceil$ nodes, whereas the ESP recursions are sequential over the players and admit no natural approximation.
\subsection{The \methodname{} Algorithm} \label{sec:algorithm}
The preceding results culminate in \methodname{}, our algorithm for fast adaptive SV estimation, as detailed in \cref{alg:shapleyog}. In the following, we outline its main differences to \shapleig{}. \Cref{app:alg_comparison,app:alg_costs} explain them in more detail and present all involved quantities and their computational costs, also in comparison to \shapleig{}.
\begin{algorithm}[t]
\caption{\methodname{}}
\label{alg:shapleyog}
\begin{algorithmic}[1]
\REQUIRE Odd GP prior $\vfodd \sim \GP(0, \ko)$ with hyperparameter prior $p(\hp)$, paired initial design of size $\Tinit$, evaluation budget $\budget$, number of GL nodes $\nq$ (exact for $\nq \geq \lceil \np/2 \rceil$)
\STATE Initialize dataset: $\D_{\Tinit} := \{(\zi, \vf(\zi))\}_{i=1}^{\Tinit}$; \quad $t := \Tinit$
\WHILE{$t < \budget$}
    \IF{$t = \Tinit$ or $t$ is a power of two}
        \STATE Refit hyperparameters: $\hp := \argmax\nolimits_{\hp} \, p(\hp \mid \Dtodd)$ \hfill (\cref{sec:oddgp})
        \STATE (Re-)sample paired candidate set $\cands \subseteq \binp \setminus \{\zi\}_{i=1}^{t}$ \hfill (\cref{app:exp_shapleyog})
        \STATE Compute $\EIGsvodd(\z)$ per pair in $\cands$ \hfill (\cref{sec:oddgp,sec:quad})
    \ELSE
        \STATE Update $\EIGsvodd(\z)$ per pair in $\cands$ given new data \hfill (\cref{app:alg_updates})
    \ENDIF
    \STATE Select pair: $\z^{(t+1)} \in \smash[b]{\argmax\nolimits_{\z \in \cands}} \, \EIGsvodd(\z)$, \quad $\z^{(t+2)} := \onevec - \z^{(t+1)}$ \hfill (\cref{thm:odd_eig})
    \STATE Update dataset: $\D_{t+2} := \Dt \cup \{(\z^{(t+1)}, \vf(\z^{(t+1)})), (\z^{(t+2)}, \vf(\z^{(t+2)}))\}$
    \STATE Update candidate set: $\cands := \cands \setminus \{\z^{(t+1)}, \z^{(t+2)}\}$; \quad $t := t + 2$
\ENDWHILE
\STATE Refit $\hp$ on $\D_{\budget,\odd}$ and \textbf{return} SV posterior $\N_{\np}(\boldsymbol{\mu}_{\sv \mid \D_{\budget,\odd}}, \boldsymbol{\Sigma}_{\sv \mid \D_{\budget,\odd}})$ \hfill (\cref{thm:odd_posterior})
\end{algorithmic}
\end{algorithm}
\runin{Odd GP and GL Quadrature.}
\methodname{} computes the SV prior covariances via GL quadrature (\cref{sec:quad}) and employs the odd GP as surrogate (\cref{sec:oddgp}). It thus acquires coalitions in pairs, as required by the odd GP (\cref{thm:odd_posterior,thm:odd_eig}), and learns the hyperparameters $\hp$ under the odd GP by MAP estimation on the odd targets (learning $\hp$ under the odd GP is more appropriate, as only the odd component matters for the SVs). %
As a result, the matrices associated with the training data and the candidate set, and the computations on them, scale with $\nodd$ and $\ncand$ instead of $t$ and $\ncands$, and the SV prior covariances cost $\bigO(\np^2 \nq^2)$ instead of $\bigO(\np^4)$. An iteration with refit thus costs $\bigO(\nodd^3 + \nodd^2 \ncand + \np^2 \nq^2)$ compared to $\bigO(t^3 + t^2 \ncands + \np^4)$ for \shapleig{} (\cref{app:alg_costs}). In particular, the prior covariance of the SVs, whose cost quartic in $\np$ typically dominates that of \shapleig{}, is computed in an inherently parallelizable manner and can be approximated at a cost only quadratic in $\np$.
\runin{Refit Schedule and Incremental EIG Updates.}
\shapleig{} refits the hyperparameters in every iteration or, for larger games, on a staged schedule \citep{RFMBF26}. \methodname{} instead refits only when the number of observations $t$ reaches a power of two, which reduces the number of refits for a budget $\budget$ from $\bigO(\budget)$ to $\bigO(\log \budget)$. %
Between refits, the hyperparameters are fixed, which \methodname{} exploits systematically. Whereas \shapleig{} recomputed Cholesky factorizations and triangular solves in every iteration even when its hyperparameters were kept fixed, we show in \cref{app:alg_updates} that the posterior covariances entering the EIG need not be recomputed but can be updated exactly and in a numerically stable manner via a bordered Cholesky factorization, an extension of the half-solves, and rank-one updates~\citep{GV13}. This reduces the cost of an iteration without refit from $\bigO(t^3 + t^2 \ncands + \np^2 \ncands)$ for \shapleig{} to $\bigO(\nodd^2 + \nodd \ncand + \np^3 + \np^2 \ncand)$ (\cref{app:alg_costs}), so that the EIG updates no longer scale cubically in the number of observations. Moreover, since all EIG quantities are recomputed at a refit anyway, \methodname{} resamples a paired candidate set of not yet evaluated coalitions at every refit essentially for free, whereas \shapleig{} depletes a fixed set, which risks that only less informative candidates remain (\cref{app:exp_shapleyog}).

\section{Experiments} \label{sec:experiments}
\begin{figure}[t]
  \begin{minipage}[c]{0.47\textwidth}
    \centering
    \includegraphics[width=\linewidth]{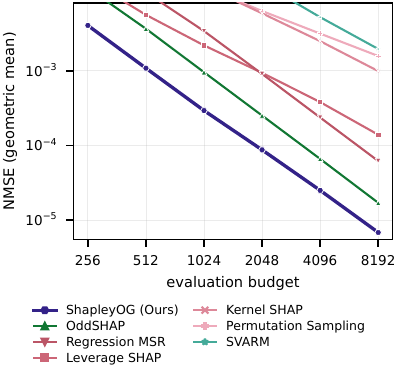}
  \end{minipage}\hfill
  \begin{minipage}[c]{0.51\textwidth}
    \centering \scriptsize \setlength{\tabcolsep}{2.5pt}
    \begin{tabular}{@{}lrrrrrr@{}}
\toprule
Game ($p$) $\backslash$ $T$ & 256 & 512 & 1024 & 2048 & 4096 & 8192 \\
\midrule
Language (IMDB) (14) & \cellcolor{gold!60}2.9 & \cellcolor{silver!60}2.6 & \cellcolor{silver!60}2.6 & \cellcolor{silver!60}2.0 & \cellcolor{gold!60}2.0 & \cellcolor{gold!60}1.7 \\
ResNet18-SP (14) & \cellcolor{gold!60}1.1 & \cellcolor{gold!60}1.0 & \cellcolor{gold!60}1.0 & \cellcolor{gold!60}1.0 & \cellcolor{gold!60}1.0 & \cellcolor{gold!60}1.0 \\
ViT4by4 (16) & \cellcolor{gold!60}2.0 & \cellcolor{gold!60}1.2 & \cellcolor{gold!60}1.2 & \cellcolor{gold!60}1.2 & \cellcolor{gold!60}1.4 & \cellcolor{gold!60}1.6 \\
DINOv2-16 (16) & \cellcolor{bronze!60}2.5 & \cellcolor{gold!60}2.1 & \cellcolor{gold!60}1.8 & \cellcolor{silver!60}1.8 & \cellcolor{silver!60}1.9 & \cellcolor{silver!60}2.3 \\
Forest Fires (13) & \cellcolor{silver!60}3.1 & \cellcolor{silver!60}3.1 & \cellcolor{bronze!60}3.3 & \cellcolor{silver!60}3.4 & \cellcolor{bronze!60}3.3 & -- \\
AdultCensus (14) & \cellcolor{gold!60}1.1 & \cellcolor{gold!60}1.2 & \cellcolor{gold!60}1.1 & \cellcolor{gold!60}1.4 & \cellcolor{silver!60}2.0 & \cellcolor{bronze!60}2.6 \\
Estate (15) & \cellcolor{gold!60}1.4 & \cellcolor{gold!60}1.1 & \cellcolor{gold!60}1.3 & \cellcolor{gold!60}1.6 & \cellcolor{gold!60}1.7 & \cellcolor{silver!60}2.1 \\
Crime (101) & \cellcolor{gold!60}1.0 & \cellcolor{gold!60}1.0 & \cellcolor{gold!60}1.0 & \cellcolor{gold!60}1.0 & \cellcolor{gold!60}1.0 & \cellcolor{gold!60}1.0 \\
Seismic Bumps (15) & \cellcolor{gold!60}1.0 & \cellcolor{gold!60}1.3 & \cellcolor{gold!60}1.1 & \cellcolor{gold!60}1.4 & \cellcolor{gold!60}1.5 & \cellcolor{silver!60}2.0 \\
Churn (19) & \cellcolor{gold!60}1.1 & \cellcolor{gold!60}1.0 & \cellcolor{gold!60}1.0 & \cellcolor{gold!60}1.0 & \cellcolor{gold!60}1.1 & \cellcolor{gold!60}1.3 \\
Credit-G (20) & \cellcolor{gold!60}1.4 & \cellcolor{silver!60}2.0 & \cellcolor{silver!60}2.2 & \cellcolor{gold!60}1.6 & \cellcolor{silver!60}1.8 & \cellcolor{silver!60}2.1 \\
Bankruptcy (64) & \cellcolor{gold!60}1.0 & \cellcolor{gold!60}1.0 & \cellcolor{gold!60}1.0 & \cellcolor{gold!60}1.0 & \cellcolor{gold!60}1.0 & \cellcolor{gold!60}1.0 \\
Superconductivity (81) & \cellcolor{gold!60}1.0 & \cellcolor{gold!60}1.0 & \cellcolor{gold!60}1.0 & \cellcolor{gold!60}1.0 & \cellcolor{gold!60}1.0 & \cellcolor{gold!60}1.0 \\
NATICUSdroid (86) & \cellcolor{gold!60}1.0 & \cellcolor{gold!60}1.0 & \cellcolor{gold!60}1.0 & \cellcolor{gold!60}1.0 & \cellcolor{gold!60}1.0 & \cellcolor{gold!60}1.0 \\
MIC (111) & \cellcolor{gold!60}1.0 & \cellcolor{gold!60}1.0 & \cellcolor{gold!60}1.0 & \cellcolor{gold!60}1.0 & \cellcolor{gold!60}1.0 & \cellcolor{gold!60}1.0 \\
\midrule
\emph{pooled} & \cellcolor{gold!60}1.4 & \cellcolor{gold!60}1.3 & \cellcolor{gold!60}1.3 & \cellcolor{gold!60}1.3 & \cellcolor{gold!60}1.4 & \cellcolor{gold!60}1.5 \\
\bottomrule
\end{tabular}

  \end{minipage}
  \caption{Left: NMSE vs.\ evaluation budget, geometric mean over $30$ seeds and the $14$ games whose budget grid covers $256$ to $8192$. Right: average NMSE rank of \methodname{} among all estimators over $30$ seeds per game (players in parentheses) and budget $\budget$ (-- beyond a game's grid; last row: $14$ games pooled); gold/silver/bronze: lowest/second/third lowest average rank of all estimators.}
  \label{fig:main_results}
\end{figure}

In this section, we evaluate \methodname{} against state-of-the-art non-adaptive SV estimators in terms of estimation accuracy and total computation time, and we analyze its individual measures along an ablation path from \shapleig{}. We first describe the experimental setup and then present the results.
\subsection{Experimental Setup} \label{sec:exp_setup}
We run \methodname{} as in \cref{alg:shapleyog} with the following configuration. The initial design is the \emph{border trick} design~\citep{FMKHH23} of size $\Tinit = 2\np + 2$, consisting of the empty and the grand coalition as well as all singletons and their complements. At every refit, a paired candidate set of $10240$ coalitions is sampled from the not yet evaluated coalitions, and the quadrature uses the exact node count. We record the SV estimates at all budgets that are powers of two from the first above $\Tinit$ up to $8192$ (capped at $2^{\np}/2$, which only affects the $13$-player Forest Fires game at $4096$), where a budget counts all evaluations including the initial design; each game is run with $30$ seeds (details in \cref{app:exp_shapleyog}). Beyond this main experiment, we run an ablation path from \shapleig{} to \methodname{} and a quadrature study that compares exact and approximate GL quadrature with the ESP route of \shapleig{} (\cref{app:exp_ablations}).
\runin{Estimation Methods.}
We compare \methodname{} against widely used and recent non-adaptive SV estimators: Kernel SHAP~\citep{LL17}, Leverage SHAP~\citep{MW25}, permutation sampling~\citep{CGT09,MCFH22}, SVARM~\citep{KBMH24}, Regression MSR~\citep{WLM25}, and OddSHAP~\citep{FBKRW26}. All are run with the \texttt{shapiq}~\citep{MBFKHH24} defaults (\cref{app:exp_baselines}) and the same evaluation budgets; the ablation path additionally compares against \shapleig{}~\citep{RFMBF26} (\cref{app:exp_ablations}).
\runin{Games.}
We consider $15$ diverse explanation games with $13$ to $111$ players, i.e., $450$ game instances over the $30$ seeds. %
They comprise local explanations of vision and language models, of the tabular foundation model TabPFN-3~\citep{grinsztajn_tabpfn-3_2026}, and of LightGBM~\citep{KMplus17} gradient-boosted tree ensembles on tabular data, and are detailed in \cref{app:exp_games}.
\runin{Metrics.}
We measure the estimation accuracy by the normalized mean squared error (NMSE) of the SV estimate, i.e., the squared error relative to the squared norm of the exact SVs. The overhead of an estimator is its wall-clock time excluding the value function evaluations, which for \methodname{} comprises the repeated hyperparameter fitting, EIG maximization, and posterior computation, and the total computation time is the overhead plus the number of evaluations times an assumed cost per evaluation (\cref{app:exp_metrics}). Details on the compute resources and on reproducibility are given in \cref{app:exp_compute}.
\subsection{Results} \label{sec:exp_results}
\runin{Estimation Accuracy.}
\Cref{fig:main_results} (left) shows the NMSE aggregated by the geometric mean across the $30$ seeds and the $14$ games whose budget grid reaches $8192$ evaluations. \methodname{} attains the lowest NMSE at every budget, consistently below that of the strongest competitor, OddSHAP, and reaches the same accuracy with substantially fewer evaluations, up to about half as many. Per-game curves and arithmetic means in \cref{app:res_pergame} show the same picture. The average NMSE ranks (\cref{fig:main_results}, right; \cref{app:res_avgrank}) agree: pooled, \methodname{} ranks lowest at every budget ($1.3$--$1.5$ vs.\ $2.0$--$2.6$ for OddSHAP). Per game, it attains the lowest average rank for most game-budget pairs and, on the remaining ones, all on games with at most $20$ players, the second or third lowest.
\begin{figure}[t]
  \begin{minipage}[t]{0.49\textwidth}
    \vspace{0pt}\centering
    \includegraphics[width=\linewidth]{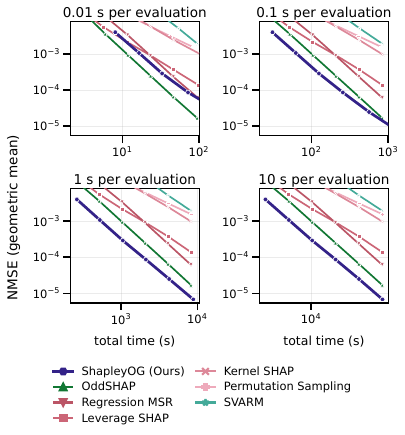}
  \end{minipage}\hfill
  \begin{minipage}[t]{0.49\textwidth}
    \vspace{0pt}\centering
    \includegraphics[width=\linewidth]{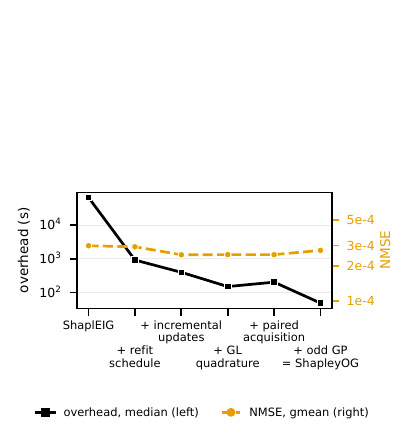}\par
    \vspace{-1.07407\linewidth}\vspace{0.05285\linewidth}%
    \parbox[c][0.28677\linewidth][c]{\linewidth}{\centering \scriptsize \setlength{\tabcolsep}{2.5pt}
      \begin{tabular}{@{}lrrrrr@{}}
\toprule
\multicolumn{6}{@{}l}{Mean overhead of ShapleyOG in minutes} \\
Game ($p$) $\backslash$ $T$ & 512 & 1024 & 2048 & 4096 & 8192 \\
\midrule
DINOv2-16 (16) & 0.2 & 0.3 & 0.7 & 1.8 & 6.0 \\
Bankruptcy (64) & 0.2 & 0.5 & 1.2 & 3.5 & 14.1 \\
MIC (111) & 0.3 & 0.6 & 1.4 & 4.2 & 15.1 \\
\bottomrule
\end{tabular}
}
  \end{minipage}
  \caption{Left: NMSE (geometric mean over the $14$ games with the full budget grid and $30$ seeds) vs.\ total computation time, i.e., overhead plus $\budget$ times an assumed cost per evaluation of $0.01$, $0.1$, $1$, and $10$ seconds; markers are the budgets $256$ to $8192$. Top right: mean overhead of \methodname{} over $30$ seeds for three games of different size. Bottom right: ablation path from \shapleig{} to \methodname{} on Crime ($\np = 101$, $\budget = 2048$), adding one measure at a time: overhead (median over $10$ seeds, left axis) and NMSE (geometric mean, right axis).}
  \label{fig:pareto}
\end{figure}%
\runin{Total Computation Time.}
Adaptivity costs overhead, so \cref{fig:pareto} (left) plots the NMSE against the total computation time, measured on our hardware (\cref{app:exp_compute}), for assumed costs per value function evaluation ranging from $0.01$\,s to $10$\,s, e.g., from high-throughput value functions over forward passes of a model to retraining small models. At $0.01$\,s, the overhead of \methodname{} dominates, and OddSHAP reaches a given accuracy faster. From $0.1$\,s on, \methodname{} attains a lower NMSE than every competitor at equal total computation time. The overhead of \methodname{} itself (\cref{fig:pareto}, top right) stays below a minute up to $1024$ evaluations even for $111$ players, takes about a minute at $2048$ and a few minutes at $4096$ evaluations, or even at $8192$ for few players, and around a quarter of an hour at $8192$ for the larger games. \Cref{app:res_overhead} further splits it into its components.
\runin{Ablation Path from \shapleig{}.}
To attribute the speed-up to the individual measures introduced with \methodname{}, we start from \shapleig{} and add one measure at a time: the sparse refit schedule, the incremental EIG updates between refits together with the candidate resampling at refits, the GL quadrature for the SV prior covariances, paired acquisition under the full GP, and finally the odd GP, which yields \methodname{}. Since \shapleig{} needs up to $23$ hours for one run to $2048$ evaluations, this study uses all $15$ games with $10$ seeds and budgets up to $2048$. \Cref{fig:pareto} (bottom right) shows the Crime game: along the path, the NMSE remains comparable to that of \shapleig{}, while the overhead falls from $18$ hours to less than a minute, i.e., by a factor of $1329$. \Cref{app:res_ablation} reports the details for all games. Overall, \methodname{} is $233$ to $1634$ times faster than \shapleig{}, and its NMSE relative to \shapleig{} ranges from $0.53$ to $1.48$ across the games, with a geometric mean of $1.03$, so that overall the accuracy is maintained. On the larger games, the refit schedule contributes most to the speed-up, followed by the odd GP, while the GL quadrature and the incremental updates also yield substantial speed-ups.
\runin{Quadrature at Large $\np$.}
In the ablation path, the benefit of the GL quadrature over the ESP route remains moderate, since both are cheap for the small to moderate numbers of players of these games. Our quadrature study in \cref{app:res_quadrature} therefore examines it on a further game, QSAR TID11 with $1024$ players, and analyzes how approximate GL quadrature with fewer nodes performs in practice. There, $8$ instead of the $514$ GL nodes leave the SV prior covariances exact to machine precision, and hence the selected coalitions and the NMSE unchanged, while the computation of the SV prior covariances drops from $12$ minutes to $14$ seconds per run. Computing the SV prior covariances with the ESP route of \shapleig{} instead exceeded a $24$-hour cap without completing them even once. The quadrature thus makes games of this size practically feasible, and a small fraction of the exact node count suffices.
\section{Conclusion} \label{sec:conclusion}
The regime in which adaptive SV estimation pays off can be broadened substantially: \methodname{} keeps the accuracy of \shapleig{} and outperforms the non-adaptive estimators on most games and budgets, while reducing the overhead of \shapleig{} by orders of magnitude, so that in typical scenarios the evaluation cost outweighs the overhead already at a tenth of a second per evaluation. Two questions are open. First, on a few small games \methodname{} is dominated by non-adaptive estimators, especially at large budgets, and it is unclear which structural properties of a game are responsible. Second, \methodname{} scales to games with $1000$ players, but its overhead still grows disproportionately with the budget and the number of players, so very large budgets and games remain costly and whether the accuracy advantage persists there is open.
\newpage
\subsection*{AI use statement}

In this work, we used generative AI tools to assist with research ideation and programming-related  execution, as well as to draft and improve parts of the manuscript. All AI-assisted research contributions and text were reviewed and verified by the authors. We take responsibility for the final content of this work, including text, claims, and artifacts produced with the aid of generative AI.

\subsection*{Ethics statement}

To the best of our knowledge, this work does not raise any specific ethical concerns requiring further discussion.

\subsection*{Reproducibility statement}

The complete code of \methodname{}, including the configuration files of all presented experiments, the scripts for preparing the games, and the aggregation and plotting scripts, will be made publicly available, so that all results are reproducible.

\subsubsection*{Acknowledgments}
Fabian Fumagalli acknowledges funding by the Deutsche Forschungsgemeinschaft (DFG, German Research Foundation): TRR 318/3 2026 -- 438445824.

\bibliography{iclr2027_conference,bibliography}
\bibliographystyle{iclr2027_conference}

\newpage

\appendix
\crefalias{section}{appendix}\crefalias{subsection}{appendix}\crefalias{subsubsection}{appendix}
\setcounter{theorem}{0}
\renewcommand{\thetheorem}{\thesection.\arabic{theorem}}
\makeatletter \@addtoreset{theorem}{section} \makeatother  %
\section{Background} \label{app:background}
\subsection{Gaussian Processes} \label{app:gp}
A Gaussian process (GP; \citealp{WR95,RW06}) over functions on $\binp$ is fully characterized by a mean function $m: \binp \to \R$ and a positive definite kernel $\kf: \binp \times \binp \to \R$ with hyperparameters $\hp$, and we write $\vf \sim \GP(m, \kf)$. Given data $\Dt$ with coalitions $\Xt \in \bin^{t \times \np}$ and targets $\Yt \in \Rp{t}$, the latter corrupted by additive Gaussian noise $\epsilon \sim \N(0, \noise)$ that is i.i.d.\ across observations, the posterior is again a GP, and the posterior predictive distribution at any $s$ coalitions $\X_\star \in \bin^{s \times \np}$ is a multivariate Gaussian, $\vf(\X_\star) \mid \Dt \sim \N_s(\boldsymbol{\mu}_{\vf(\X_\star) \mid \Dt}, \boldsymbol{\Sigma}_{\vf(\X_\star) \mid \Dt})$. For a zero mean function, $m = 0$,
\begin{align}
    \boldsymbol{\mu}_{\vf(\X_\star) \mid \Dt} &= \K(\X_\star, \Xt) \, \KXXninv \Yt, \label{eq:gp_ppd_mean} \\
    \boldsymbol{\Sigma}_{\vf(\X_\star) \mid \Dt} &= \K(\X_\star, \X_\star) - \K(\X_\star, \Xt) \, \KXXninv \K(\Xt, \X_\star), \label{eq:gp_ppd_cov}
\end{align}
where, for matrices $\mathbf{U}$ and $\mathbf{V}$ whose rows are coalitions, $\K(\mathbf{U}, \mathbf{V})$ denotes the kernel matrix whose $(i,j)$-th entry is $\kf(\mathbf{U}_i, \mathbf{V}_j)$, with $\mathbf{U}_i$ the $i$-th row of $\mathbf{U}$.
\paragraph{Computational Cost.}
Forming the kernel matrices costs $\bigO(t^2 \np)$ for the training kernel matrix $\KXXn \in \Rmat{t}{t}$, $\bigO(s t \np)$ for the cross kernel matrix $\K(\X_\star, \Xt)$, and $\bigO(s^2 \np)$ for $\K(\X_\star, \X_\star)$, or $\bigO(s)$ for its diagonal only, which is constant, $\kf(\z, \z) = \oscale$, for the Hamming kernel. Given the kernel matrices, the training kernel matrix is typically decomposed via a Cholesky factorization in $\bigO(t^3)$. The posterior mean in \cref{eq:gp_ppd_mean} then requires one triangular solve with the targets in $\bigO(t^2)$ and a matrix-vector product in $\bigO(s t)$, i.e., $\bigO(t^3 + s t)$ in total. The posterior covariance in \cref{eq:gp_ppd_cov} requires triangular solves with $s$ right-hand sides in $\bigO(s t^2)$ and a matrix product in $\bigO(s^2 t)$, i.e., $\bigO(t^3 + s t^2 + s^2 t)$ in total. If only the predictive variances, i.e., the diagonal of \cref{eq:gp_ppd_cov}, are required, the matrix product reduces to column sums of squares in $\bigO(s t)$, i.e., $\bigO(t^3 + s t^2)$ in total.
\paragraph{Hamming Kernel.} The Hamming kernel of \cref{eq:hamming_kernel} can be written as
\begin{align} \label{eq:hamming_kernel_full}
    \kf(\z, \z') = \oscale \prod_{j \in \players} \exp\big( -\Ind{z_j \neq z'_j} / \ls_j \big) = \oscale \prod_{j \in \players} \big( \alpha_j \Ind{z_j = z'_j} + \beta_j \Ind{z_j \neq z'_j} \big)
\end{align}
with $\alpha_j := 1$ and $\beta_j := \exp(-1/\ls_j)$, i.e., it quantifies the similarity of two coalitions via a weighted Hamming distance $\sum_{j} \Ind{z_j \neq z'_j} / \ls_j$. Player $j$ contributes the factor $\alpha_j$ if both coalitions agree on its membership and $\beta_j$ if they differ, so that agreement leaves the covariance unchanged, whereas each disagreement reduces it by the player-specific factor $\beta_j < 1$. Small lengthscales thus encode players whose membership changes the value function strongly. The general form with agreement factors $\alpha_j$ and disagreement factors $\beta_j$ is used in \cref{app:quad_proofs}.
\paragraph{Hyperparameter Fitting.}
The kernel hyperparameters $\hp$ are learned by maximizing the log marginal likelihood $\log p(\Yt \mid \Xt, \hp)$ of the data or, given a prior $p(\hp)$, the log posterior $\log p(\Yt \mid \Xt, \hp) + \log p(\hp)$, i.e., by maximum a posteriori (MAP) estimation. Both objectives and their gradients with respect to $\hp$ are available in closed form~\citep{RW06}, so that gradient-based optimizers apply, where each objective evaluation requires a Cholesky factorization of the training kernel matrix in $\bigO(t^3)$.
\paragraph{Quasi-Noiseless GPs.}
A noiseless GP is the limiting case $\noise = 0$, in which the posterior mean interpolates the training data exactly if the training kernel matrix is non-singular~\citep{S99,RW06}. In practice, $\noise$ is typically fixed to a small positive constant, which keeps the Cholesky factorization numerically stable and is equivalent to assuming a small observation noise; the resulting GP is referred to as \emph{quasi-noiseless}.
\subsection{Adaptive SV Estimation with \shapleig{}} \label{app:shapleig}
In each iteration, \shapleig{} evaluates the EIG of \cref{eq:eig_closed_form,eq:Q_def} for all candidates $\z \in \cands$ under the full GP with the current hyperparameters $\hp$,
\begin{align}
    \EIGsv(\z) &\propto \log \Var(\vf(\z) \mid \Dt) - \log \big( \Var(\vf(\z) \mid \Dt) - \Qz(\z) \big), \label{eq:app_eig} \\
    \Qz(\z) &= \Cov(\sv, \vf(\z) \mid \Dt)\trans \, \Cov(\sv, \sv \mid \Dt)^{-1} \, \Cov(\sv, \vf(\z) \mid \Dt) \notag \\
    &= \big( \ASig \ez \big)\trans \big( \ASigA \big)^{-1} \big( \ASig \ez \big). \notag
\end{align}
Ultimately, \shapleig{} returns the SV posterior of \cref{eq:sv_posterior_mean,eq:sv_posterior_cov} as its final output.

In the following, we describe how the EIG over the candidate set $\cands$ is computed in five steps, and how the SV posterior is then obtained from the same quantities. For every step, we state the size of the involved quantities, their cost, and what is recycled from previous steps. We collect the basis vectors $\ez$ of the candidates as columns of $\mathbf{E}_{\cands} \in \Rmat{\twop}{\ncands}$ and abbreviate the training kernel matrix as $\widetilde{\K} := \KXXn \in \Rmat{t}{t}$.
\paragraph{Step 1: Kernel Matrices.}
The training kernel matrix $\widetilde{\K}$ is formed in $\bigO(t^2 \np)$ and factorized via Cholesky as $\widetilde{\K} = \mathbf{L} \mathbf{L}\trans$ in $\bigO(t^3)$, and the candidate kernel matrix $\K(\Xt, \cands) \in \Rmat{t}{\ncands}$ is formed in $\bigO(t \ncands \np)$.
\paragraph{Step 2: SV Prior Covariances.}
The SV prior covariances $\A \KZZ \A\trans \in \Rmat{\np}{\np}$, $\A \KZX \in \Rmat{\np}{t}$, and $\A \K(\Zall, \cands) \in \Rmat{\np}{\ncands}$ are computed via ESPs in $\bigO(\np^4)$, $\bigO(t \np^2)$, and $\bigO(\ncands \np^2)$, respectively (\cref{sec:shapleig}).
\paragraph{Step 3: Half-Solves.}
The half-solve $\Y_{\X} := \mathbf{L}^{-1} (\A \KZX)\trans \in \Rmat{t}{\np}$ costs $\bigO(t^2 \np)$ and reuses $\mathbf{L}$ from Step~1 and $\A \KZX$ from Step~2, and the half-solve $\Y_{\cands} := \mathbf{L}^{-1} \K(\Xt, \cands) \in \Rmat{t}{\ncands}$ costs $\bigO(t^2 \ncands)$ and reuses $\mathbf{L}$ and $\K(\Xt, \cands)$ from Step~1.
\paragraph{Step 4: Posterior Covariances.}
The EIG of \cref{eq:app_eig} involves three posterior quantities: the predictive variance $\Var(\vf(\z) \mid \Dt)$ of the candidate, the cross-covariance $\Cov(\sv, \vf(\z) \mid \Dt) = \ASig \ez$ between the SVs and the candidate, and the SV posterior covariance $\Cov(\sv, \sv \mid \Dt) = \ASigA = \Sigsv$. They are computed in the same way, since by \cref{eq:gp_ppd_cov} each of them is a prior covariance minus a product of the form $\mathbf{M}\trans \widetilde{\K}^{-1} \mathbf{N}$, with the inverse training kernel matrix in the middle and possibly different matrices on either side. Every such product equals the product $(\mathbf{L}^{-1} \mathbf{M})\trans (\mathbf{L}^{-1} \mathbf{N})$ of two half-solves, so that from Steps~2 and 3, the three quantities are obtained for all candidates by matrix products:
\begin{align}
    \mathbf{v} &:= \big( \Var(\vf(\z) \mid \Dt) \big)_{\z \in \cands} = \oscale \onevec - \diag\big( \Y_{\cands}\trans \Y_{\cands} \big) \in \Rp{\ncands}, && \bigO(t \ncands), \label{eq:app_var} \\
    \mathbf{B} &:= \A \Sigvf \mathbf{E}_{\cands} = \A \K(\Zall, \cands) - \Y_{\X}\trans \Y_{\cands} \in \Rmat{\np}{\ncands}, && \bigO(t \np \ncands), \label{eq:app_cross} \\
    \Sigsv &= \ASigA = \A \KZZ \A\trans - \Y_{\X}\trans \Y_{\X} \in \Rmat{\np}{\np}, && \bigO(t \np^2), \label{eq:app_prior}
\end{align}
where $\diag(\cdot)$ extracts the diagonal, i.e., the column sums of squares, and $\kf(\z, \z) = \oscale$ for the Hamming kernel. \Cref{eq:app_prior} is the SV posterior covariance of \cref{eq:sv_posterior_cov}.
\paragraph{Step 5: EIG.}
With the Cholesky factor $\mathbf{L}_{\sv}$ of $\Sigsv$, computed in $\bigO(\np^3)$, one triangular solve with $\ncands$ right-hand sides in $\bigO(\np^2 \ncands)$ yields
\begin{align}
    \mathbf{q} &:= \big( \Qz(\z) \big)_{\z \in \cands} = \diag\big( (\mathbf{L}_{\sv}^{-1} \mathbf{B})\trans (\mathbf{L}_{\sv}^{-1} \mathbf{B}) \big) \in \Rp{\ncands}, \label{eq:app_q} \\
    \big( \EIGsv(\z) \big)_{\z \in \cands} &\propto \log \mathbf{v} - \log(\mathbf{v} - \mathbf{q}), \notag
\end{align}
with elementwise logarithms. In total, the EIG over $\cands$ costs $\bigO(t^3 + t^2 \ncands + \np^4)$ for $t \geq \np$, as stated in \cref{sec:shapleig}.
\paragraph{SV Posterior.}
The SV posterior of \cref{eq:sv_posterior_mean,eq:sv_posterior_cov}, the final output of \shapleig{}, consists of the same components: its covariance $\Sigsv$ is already available from \cref{eq:app_prior}, and its mean
\begin{align} \label{eq:app_sv_mean}
    \musv = \Y_{\X}\trans \mathbf{w}, \qquad \mathbf{w} := \mathbf{L}^{-1} \Yt \in \Rp{t},
\end{align}
additionally requires the triangular solve for $\mathbf{w}$ in $\bigO(t^2)$ and the product with $\Y_{\X}$ in $\bigO(t \np)$. Given Steps~1 to 3, the SV posterior is thus available in $\bigO(t^3 + t^2 \np + \np^4)$, as stated in \cref{sec:shapleig}.
\paragraph{Recycling between Iterations.}
Since \shapleig{} refits $\hp$ in every iteration, it generally recomputes all steps. Whenever $\hp$ is kept fixed, the kernel matrices $\widetilde{\K}$ and $\K(\Xt, \cands)$ of Step~1 change by a single row and column, and Step~2 is recycled entirely, as $\A \KZZ \A\trans$ remains valid and $\A \KZX$ and $\A \K(\Zall, \cands)$ change by a single column. \shapleig{} exploits this but recomputes the Cholesky factorization of Step~1, both half-solves of Step~3, and Steps~4 and 5.
\clearpage
\section{Proofs and Details for \cref{sec:oddgp}} \label{app:oddgp_proofs}
Throughout this section, the hyperparameters $\hp$ are fixed, and we use the notation of \cref{sec:oddgp}. For the convenience of the reader, we restate the results of the main text before their proofs.
\subsection{Complement Invariance of the Hamming Kernel} \label{app:complement}
The Hamming kernel depends on its arguments only through the mismatch indicators $\Ind{z_j \neq z'_j}$, $j \in \players$ (\cref{app:gp}), which complementing both arguments leaves unchanged. Hence it is \emph{complement-invariant}, $\kf(\zbar, \zbar') = \kf(\z, \z')$, and consequently $\kf(\zbar, \z') = \kf(\z, \zbar')$ for all $\z, \z' \in \binp$. Both identities are used throughout this appendix.
\subsection{Even-Odd Decomposition of the GP Prior} \label{app:decomp_proof}
\noindent\textbf{\Cref{lem:decomp}} (Even-odd decomposition of the Hamming kernel GP). \emph{Let $\vf \sim \GP(m, \kf)$ be the Hamming kernel GP with constant mean $m \in \R$. Then $\vf = \vfodd + \vfeven$ with independent GPs $\vfodd$ and $\vfeven$, where the odd component follows the odd GP $\vfodd \sim \GP(0, \ko)$ with $\ko$ as in \cref{eq:odd_kernel}.}

The lemma follows from the complement invariance of the Hamming kernel~\citep{GBRC12}: by (anti-)symmetrization of the kernel~\citep{HB07,KGNN21}, the GP decomposes into independent even and odd GPs~\citep[Cor.~2(b)]{GRSDL16}. We include a direct proof in our notation for completeness.
\begin{proof}
We consider the GP prior $\vf \sim \GP(m, \kf)$ on the value function, i.e., $\E[\vf(\z)] = m$ and $\Cov(\vf(\z), \vf(\z')) = \kf(\z, \z')$ for all $\z, \z' \in \binp$, and proceed in four steps.

\emph{Step 1: The complemented process.} Define the complemented process $\bar{\vf}(\z) := \vf(\zbar)$, i.e., the value function evaluated at the complement coalition. Since $\bar{\vf}$ consists of the same random variables as $\vf$, only indexed differently, $\vf$ and $\bar{\vf}$ are jointly Gaussian processes~\citep{RW06}. Their moments follow directly from those of $\vf$ and \cref{app:complement}:
\begin{align*}
    \E[\bar{\vf}(\z)] &= \E[\vf(\zbar)] = m, \\
    \Cov\big(\bar{\vf}(\z), \bar{\vf}(\z')\big) &= \Cov\big(\vf(\zbar), \vf(\zbar')\big) = \kf(\zbar, \zbar') = \kf(\z, \z'), \\
    \Cov\big(\vf(\z), \bar{\vf}(\z')\big) &= \Cov\big(\vf(\z), \vf(\zbar')\big) = \kf(\z, \zbar').
\end{align*}
In particular, $\bar{\vf} \sim \GP(m, \kf)$ has the same distribution as $\vf$, and the two processes are correlated through $\kf(\z, \zbar')$.

\emph{Step 2: The odd and even components are GPs.} By \cref{eq:odd_even}, the odd and even components are the linear transformations
\begin{align*}
    \vfodd = \tfrac{1}{2} \big( \vf - \bar{\vf} \big),
    \qquad
    \vfeven = \tfrac{1}{2} \big( \vf + \bar{\vf} \big)
\end{align*}
of the jointly Gaussian pair $(\vf, \bar{\vf})$. Linear transformations of jointly Gaussian processes are again jointly Gaussian processes~\citep{RW06,GRD16}, so $\vfodd$ and $\vfeven$ are GPs, and $\vf = \vfodd + \vfeven$ holds by construction. It remains to compute their means and covariances.

\emph{Step 3: Means and covariances.} By linearity of expectation and Step 1,
\begin{align*}
    \E[\vfodd(\z)] = \tfrac{1}{2} (m - m) = 0,
    \qquad
    \E[\vfeven(\z)] = \tfrac{1}{2} (m + m) = m.
\end{align*}
By bilinearity of the covariance and the moments of Step 1,
\begin{align*}
    \Cov\big(\vfodd(\z), \vfodd(\z')\big)
    &= \tfrac{1}{4} \Big[ \Cov\big(\vf(\z), \vf(\z')\big) - \Cov\big(\vf(\z), \bar{\vf}(\z')\big) \\
    &\qquad\quad - \Cov\big(\bar{\vf}(\z), \vf(\z')\big) + \Cov\big(\bar{\vf}(\z), \bar{\vf}(\z')\big) \Big] \\
    &= \tfrac{1}{4} \Big[ \kf(\z, \z') - \kf(\z, \zbar') - \kf(\zbar, \z') + \kf(\z, \z') \Big] \\
    &= \tfrac{1}{4} \Big[ \kf(\z, \z') - \kf(\z, \zbar') - \kf(\z, \zbar') + \kf(\z, \z') \Big] \\
    &= \tfrac{1}{2} \big( \kf(\z, \z') - \kf(\z, \zbar') \big) = \ko(\z, \z'),
\end{align*}
where the third equality uses $\kf(\zbar, \z') = \kf(\z, \zbar')$ from \cref{app:complement}; $\ko$ is the \emph{antisymmetrization} of $\kf$ under complementation~\citep{HB07,KGNN21}. Analogously,
\begin{align*}
    \Cov\big(\vfeven(\z), \vfeven(\z')\big)
    &= \tfrac{1}{4} \Big[ \kf(\z, \z') + \kf(\z, \zbar') + \kf(\zbar, \z') + \kf(\z, \z') \Big] \\
    &= \tfrac{1}{2} \big( \kf(\z, \z') + \kf(\z, \zbar') \big) =: \ke(\z, \z'),
\end{align*}
the \emph{symmetrization} of $\kf$~\citep{HB07,KGNN21}, and for the cross-covariance
\begin{align*}
    \Cov\big(\vfodd(\z), \vfeven(\z')\big)
    &= \tfrac{1}{4} \Big[ \Cov\big(\vf(\z), \vf(\z')\big) + \Cov\big(\vf(\z), \bar{\vf}(\z')\big) \\
    &\qquad\quad - \Cov\big(\bar{\vf}(\z), \vf(\z')\big) - \Cov\big(\bar{\vf}(\z), \bar{\vf}(\z')\big) \Big] \\
    &= \tfrac{1}{4} \Big[ \kf(\z, \z') + \kf(\z, \zbar') - \kf(\z, \zbar') - \kf(\z, \z') \Big] = 0.
\end{align*}
Hence $\vfodd \sim \GP(0, \ko)$ and $\vfeven \sim \GP(m, \ke)$.

\emph{Step 4: Independence and kernel decomposition.} Since $\vfodd$ and $\vfeven$ are jointly Gaussian (Step 2) with vanishing cross-covariance (Step 3), they are independent~\citep{RW06}; this is the independence criterion for projections of GPs~\citep[Cor.~2(b)]{GRSDL16}. Finally, adding the two kernels in \cref{eq:odd_kernel} gives $\ko(\z, \z') + \ke(\z, \z') = \kf(\z, \z')$, which is consistent with $\vf = \vfodd + \vfeven$ and the independence of the two components, as the sum of independent GPs has the sum of their kernels as covariance function~\citep[Sec.~4.2.4]{RW06}.
\end{proof}
\begin{lemma}[The posterior mean of the odd GP is odd] \label{lem:odd_gp_odd}
Let $\vfodd \sim \GP(0, \ko)$ with $\ko$ as in \cref{eq:odd_kernel}. Then, for any dataset $\Dtodd$ of observations of $\vfodd$ at coalitions $\Xtodd$ with targets $\Ytodd$ and noise variance $\noise \geq 0$, the posterior mean $\mu_{\odd}(\z) := \mu_{\vfodd(\z) \mid \Dtodd} = \E[\vfodd(\z) \mid \Dtodd]$ satisfies $\mu_{\odd}(\zbar) = -\mu_{\odd}(\z)$ for all $\z \in \binp$.
\end{lemma}
The lemma is a special case of \citet[Props.~3 and 8]{GRD16}.
\begin{proof}
By complement invariance (\cref{app:complement}),
\begin{align*}
    \ko(\zbar, \z') = \tfrac{1}{2} \big( \kf(\zbar, \z') - \kf(\zbar, \zbar') \big) = \tfrac{1}{2} \big( \kf(\z, \zbar') - \kf(\z, \z') \big) = -\ko(\z, \z'),
\end{align*}
i.e., $\ko$ is odd in its first argument and, by symmetry, in its second. Consequently, $\Ko(\zbar, \Xtodd) = -\Ko(\z, \Xtodd)$, where $\Ko(\mathbf{U}, \mathbf{V})$ denotes the kernel matrix of $\ko$ in analogy to $\K(\mathbf{U}, \mathbf{V})$ in \cref{app:gp}, and the posterior mean $\mu_{\odd}(\z) = \Ko(\z, \Xtodd) \big[ \Ko(\Xtodd, \Xtodd) + \noise \I \big]^{-1} \Ytodd$ (\cref{eq:gp_ppd_mean}) satisfies $\mu_{\odd}(\zbar) = -\mu_{\odd}(\z)$.
\end{proof}
\subsection{Posterior Equivalence under Paired Sampling} \label{app:posterior_proof}
\noindent\textbf{\Cref{thm:odd_posterior}} (SV posterior equivalence under paired sampling). \emph{Let $\vf \sim \GP(m, \kf)$ be the Hamming kernel GP with fixed hyperparameters $\hp$ and constant mean $m \in \R$, and let $\Dt$ be paired with odd dataset $\Dtodd$. Then $p(\sv \mid \Dt) = p(\sv \mid \Dtodd)$, where the left-hand side denotes the SV posterior of the full GP $\vf$ conditioned on $\Dt$, and the right-hand side the SV posterior of the odd GP $\vfodd \sim \GP(0, \ko)$ from \cref{eq:odd_kernel} conditioned on the odd dataset $\Dtodd$.}

The proof uses the Hamming kernel only through its complement invariance (\cref{app:complement}).
\begin{proof}
We prove the statement for observations corrupted by i.i.d.\ Gaussian noise with variance $\noise \geq 0$; the theorem as stated in the main text is the case $\noise = 0$. Let $\reps$ be the set of representatives from \cref{sec:oddgp}, i.e., the rows of $\Xtodd$. The paired dataset consists of the observations
\begin{align*}
    y(\z) = \vf(\z) + \epsilon_{\z}, \qquad y(\zbar) = \vf(\zbar) + \epsilon_{\zbar}, \qquad \z \in \reps,
\end{align*}
with all $\epsilon_{\z}, \epsilon_{\zbar} \overset{\text{i.i.d.}}{\sim} \N(0, \noise)$ independent of $\vf$ (for $\noise = 0$ the observations are exact), i.e., $\Dt = \{(\z, y(\z)), (\zbar, y(\zbar)) : \z \in \reps\}$. For each pair, define the odd and even targets
\begin{align*}
    y_{\odd}(\z) &:= \tfrac{1}{2}\big(y(\z) - y(\zbar)\big) = \vfodd(\z) + \epsilon_{\odd}(\z), \\
    y_{\even}(\z) &:= \tfrac{1}{2}\big(y(\z) + y(\zbar)\big) = \vfeven(\z) + \epsilon_{\even}(\z),
\end{align*}
with $\epsilon_{\odd}(\z) := \tfrac{1}{2}(\epsilon_{\z} - \epsilon_{\zbar})$ and $\epsilon_{\even}(\z) := \tfrac{1}{2}(\epsilon_{\z} + \epsilon_{\zbar})$, and collect them, together with the representatives, in the odd and even datasets
\begin{align*}
    \Dtodd := \{(\z, y_{\odd}(\z)) : \z \in \reps\}
    \qquad \text{and} \qquad
    \Dteven := \{(\z, y_{\even}(\z)) : \z \in \reps\},
\end{align*}
with targets stacked in $\Ytodd, \Yteven \in \Rp{\nodd}$ in the row order of $\Xtodd$. For $\noise = 0$, $\Dtodd$ coincides with the odd dataset of the main text. 

\emph{Step 1: Conditioning on $\Dt$ is conditioning on $(\Dtodd, \Dteven)$.} For each $\z \in \reps$, the observations and the odd and even targets are related by the linear map
\begin{align*}
    \begin{pmatrix} y_{\odd}(\z) \\ y_{\even}(\z) \end{pmatrix}
    = \tfrac{1}{2} \begin{pmatrix} 1 & -1 \\ 1 & 1 \end{pmatrix}
    \begin{pmatrix} y(\z) \\ y(\zbar) \end{pmatrix},
    \qquad \text{with inverse} \qquad
    \begin{pmatrix} y(\z) \\ y(\zbar) \end{pmatrix}
    = \begin{pmatrix} 1 & 1 \\ -1 & 1 \end{pmatrix}
    \begin{pmatrix} y_{\odd}(\z) \\ y_{\even}(\z) \end{pmatrix}.
\end{align*}
Since the representatives are the same on both sides, $\Dt$ and $(\Dtodd, \Dteven)$ are in one-to-one correspondence and thus carry exactly the same information. Consequently, conditioning on either yields the same posterior,
\begin{align*}
    p(\sv \mid \Dt) = p(\sv \mid \Dtodd, \Dteven).
\end{align*}

\emph{Step 2: The noise decomposes independently.} The vector $(\epsilon_{\odd}(\z), \epsilon_{\even}(\z))$ is Gaussian with
\begin{align*}
    \Var(\epsilon_{\odd}(\z)) &= \Var(\epsilon_{\even}(\z)) = \tfrac{1}{4}(\noise + \noise) = \tfrac{\noise}{2}, \\
    \Cov(\epsilon_{\odd}(\z), \epsilon_{\even}(\z)) &= \tfrac{1}{4}(\noise - \noise) = 0,
\end{align*}
so $\epsilon_{\odd}(\z)$ and $\epsilon_{\even}(\z)$ are independent~\citep{RW06}, and both are independent across $\z \in \reps$ and of $\vf$.

\emph{Step 3: The even dataset is uninformative about the SVs.} By \cref{lem:decomp} and Step 2, $(\vfodd, \epsilon_{\odd})$ and $(\vfeven, \epsilon_{\even})$ are independent. Since the SVs of any even function vanish \citep{FBKRW26},
\begin{align*}
    \sv(\vf) = \A \vfv = \A \vfodd(\Zall) + \A \vfeven(\Zall) = \A \vfodd(\Zall) + \mathbf{0} = \sv(\vfodd),
\end{align*}
so $\sv$ and $\Dtodd$ are functions of $(\vfodd, \epsilon_{\odd})$ only, whereas $\Dteven$ is a function of $(\vfeven, \epsilon_{\even})$. Hence $(\sv, \Dtodd)$ is independent of $\Dteven$, and
\begin{align*}
    p(\sv \mid \Dtodd, \Dteven) = p(\sv \mid \Dtodd).
\end{align*}
Moreover, by \cref{lem:decomp} and Step 2, $\Dtodd$ consists of observations of $\vfodd \sim \GP(0, \ko)$ at $\Xtodd$ with i.i.d.\ Gaussian noise of variance $\noise / 2$, so that the right-hand side is the SV posterior of the odd GP. Together with Step 1, this gives $p(\sv \mid \Dt) = p(\sv \mid \Dtodd, \Dteven) = p(\sv \mid \Dtodd)$, which proves the claim.

Explicitly, standard GP regression (\cref{app:gp}; \citealp{RW06}) and the linearity of the SVs in the value function (\cref{sec:sv}) yield $\sv \mid \Dtodd \sim \N_{\np}(\musvodd, \Sigsvodd)$ with
\begin{align}
    \musvodd &= \A \Ko(\Zall, \Xtodd) \, \Kot^{-1} \Ytodd, \label{eq:odd_posterior_mean} \\
    \Sigsvodd &= \A \Ko(\Zall, \Zall) \A\trans - \A \Ko(\Zall, \Xtodd) \, \Kot^{-1} \Ko(\Xtodd, \Zall) \A\trans, \label{eq:odd_posterior_cov}
\end{align}
where $\Kot := \Ko(\Xtodd, \Xtodd) + \tfrac{\noise}{2} \I$.
\end{proof}
\subsection{Odd Expected Information Gain} \label{app:odd_eig}
\noindent\textbf{\Cref{thm:odd_eig}} (EIG equivalence under paired acquisition; stated with i.i.d.\ Gaussian observation noise as in \cref{app:posterior_proof}). \emph{Under the assumptions of \cref{thm:odd_posterior}, let $\z \in \binp$ be a candidate coalition and let $y_{\odd}(\z) := \tfrac{1}{2}(y(\z) - y(\zbar))$ denote the odd target obtained from evaluating both $\z$ and $\zbar$. Then
\begin{align*}
    \MI\big(\sv; y(\z), y(\zbar) \mid \Dt\big) = \MI\big(\sv; y_{\odd}(\z) \mid \Dtodd\big) =: \EIGsvodd(\z),
\end{align*}
where the right-hand side is the EIG under the odd GP conditioned on the odd dataset. Moreover, $\EIGsvodd(\z) = \EIGsvodd(\zbar)$.}
\begin{proof}
For the first claim, the invertible map of Step 1 in \cref{app:posterior_proof} gives $\MI(\sv; y(\z), y(\zbar) \mid \Dt) = \MI(\sv; y_{\odd}(\z), y_{\even}(\z) \mid \Dt)$, since mutual information is invariant under invertible transformations of its arguments~\citep{CT06}. By the chain rule of mutual information~\citep{CT06},
\begin{align*}
    \MI\big(\sv; y_{\odd}(\z), y_{\even}(\z) \mid \Dt\big) = \MI\big(\sv; y_{\odd}(\z) \mid \Dt\big) + \MI\big(\sv; y_{\even}(\z) \mid y_{\odd}(\z), \Dt\big).
\end{align*}
By Step 3 of the proof of \cref{thm:odd_posterior}, and since Step 2 of that proof applies verbatim to the noise of the candidate pair, $(\sv, y_{\odd}(\z), \Dtodd)$ is a function of $(\vfodd, \epsilon_{\odd})$, which is independent of $(\vfeven, \epsilon_{\even})$ and hence of $(y_{\even}(\z), \Dteven)$. Thus the conditional mutual information on the right vanishes, and conditioning on $\Dt = (\Dtodd, \Dteven)$ yields the same joint law of $(\sv, y_{\odd}(\z))$ as conditioning on $\Dtodd$ alone, i.e., $\MI(\sv; y_{\odd}(\z) \mid \Dt) = \MI(\sv; y_{\odd}(\z) \mid \Dtodd)$. The latter is the EIG under the odd GP, since $\Dtodd$ consists of observations of $\vfodd \sim \GP(0, \ko)$ with i.i.d.\ Gaussian noise.

For the second claim, recall from \cref{eq:eig_closed_form,eq:Q_def} that the EIG depends on the candidate only through its posterior predictive variance and through $\Cov(\sv, \vfodd(\z) \mid \Dtodd)$, the latter entering quadratically via $\Qz(\z)$. Since $\ko$ is odd in each argument (proof of \cref{lem:odd_gp_odd}), $\Ko(\zbar, \Xtodd) = -\Ko(\z, \Xtodd)$ and $\ko(\zbar, \zbar) = \ko(\z, \z)$, so that by \cref{eq:gp_ppd_cov} $\Var(\vfodd(\zbar) \mid \Dtodd) = \Var(\vfodd(\z) \mid \Dtodd)$. Likewise, $\Cov(\sv, \vfodd(\z) \mid \Dtodd) = \A \big[ \Ko(\Zall, \z) - \Ko(\Zall, \Xtodd) \, \Kot^{-1} \Ko(\Xtodd, \z) \big]$, where both terms change sign under $\z \mapsto \zbar$, so that $\Cov(\sv, \vfodd(\zbar) \mid \Dtodd) = -\Cov(\sv, \vfodd(\z) \mid \Dtodd)$. Hence $\Qz(\zbar) = \Qz(\z)$, and the EIG values coincide.
\end{proof}

\subsection{Covariance Identities} \label{app:identities}
\begin{proposition}[Covariance identities] \label{prop:odd_identities}
Under the assumptions of \cref{lem:decomp}, for all $\z \in \binp$,
\begin{align*}
    \A \Ko(\Zall, \z) = \A \K(\Zall, \z)
    \qquad \text{and} \qquad
    \A \Ko(\Zall, \Zall) \A\trans = \A \K(\Zall, \Zall) \A\trans.
\end{align*}
\end{proposition}
\begin{proof}
For the first identity, write $\vf(\z) = \vfodd(\z) + \vfeven(\z)$ and use $\A \vfv = \A \vfodd(\Zall)$ (Step 3 in \cref{app:posterior_proof}) together with \cref{lem:decomp}:
\begin{align*}
    \A \K(\Zall, \z) &= \Cov\big(\A \vfv, \vf(\z)\big) = \Cov\big(\A \vfv, \vfodd(\z) + \vfeven(\z)\big) \\
    &= \Cov\big(\A \vfodd(\Zall), \vfodd(\z)\big) + \Cov\big(\A \vfodd(\Zall), \vfeven(\z)\big)
    = \A \Ko(\Zall, \z) + \mathbf{0}.
\end{align*}
For the second, analogously,
\begin{align*}
    \A \K(\Zall, \Zall) \A\trans = \Cov\big(\A \vfv, \A \vfv\big) = \Cov\big(\A \vfodd(\Zall), \A \vfodd(\Zall)\big) = \A \Ko(\Zall, \Zall) \A\trans.
\end{align*}
\end{proof}
Hence, the SV prior covariances need not be computed specifically for the odd kernel but coincide with those of the full kernel. Computing them for the odd kernel would require expanding it as the difference of two Hamming kernels, $\ko(\z, \z') = \tfrac{1}{2} ( \kf(\z, \z') - \kf(\z, \zbar') )$, and thus double their cost.
\subsection{Computational Consequences of the Odd GP} \label{app:odd_consequences}
By \cref{thm:odd_posterior}, the SV posterior of the odd GP has the same form as \cref{eq:sv_posterior_mean,eq:sv_posterior_cov} (see \cref{eq:odd_posterior_mean,eq:odd_posterior_cov} in \cref{app:posterior_proof}), but with the training kernel matrix $\Ko(\Xtodd, \Xtodd) \in \Rmat{\nodd}{\nodd}$, the targets $\Ytodd \in \Rp{\nodd}$, and the cross-covariance $\A \Ko(\Zall, \Xtodd) \in \Rmat{\np}{\nodd}$ in place of their counterparts with $t$ rows. Since $\nodd = t/2$, this reduces the cost of the Cholesky decomposition, which scales cubically in the number of rows, by a factor of up to $8$, that of the triangular solves, which scale quadratically, by a factor of $4$, and that of the prior cross-covariance computation, which scales linearly, by a factor of $2$. By \cref{thm:odd_eig}, the same reduction applies to all EIG quantities involving the training data (\cref{eq:eig_closed_form,eq:Q_def}). In addition, since the odd EIG is invariant under complements, only one representative per candidate pair has to be scored, which halves the number of scored candidates to $\ncand = \ncands/2$ and hence the size of all matrices over the candidate set as well as the cost of all operations on them, e.g., of the half-solve $\Y_{\cands} = \mathbf{L}^{-1} \K(\X, \cands)$ (\cref{app:shapleig}). Finally, each iteration acquires two evaluations, which halves the number of iterations for a given budget. See \cref{app:alg_costs} for a more detailed comparison.

\clearpage
\section{Proofs and Details for \cref{sec:quad}} \label{app:quad_proofs}
Throughout this section, the hyperparameters $\hp$ of the Hamming kernel GP surrogate are fixed, and we write $\alpha_j = 1$ and $\beta_j = \exp(-1/\ls_j)$ for the match and mismatch factors of the Hamming kernel (\cref{eq:hamming_kernel_full}). The quadrature route was introduced for explaining tree-based models by TreeGrad~\citep{LYL26b}, extended to Shapley interactions by Quadrature-TreeSHAP~\citep{WMY26}, and to computing the SVs of general product games, including product kernel methods, by QuadraSHAP~\citep{MRSMC26}.
\subsection{Shapley Weights as Beta Integrals} \label{app:quad_beta}
\begin{lemma}[Beta representation of the Shapley weights; see, e.g., \citealp{AAR99,O72}] \label{lem:beta}
For $s \in \{0, \dots, \np - 1\}$,
\begin{align*}
    \frac{1}{\np \binom{\np-1}{s}} = B(s+1, \np-s) = \int_0^1 \iv^{s} (1-\iv)^{\np-1-s} \, d\iv.
\end{align*}
\end{lemma}
Here, $B(\cdot, \cdot)$ is the Beta function~\citep{AAR99}. Its use for the Shapley weights goes back to the multilinear extension of \citet{O72}.

For every player $i \in \players$ and every coalition $S \subseteq \players$, the entry $\A_{i,S}$ of the Shapley transformation (\cref{eq:shapley_set_def}) is, up to sign, such a weight with $s = |S| - 1$ if $i \in S$ and $s = |S|$ otherwise,
\begin{align*}
    \A_{i,S} = \begin{cases}
        \dfrac{1}{\np \binom{\np-1}{|S|-1}} = \displaystyle\int_0^1 \iv^{|S|-1} (1-\iv)^{\np-|S|} \, d\iv, & i \in S, \\[12pt]
        -\dfrac{1}{\np \binom{\np-1}{|S|}} = -\displaystyle\int_0^1 \iv^{|S|} (1-\iv)^{\np-1-|S|} \, d\iv, & i \notin S.
    \end{cases}
\end{align*}
Both integrands have the same shape $\iv^{a} (1-\iv)^{b}$ with $a + b = \np - 1$. They differ only in which exponent is reduced by one relative to $|S|$ and $\np - |S|$: that of $\iv$ if $i \in S$ and that of $1 - \iv$ if $i \notin S$, i.e., the factor that player $i$ itself would contribute is left out. The following lemma expresses both cases through a single integrand that is a product over the remaining players.
\begin{lemma}[Leave-one-out product form; \citealp{O72}] \label{lem:loo_weights}
For a player $i \in \players$, a coalition $S \subseteq \players$, and $\iv \in [0,1]$, define the per-player factor $w_{\iv}(r, S) := \iv \Ind{r \in S} + (1-\iv) \Ind{r \notin S}$. Then
\begin{align*}
    \int_0^1 \prod_{r \neq i} w_{\iv}(r, S) \, d\iv
    = \begin{cases}
        \dfrac{1}{\np \binom{\np-1}{|S|-1}}, & i \in S, \\[8pt]
        \dfrac{1}{\np \binom{\np-1}{|S|}}, & i \notin S,
    \end{cases}
\end{align*}
and consequently the entry of $\A$ for player $i$ and coalition $S$ can be written as
\begin{align} \label{eq:A_entry_integral}
    \A_{i,S} = \big( \Ind{i \in S} - \Ind{i \notin S} \big) \int_0^1 \prod_{r \neq i} w_{\iv}(r, S) \, d\iv.
\end{align}
\end{lemma}
\subsection{The Hamming Kernel as a Product Game} \label{app:quad_product}
By \cref{eq:hamming_kernel}, the Hamming kernel is a product kernel, $\kf(\z, \z') = \oscale \prod_{j \in \players} ( \alpha_j \Ind{z_j = z'_j} + \beta_j \Ind{z_j \neq z'_j} )$. Fix a coalition $\z \in \binp$ and consider the kernel game $\vf_{\z}(S) = \kf(\indvec{S}, \z)$ from \cref{sec:quad}, i.e., the kernel value between $\z$ and the indicator vector $\indvec{S}$ of a coalition $S \subseteq \players$. Player $j$ contributes $\alpha_j$ if $(\indvec{S})_j = z_j$ and $\beta_j$ otherwise. Defining
\begin{align*}
    \gamma_j(\z) := \begin{cases} \alpha_j, & z_j = 0, \\ \beta_j, & z_j = 1, \end{cases}
    \qquad
    \delta_j(\z) := \begin{cases} \beta_j, & z_j = 0, \\ \alpha_j, & z_j = 1, \end{cases}
\end{align*}
i.e., the factor of player $j$ if $j \notin S$ and if $j \in S$, respectively, as in \citet{RFMBF26}, we obtain
\begin{align} \label{eq:hamming_product_game_gd}
    \vf_{\z}(S) = \oscale \prod_{j \notin S} \gamma_j(\z) \prod_{j \in S} \delta_j(\z)
    = \oscale \prod_{j \notin S} \beta_j^{z_j} \prod_{j \in S} \beta_j^{1-z_j},
\end{align}
where we used that $\alpha_j = 1$ implies $\gamma_j(\z) = \beta_j^{z_j}$ and $\delta_j(\z) = \beta_j^{1-z_j}$. Now consider the factor $\prod_{j \in S} \beta_j^{z_j} \beta_j^{-z_j} = 1$. Multiplying \cref{eq:hamming_product_game_gd} by it and absorbing the factors $\beta_j^{z_j}$ into the product over $j \notin S$ and the factors $\beta_j^{-z_j}$ into the product over $j \in S$ yields
\begin{align} \label{eq:hamming_product_game}
    \vf_{\z}(S)
    = \oscale \prod_{j \notin S} \beta_j^{z_j} \prod_{j \in S} \beta_j^{1-z_j} \prod_{j \in S} \beta_j^{z_j} \beta_j^{-z_j}
    = \oscale \Big( \prod_{j \in \players} \beta_j^{z_j} \Big) \prod_{j \in S} \beta_j^{1-2z_j}.
\end{align}
The first factor equals $\kf(\boldsymbol{0}, \z)$, the kernel value between $\z$ and the empty coalition. Hence,
\begin{align*}
    \vf_{\z}(S) = \kf(\boldsymbol{0}, \z) \prod_{j \in S} \beta_j^{1-2z_j}
\end{align*}
is the product of a constant that is independent of the argument $S$ and a product game in the sense of QuadraSHAP~\citep{MRSMC26}, i.e., a game of the form $\vf(S) = \prod_{j \in S} u_j$ with player-specific factors $u_j \in \R$, here with $u_j = \beta_j^{1-2z_j}$. Since the SVs are linear in the value function, the results for product games~\citep{MRSMC26} also apply after rescaling by the constant $\kf(\boldsymbol{0}, \z)$, i.e., also to $\vf_{\z}$ (see \cref{app:quad_linear}).
\subsection{The Linear Term} \label{app:quad_linear}
\Cref{eq:quad_linear} is the integral representation of the SVs of a product game~\citep{LYL26b,WMY26,MRSMC26} applied to the kernel game $\vf_{\z}$ of \cref{app:quad_product}, whose SVs are the entries of the SV prior covariance $\A \K(\Zall, \z)$, i.e.,
\begin{align} \tag{\ref{eq:quad_linear}}
    \big( \A \K(\Zall, \z) \big)_i = \phi_i(\vf_{\z}) = \kf(\boldsymbol{0}, \z) \, \big( \beta_i^{1-2z_i} - 1 \big) \int_0^1 \prod_{j \neq i} \big( (1-\iv) + \iv \, \beta_j^{1-2z_j} \big) \, d\iv .
\end{align}
For completeness, we derive the entries of $\A \K(\Zall, \z)$ via \cref{lem:loo_weights} (\cref{app:quad_beta}) directly from the Hamming kernel in its factorized form with the factors $\gamma_j(\z)$ and $\delta_j(\z)$ of \cref{app:quad_product}, i.e., without the rescaling to a product game, and show that the result coincides with \cref{eq:quad_linear}.
\begin{proof}
The $i$-th entry of $\A \K(\Zall, \z)$ is the SV of player $i$ for the game $\vf_{\z}(S) = \kf(\indvec{S}, \z)$, i.e., by \cref{eq:A_entry_integral} from \cref{lem:loo_weights} and \cref{eq:hamming_product_game_gd} from \cref{app:quad_product},
\begin{align*}
    \big( \A \K(\Zall, \z) \big)_i
    &= \sum_{S \subseteq \players} \A_{i,S} \, \kf(\indvec{S}, \z) \\
    &= \oscale \int_0^1 \sum_{S \subseteq \players} \big( \Ind{i \in S} - \Ind{i \notin S} \big) \prod_{r \neq i} w_{\iv}(r, S) \prod_{j \notin S} \gamma_j \prod_{j \in S} \delta_j \, d\iv ,
\end{align*}
where $\gamma_j = \gamma_j(\z)$ and $\delta_j = \delta_j(\z)$, and where we exchanged the finite sum with the integral. For fixed $\iv$, consider the integrand: each term of the sum is a product of per-player factors that depend only on the membership of the respective player, so that the sum over all subsets factorizes into a product over players of the two membership cases. In particular, player $i$ contributes $\delta_i$ if $i \in S$ and $\gamma_i$ if $i \notin S$, weighted by the signs $+1$ and $-1$, and does not appear in $\prod_{r \neq i} w_{\iv}(r, S)$; every other player $j \neq i$ contributes $\iv \, \delta_j$ if $j \in S$ and $(1-\iv) \, \gamma_j$ if $j \notin S$. This yields
\begin{align} \label{eq:quad_linear_gd}
    \big( \A \K(\Zall, \z) \big)_i
    = \oscale \, \big( \delta_i - \gamma_i \big) \int_0^1 \prod_{j \neq i} \big( \gamma_j (1-\iv) + \delta_j \, \iv \big) \, d\iv .
\end{align}
Finally, we bring \cref{eq:quad_linear_gd} into the form of \cref{eq:quad_linear}. Multiplying by $\prod_{j \in \players} \gamma_j / \gamma_j = 1$, where the factor $1 / \gamma_j$ is assigned to the respective player $j$, and identifying $\oscale \prod_{j \in \players} \gamma_j = \kf(\boldsymbol{0}, \z)$ and $\delta_j / \gamma_j = \beta_j^{1-2z_j}$ (\cref{app:quad_product}) yields
\begin{align*}
    &\oscale \, \big( \delta_i - \gamma_i \big) \int_0^1 \prod_{j \neq i} \big( \gamma_j (1-\iv) + \delta_j \, \iv \big) \, d\iv \\
    &\quad = \oscale \Big( \prod_{j \in \players} \gamma_j \Big) \frac{\delta_i - \gamma_i}{\gamma_i} \int_0^1 \prod_{j \neq i} \frac{\gamma_j (1-\iv) + \delta_j \, \iv}{\gamma_j} \, d\iv \\
    &\quad = \kf(\boldsymbol{0}, \z) \, \big( \beta_i^{1-2z_i} - 1 \big) \int_0^1 \prod_{j \neq i} \big( (1-\iv) + \iv \, \beta_j^{1-2z_j} \big) \, d\iv ,
\end{align*}
which is \cref{eq:quad_linear}.
\end{proof}
\subsection{The Quadratic Term} \label{app:quad_quadratic}
\Cref{thm:quad_quadratic} states that, for the Hamming kernel with $\alpha_r = 1$, for $r \in \players$ and $\iv, \ivb \in [0,1]$ with $d_r := 1 - \beta_r$ and $h_r(\iv,\ivb) := 1 - d_r (\iv + \ivb - 2\iv\ivb)$, and for all players $i \neq j$,
\begin{align}
    \big( \A \K(\Zall, \Zall) \A\trans \big)_{i,j}
    &= \oscale \, d_i \, d_j \iint_{[0,1]^2} (1-2\iv)(1-2\ivb) \prod_{r \notin \{i,j\}} h_r(\iv,\ivb) \, d\iv \, d\ivb, \tag{\ref{eq:quad_offdiag}} \\
    \big( \A \K(\Zall, \Zall) \A\trans \big)_{i,i}
    &= 2 \, \oscale \, d_i \iint_{[0,1]^2} \prod_{r \neq i} h_r(\iv,\ivb) \, d\iv \, d\ivb. \tag{\ref{eq:quad_diag}}
\end{align}
We prove the statement for the general product kernel of \cref{eq:hamming_kernel_full},
\begin{align*}
    \kf(\z, \z') = \oscale \prod_{r \in \players} \big( \alpha_r \Ind{z_r = z'_r} + \beta_r \Ind{z_r \neq z'_r} \big),
\end{align*}
with match and mismatch factors $\alpha_r$ and $\beta_r$ of player $r$, and specialize to the Hamming kernel with $\alpha_r = 1$ at the end of the proof.
\begin{proof}
We proceed in six steps: we write the entry as a double sum over coalition pairs (Step 1), represent both Shapley weights as integrals (\cref{lem:loo_weights} in \cref{app:quad_beta}; Step 2), exchange summation and integration (Step 3), collapse the sum over the players other than $i$ and $j$ into a product (Step 4), evaluate the contributions of $i$ and $j$ (Step 5), and specialize to the Hamming kernel (Step 6).

\emph{Step 1: The entry as a double sum.} The $(i,j)$ entry of $\A \K(\Zall, \Zall) \A\trans$ is the prior covariance between the SVs of players $i$ and $j$, which \shapleig{}~\citep{RFMBF26} writes as
\begin{align*}
    \big( \A \K(\Zall, \Zall) \A\trans \big)_{i,j} = \sum_{S, T \subseteq \players} \A_{i,S} \, \A_{j,T} \, \kf(\indvec{S}, \indvec{T}),
\end{align*}
where, by the product form of the kernel, player $r$ contributes $\alpha_r$ to $\kf(\indvec{S}, \indvec{T})$ if $r$ is in both or in neither of $S$ and $T$, and $\beta_r$ if $r$ is in exactly one of them.

\emph{Step 2: Both Shapley weights as integrals.} Let $\chi_i(S) := \Ind{i \in S} - \Ind{i \notin S}$ denote the sign of $\A_{i,S}$. By \cref{eq:A_entry_integral} from \cref{lem:loo_weights}, applied to $(i, S)$ with integration variable $\iv$ and to $(j, T)$ with integration variable $\ivb$,
\begin{align*}
    \A_{i,S} \, \A_{j,T}
    = \chi_i(S) \, \chi_j(T) \iint_{[0,1]^2} \prod_{r \neq i} w_{\iv}(r, S) \prod_{r \neq j} w_{\ivb}(r, T) \, d\iv \, d\ivb .
\end{align*}

\emph{Step 3: Exchanging summation and integration.} Substituting into Step 1 and exchanging the finite sum with the integrals,
\begin{align} \label{eq:quad_step3}
    \big( \A \K(\Zall, \Zall) \A\trans \big)_{i,j}
    = \oscale \iint_{[0,1]^2} \sum_{S, T \subseteq \players}
    \chi_i(S) \, \chi_j(T) \prod_{r \neq i} w_{\iv}(r, S) \prod_{r \neq j} w_{\ivb}(r, T) \, \tilde{k}(S, T) \, d\iv \, d\ivb,
\end{align}
where $\tilde{k}(S,T) := \kf(\indvec{S}, \indvec{T}) / \oscale$. In the following, we show that the sum factorizes into a product over players.

\emph{Step 4: The free players.} For fixed $(\iv,\ivb)$, consider the integrand of \cref{eq:quad_step3}: each summand is a product of per-player factors that depend only on the membership of the respective player in $S$ and $T$, namely $w_{\iv}(r,S)$, $w_{\ivb}(r,T)$, the kernel factor of $r$, and, for $r \in \{i,j\}$, the sign. Consequently, the sum over all coalition pairs $(S,T)$ factorizes into a product over players of sums over the four membership states $(\Ind{r \in S}, \Ind{r \in T}) \in \{0,1\}^2$.

Consider first a player $r \notin \{i, j\}$. Its membership in $S$ and $T$ is unconstrained, and it contributes $w_{\iv}(r,S)$, $w_{\ivb}(r,T)$, and its kernel factor. Summing over the four states,
\begin{align*}
    \underbrace{\iv \, \ivb \, \alpha_r}_{r \in S, \, r \in T}
    + \underbrace{\iv (1-\ivb) \, \beta_r}_{r \in S, \, r \notin T}
    + \underbrace{(1-\iv) \, \ivb \, \beta_r}_{r \notin S, \, r \in T}
    + \underbrace{(1-\iv)(1-\ivb) \, \alpha_r}_{r \notin S, \, r \notin T}
    =: h_r(\iv,\ivb).
\end{align*}
This factor is the same for all four cases of the memberships of $i$ and $j$, so every summand of \cref{eq:quad_step3} contains the common product $\prod_{r \notin \{i,j\}} h_r(\iv,\ivb)$.

\emph{Step 5a: The players $i$ and $j$ for $i \neq j$.} Player $i$ is excluded from $\prod_{r \neq i} w_{\iv}(r,S)$ but appears in $\prod_{r \neq j} w_{\ivb}(r,T)$ (as $i \neq j$) and in the kernel. For fixed membership of $i$ in $S$, summing over its membership in $T$ gives $\ivb \, \alpha_i + (1-\ivb) \, \beta_i$ if $i \in S$ (the two states $i \in T$ and $i \notin T$ contribute $\alpha_i$ and $\beta_i$) and $\ivb \, \beta_i + (1-\ivb) \, \alpha_i$ if $i \notin S$. Weighting by $\chi_i(S) = \pm 1$ and adding,
\begin{align*}
    \big( \ivb \, \alpha_i + (1-\ivb) \, \beta_i \big) - \big( \ivb \, \beta_i + (1-\ivb) \, \alpha_i \big) = (\beta_i - \alpha_i)(1 - 2\ivb).
\end{align*}
By symmetry, player $j$ contributes $(\beta_j - \alpha_j)(1 - 2\iv)$. Multiplying both with the common product of Step 4 and integrating, \cref{eq:quad_step3} becomes
\begin{align} \label{eq:quad_offdiag_general}
    \big( \A \K(\Zall, \Zall) \A\trans \big)_{i,j}
    = \oscale \, (\beta_i - \alpha_i)(\beta_j - \alpha_j) \iint_{[0,1]^2} (1-2\iv)(1-2\ivb) \prod_{r \notin \{i,j\}} h_r(\iv,\ivb) \, d\iv \, d\ivb .
\end{align}

\emph{Step 5b: The diagonal $i = j$.} Now the free players are all $r \neq i$ and contribute $\prod_{r \neq i} h_r(\iv,\ivb)$ as in Step 4. Player $i$ is excluded from both weight products and contributes only its kernel factor, weighted by $\chi_i(S) \, \chi_i(T)$:
\begin{align*}
    \underbrace{\alpha_i}_{i \in S, \, i \in T}
    \underbrace{- \, \beta_i}_{i \in S, \, i \notin T}
    \underbrace{- \, \beta_i}_{i \notin S, \, i \in T}
    \underbrace{+ \, \alpha_i}_{i \notin S, \, i \notin T}
    = 2 (\alpha_i - \beta_i).
\end{align*}
Multiplying with the common product and integrating, \cref{eq:quad_step3} becomes
\begin{align} \label{eq:quad_diag_general}
    \big( \A \K(\Zall, \Zall) \A\trans \big)_{i,i}
    = 2 \, \oscale \, (\alpha_i - \beta_i) \iint_{[0,1]^2} \prod_{r \neq i} h_r(\iv,\ivb) \, d\iv \, d\ivb .
\end{align}

\emph{Step 6: Specialization to the Hamming kernel.} For the Hamming kernel, $\alpha_r = 1$ for all $r \in \players$ (\cref{eq:hamming_kernel}). Using
\begin{align*}
    (1-\iv)(1-\ivb) + \iv \, \ivb &= 1 - (\iv + \ivb - 2\iv\ivb), \\
    \iv (1-\ivb) + (1-\iv) \, \ivb &= \iv + \ivb - 2\iv\ivb,
\end{align*}
the factor of Step 4 simplifies to
\begin{align*}
    h_r(\iv,\ivb) &= (1-\iv)(1-\ivb) + \iv \, \ivb + \beta_r \big( \iv (1-\ivb) + (1-\iv) \, \ivb \big) \\
    &= 1 - d_r (\iv + \ivb - 2\iv\ivb),
\end{align*}
and the prefactors in \cref{eq:quad_offdiag_general,eq:quad_diag_general} become
\begin{align*}
    (\beta_i - 1)(\beta_j - 1) = d_i \, d_j \qquad \text{and} \qquad 2 (1 - \beta_i) = 2 \, d_i.
\end{align*}
This gives \cref{eq:quad_offdiag,eq:quad_diag}.
\end{proof}
\subsection{Exact Gauss--Legendre Quadrature} \label{app:quad_gl}
\Cref{cor:quad_gl} states that, for the GL nodes and weights $(\tau_q, \omega_q)_{q=1}^{\nq}$ on $[0,1]$, replacing the integral in \cref{eq:quad_linear} by $\sum_{q} \omega_q \, (\cdot)(\tau_q)$ and the integrals in \cref{eq:quad_offdiag,eq:quad_diag} by $\sum_{q, q'} \omega_q \omega_{q'} \, (\cdot)(\tau_q, \tau_{q'})$ costs $\bigO(\np \nq)$ for all $\np$ entries of $\A \K(\Zall, \z)$ per coalition $\z$ and $\bigO(\np^2 \nq^2)$ for all $\np^2$ entries of $\A \K(\Zall, \Zall) \A\trans$, and is exact for $\nq \geq \lceil \np/2 \rceil$, i.e., at costs $\bigO(\np^2)$ and $\bigO(\np^4)$.
\begin{proof}
The integrands of \cref{eq:quad_linear,eq:quad_offdiag,eq:quad_diag} are polynomials of degree $\np - 1$ in each integration variable, and the GL rule with $\nq$ nodes integrates polynomials of degree up to $2\nq - 1$ exactly~\citep{GW69}. For the bivariate integrals, this applies to the tensor-product rule $\sum_{q, q'} \omega_q \omega_{q'} \, (\cdot)(\tau_q, \tau_{q'})$ with the degree counted in each variable separately. Hence, the rules are exact if $2\nq - 1 \geq \np - 1$, i.e., if $\nq \geq \lceil \np/2 \rceil$. The cost for all $\np$ entries of $\A \K(\Zall, \z)$ is $\bigO(\np \nq)$ per coalition $\z$, and the cost for all $\np^2$ entries of $\A \K(\Zall, \Zall) \A\trans$ is $\bigO(\np^2 \nq^2)$, since the products over players for the different entries of $\A \K(\Zall, \z)$ and $\A \K(\Zall, \Zall) \A\trans$ are computed jointly via the leave-one-out and leave-two-out products of \cref{app:quad_impl}.
\end{proof}
For $\nq < \lceil \np/2 \rceil$, the rules are approximations. For the one-dimensional integral in \cref{eq:quad_linear}, the GL error decays geometrically in $\nq$~\citep{MRSMC26,T08}; for the bivariate integrals in \cref{eq:quad_offdiag,eq:quad_diag}, we examine the accuracy attained with a fixed $\nq$ empirically in \cref{sec:experiments}. %
\subsection{Implementation} \label{app:quad_impl}
In the following, we describe how the linear term $\A \K(\Zall, \z)$ and the quadratic term $\A \K(\Zall, \Zall) \A\trans$ are computed via GL quadrature, and how the costs of \cref{cor:quad_gl} arise.

\paragraph{Leave-one-out products for the linear term.}
Consider a single coalition $\z \in \binp$. We evaluate the linear term in the form of \cref{eq:quad_linear_gd} from the proof in \cref{app:quad_linear},
\begin{align*}
    \big( \A \K(\Zall, \z) \big)_i = \oscale \, \big( \delta_i - \gamma_i \big) \int_0^1 \prod_{j \neq i} \big( \gamma_j (1-\iv) + \delta_j \, \iv \big) \, d\iv,
\end{align*}
with $\gamma_j = \gamma_j(\z)$ and $\delta_j = \delta_j(\z)$ of \cref{app:quad_product}, rather than in the product game form of \cref{eq:quad_linear}. The reason is numerical: $\gamma_j, \delta_j \in \{\alpha_j, \beta_j\} \subset (0, 1]$, so that every factor of the integrand lies in $(0, 1]$ for $\iv \in [0,1]$, whereas \cref{eq:quad_linear} involves the factors $\beta_j^{1-2z_j} = 1/\beta_j > 1$ for $z_j = 1$, which can get large for small lengthscales, and the potentially small constant $\kf(\boldsymbol{0}, \z)$. Positive and bounded factors allow a stable evaluation of the products in log space.

The GL nodes and weights $(\tau_q, \omega_q)_{q=1}^{\nq}$ on $[0,1]$ are computed once~\citep{GW69}, and for the coalition $\z$, we form the $\np \times \nq$ table
\begin{align*}
    H_{j,q} := \gamma_j (1-\tau_q) + \delta_j \, \tau_q \in (0, 1],
\end{align*}
which holds the factor of player $j$ in the integrand at node $\tau_q$, in $\bigO(\np \nq)$. Its column sums in log space, $L_q := \sum_{j \in \players} \log H_{j,q}$ for $q = 1, \dots, \nq$, are the logarithms of the products over all players at each node and cost $\bigO(\np \nq)$. They are shared across all entries $i$: by the standard log-space trick, the leave-one-out product at node $\tau_q$ is obtained by subtraction, which gives the $\np \times \nq$ table
\begin{align*}
    E_{i,q} := \prod_{j \neq i} H_{j,q} = \exp\big( L_q - \log H_{i,q} \big)
\end{align*}
of leave-one-out products in $\bigO(\np \nq)$. The quadrature sums of \cref{cor:quad_gl} for all $\np$ entries then form a single matrix-vector product of $E$ with the weight vector $\boldsymbol{\omega} := (\omega_1, \dots, \omega_{\nq})\trans$,
\begin{align*}
    \big( \A \K(\Zall, \z) \big)_i = \oscale \, \big( \delta_i - \gamma_i \big) \sum_{q=1}^{\nq} \omega_q \, E_{i,q} = \oscale \, \big( \delta_i - \gamma_i \big) \, (E \boldsymbol{\omega})_i,
\end{align*}
in $\bigO(\np \nq)$. In total, the cost per coalition is $\bigO(\np \nq)$, as stated in \cref{cor:quad_gl}. To avoid underflow for large $\np$, the exponent $L_q - \log H_{i,q}$ is shifted by its maximum over the nodes $q$ for each entry $i$ before exponentiating, and the shift is undone after the weighted sum.

For a batch of $n$ coalitions, e.g., the training coalitions or the candidates, the tables $E$ of all coalitions are stacked into an $n \np \times \nq$ matrix, so that the quadrature sums of the whole batch form a single matrix-vector product with $\boldsymbol{\omega}$, in $\bigO(n \np \nq)$. The batch is processed in chunks of coalitions whose working set fits into the cache. All steps are vectorized across players, nodes, and coalitions: the tables are elementwise operations, and the column sums, the maxima for the shift, and the weighted sums are reductions along one axis, the players or the nodes, that are independent along the other axes. In particular, no step is sequential over the players.

\paragraph{Leave-two-out products for the quadratic term.}
For the quadratic term, we evaluate \cref{eq:quad_offdiag,eq:quad_diag} on the tensor-product grid of the GL rule. We flatten the grid into $G := \nq^2$ node pairs $g = (q, q')$ with nodes $(\tau_q, \tau_{q'})$ and product weights $\omega_q \omega_{q'}$, and form the $\np \times G$ table
\begin{align*}
    H_{r,g} := h_r(\tau_q, \tau_{q'}) \in (0, 1],
\end{align*}
which holds the factor of player $r$ in the integrand at node pair $g$, in $\bigO(\np \nq^2)$. Its column sums in log space, $L_g := \sum_{r \in \players} \log H_{r,g}$ for $g = 1, \dots, G$, cost $\bigO(\np \nq^2)$ and are shared across all entries $(i,j)$.

\emph{Off-diagonal entries.} The leave-two-out product at node pair $g$ factorizes into two leave-one-out terms,
\begin{align*}
    \prod_{r \notin \{i,j\}} H_{r,g} = \exp\big( L_g - \log H_{i,g} - \log H_{j,g} \big) = V_{i,g} \, V_{j,g},
    \qquad
    V_{r,g} := \exp\big( \tfrac{1}{2} L_g - \log H_{r,g} \big),
\end{align*}
which gives the $\np \times G$ table $V$ in $\bigO(\np \nq^2)$. With $d_i = 1 - \beta_i$ from \cref{thm:quad_quadratic} and the node-pair weights $\tilde{\omega}_g := \omega_q \omega_{q'} (1 - 2\tau_q)(1 - 2\tau_{q'})$, the quadrature sums of \cref{cor:quad_gl} for \cref{eq:quad_offdiag} read
\begin{align*}
    \big( \A \K(\Zall, \Zall) \A\trans \big)_{i,j} &= \oscale \, d_i \, d_j \sum_{g=1}^{G} \tilde{\omega}_g \, V_{i,g} \, V_{j,g}, \\
    \text{i.e.,} \qquad
    \A \K(\Zall, \Zall) \A\trans \big|_{i \neq j} &= \oscale \, \big( \mathbf{d} \mathbf{d}\trans \big) \odot \big( V \diag(\tilde{\boldsymbol{\omega}}) V\trans \big),
\end{align*}
where $\odot$ denotes the elementwise product and $\mathbf{d} := (d_1, \dots, d_{\np})\trans$. All $\np^2$ off-diagonal entries thus form a single matrix product of the $\np \times G$ matrix $V \diag(\tilde{\boldsymbol{\omega}})$ with the $G \times \np$ matrix $V\trans$, in $\bigO(\np^2 \nq^2)$. To avoid underflow, the exponent $\tfrac{1}{2} L_g - \log H_{r,g}$ is shifted by its maximum $m_g$ over the players $r$ for each node pair $g$ before exponentiating, so that all entries of $V$ lie in $(0, 1]$. Both factors $V_{i,g}$ and $V_{j,g}$ of the summand for node pair $g$ carry this shift. Since the shift depends on $g$, it is undone inside the summand by multiplying the weight $\tilde{\omega}_g$ by $\exp(2 m_g)$.

\emph{Diagonal entries.} The diagonal is computed separately from \cref{eq:quad_diag}. With the leave-one-out products
\begin{align*}
    E_{i,g} := \prod_{r \neq i} H_{r,g} = \exp\big( L_g - \log H_{i,g} \big),
\end{align*}
which give a $\np \times G$ table $E$ in $\bigO(\np \nq^2)$, and the product weights $\omega_q \omega_{q'}$, the quadrature sums of \cref{cor:quad_gl} for \cref{eq:quad_diag} read
\begin{align*}
    \big( \A \K(\Zall, \Zall) \A\trans \big)_{i,i} = 2 \, \oscale \, d_i \sum_{g=1}^{G} \omega_q \omega_{q'} \, E_{i,g},
\end{align*}
i.e., a matrix-vector product of $E$ with the vector of product weights, in $\bigO(\np \nq^2)$. As for the linear term, the exponent $L_g - \log H_{i,g}$ is shifted by its maximum over the node pairs $g$ for each entry $i$ before exponentiating, and the shift is undone after the weighted sum.

\emph{Memory, total cost, and parallelization.} The tables $H$, $V$, and $E$ are formed and consumed in chunks of node pairs, which bounds the memory by the chunk size. In total, the cost is $\bigO(\np \nq^2)$ for the tables and the matrix-vector product of the diagonal entries, and $\bigO(\np^2 \nq^2)$ for the matrix product of the off-diagonal entries. As for the linear term, all steps are vectorized across players and node pairs: the tables are elementwise operations, and the column sums, the maxima for the shifts, and the final entries are reductions along one axis, the players or the node pairs, that are independent along the other axes. Overall, the computation is dominated by a single dense matrix product in $\bigO(\np^2 \nq^2)$, as stated in \cref{cor:quad_gl}, which is parallel over the player pairs.
\subsection{Comparison with the ESP Route of \shapleig{}} \label{app:quad_vs_esp}
The quadrature route for computing the prior covariances $\A \K(\Zall, \z)$ and $\A \K(\Zall, \Zall) \A\trans$ folds the kernel factors into the integral representation of the Shapley weights and evaluates the resulting integrand at the quadrature nodes. Coefficient-based schemes, as used by \shapleig{}~\citep{RFMBF26}, in contrast, do not involve any integral: they compute the sums of kernel values over all coalitions, respectively coalition pairs, of each size as the coefficients of a univariate, respectively bivariate, generating polynomial, using ESPs, and combine these coefficients with the size-dependent Shapley weights. The implementation of this route that we resort to in \cref{sec:experiments} avoids polynomial division, which is numerically unstable: as proposed by \citet{MCM25} for the linear term and by \shapleig{} for the quadratic term, it contracts the Shapley weights into suffix and, for the quadratic term, also prefix coefficient tables that are evolved over the players, once per coalition for the linear term and once per player for the quadratic term. This computes the two terms in $\bigO(\np^2)$ per coalition and $\bigO(\np^4)$, respectively, but the recursions are inherently sequential over the players, respectively player pairs.

In summary, the two routes differ as follows. At the exact node count $\nq = \lceil \np/2 \rceil$, both have the same asymptotic cost. However, the quadrature route is vectorized across players, nodes, and coalitions and reduces to matrix products (\cref{app:quad_impl}), whereas the ESP recursions are sequential over the players and player pairs. Moreover, the ESP route has no natural approximation, since all coefficients are required exactly, whereas the quadrature route can be approximated with fewer nodes.

\clearpage
\section{Details for \cref{alg:shapleyog}} \label{app:algorithm}

\subsection{Comparison with \shapleig{}} \label{app:alg_comparison}
\Cref{tab:alg_comparison} summarizes the algorithmic choices of \methodname{} (\cref{alg:shapleyog}) and contrasts them with \shapleig{} (Algorithm~1 in \citealp{RFMBF26}) along all aspects in which the two algorithms differ.
\begin{table}[h]
\centering \small
\caption{Algorithmic differences between \shapleig{} and \methodname{}.}
\label{tab:alg_comparison}
\begin{tabular}{@{}p{0.19\linewidth}p{0.37\linewidth}p{0.38\linewidth}@{}}
\toprule
 & \shapleig{} & \methodname{} \\
\midrule
Surrogate & Full GP with constant mean on all $t$ observations & Odd GP with zero mean on the $\nodd = t/2$ odd targets (\cref{sec:oddgp}) \\
\addlinespace
Hyperparameters & MAP estimate on all $t$ observations; refit in every iteration (staged schedule for large $\np$) & MAP estimate on the $\nodd$ odd targets; refit at $t = \Tinit$, whenever $t$ is a power of two, and on the final data (\cref{sec:algorithm}) \\
\addlinespace
SV prior covariances & Computation via ESPs (\cref{sec:shapleig}) & Computation via GL quadrature; exact for $\nq \geq \lceil \np/2 \rceil$ (\cref{sec:quad}) \\
\addlinespace
Initial design & Arbitrary & Paired \\
\addlinespace
Candidate set & Fixed set of size $\ncands$ provided as input; depleted over the run & Paired set of size $\ncands$ with $\ncand = \ncands/2$ scored representatives, resampled at every refit (\cref{sec:algorithm,app:exp_shapleyog}) \\
\addlinespace
Selection & One coalition per iteration; EIG of a single evaluation & One complement pair per iteration; joint EIG of both evaluations, scored via one representative (\cref{thm:odd_eig}) \\
\addlinespace
Between refits & Reuse of the kernel matrices and SV prior covariances, which change by single rows or columns; factorization and all solves recomputed & The same reuse; in addition, bordered Cholesky factorization, extension of the half-solves, and rank-one updates of the posterior covariances (\cref{app:alg_costs,app:alg_updates}) \\
\addlinespace
Iterations for budget $\budget$ & $\budget - \Tinit$ & $(\budget - \Tinit)/2$ \\
\bottomrule
\end{tabular}
\end{table}
\subsection{Computation Scheme of \methodname{}: Quantities, Sizes, and Costs} \label{app:alg_costs}
We now present the computation scheme of \methodname{} in detail and contrast it with that of \shapleig{}. For \shapleig{}, \cref{app:shapleig} describes the computation of the EIG over the candidate set in one iteration in five steps. The SV posterior can be computed in addition, e.g., in the final iteration. We keep this structure here. \Cref{tab:alg_costs} lists, in the order of the five steps, every quantity involved in an iteration together with its size and its cost when computed from scratch, i.e., at a refit, for \shapleig{} and \methodname{}. \Cref{tab:alg_costs_update} lists, for the same quantities, the cost of updating them by one observation in an iteration without refit. The update techniques of \methodname{} that appear there, i.e., the bordered Cholesky factorization, the extension of the half-solves, and the rank-one updates of the posterior covariances, are introduced in \cref{app:alg_updates}, together with how they apply to the respective quantities. The last rows of both tables give the dominant terms per iteration.
\begin{table}[H]
\centering \small
\caption{Quantities computed per iteration for the EIG over the candidate set and for the SV posterior, in the order of the steps of \cref{app:shapleig}, with their sizes and their costs when computed from scratch, i.e., at a refit, for \shapleig{} and \methodname{}, where $\X$ and $\K$ stand for $\Xt$ and $\K$ for \shapleig{} and for $\Xtodd$ and $\Ko$ for \methodname{}; for GL quadrature with $\nq$ nodes, the cost of the exact case $\nq = \lceil \np/2 \rceil$ is given in brackets.}
\label{tab:alg_costs}
\footnotesize
\setlength{\tabcolsep}{2.5pt}
\begin{tabular}{@{}lllll@{}}
\toprule
& \multicolumn{2}{c}{\shapleig{}} & \multicolumn{2}{c@{}}{\methodname{}} \\
\cmidrule(lr){2-3} \cmidrule(l){4-5}
Quantity & Size & Cost & Size & Cost \\
\midrule
\multicolumn{5}{@{}l}{\emph{Step 1: kernel matrices}} \\
$\widetilde{\K} = \K(\X, \X) + \noise \I$ & $t \times t$ & $\bigO(t^2 \np)$ & $\nodd \times \nodd$ & $\bigO(\nodd^2 \np)$ \\
$\mathbf{L}$ (Cholesky factor of $\widetilde{\K}$) & $t \times t$ & $\bigO(t^3)$ & $\nodd \times \nodd$ & $\bigO(\nodd^3)$ \\
$\K(\X, \cands)$ & $t \times \ncands$ & $\bigO(t \ncands \np)$ & $\nodd \times \ncand$ & $\bigO(\nodd \ncand \np)$ \\
\addlinespace
\multicolumn{5}{@{}l}{\emph{Step 2: SV prior covariances}} \\
$\A \KZZ \A\trans$ & $\np \times \np$ & $\bigO(\np^4)$ & $\np \times \np$ & $\bigO(\np^2 \nq^2)$ [$\bigO(\np^4)$] \\
$\A \K(\Zall, \X)$ & $\np \times t$ & $\bigO(t \np^2)$ & $\np \times \nodd$ & $\bigO(\nodd \np \nq)$ [$\bigO(\nodd \np^2)$] \\
$\A \K(\Zall, \cands)$ & $\np \times \ncands$ & $\bigO(\ncands \np^2)$ & $\np \times \ncand$ & $\bigO(\ncand \np \nq)$ [$\bigO(\ncand \np^2)$] \\
\addlinespace
\multicolumn{5}{@{}l}{\emph{Step 3: half-solves}} \\
$\Y_{\X} = \mathbf{L}^{-1} (\A \K(\Zall, \X))\trans$ & $t \times \np$ & $\bigO(t^2 \np)$ & $\nodd \times \np$ & $\bigO(\nodd^2 \np)$ \\
$\Y_{\cands} = \mathbf{L}^{-1} \K(\X, \cands)$ & $t \times \ncands$ & $\bigO(t^2 \ncands)$ & $\nodd \times \ncand$ & $\bigO(\nodd^2 \ncand)$ \\
\addlinespace
\multicolumn{5}{@{}l}{\emph{Step 4: posterior covariances}} \\
$\mathbf{v}$ (predictive variances) & $\ncands$ & $\bigO(t \ncands)$ & $\ncand$ & $\bigO(\nodd \ncand)$ \\
$\mathbf{B}$ (cross-covariances) & $\np \times \ncands$ & $\bigO(t \np \ncands)$ & $\np \times \ncand$ & $\bigO(\nodd \np \ncand)$ \\
$\Sigsv$ (SV posterior covariance) & $\np \times \np$ & $\bigO(t \np^2)$ & $\np \times \np$ & $\bigO(\nodd \np^2)$ \\
\addlinespace
\multicolumn{5}{@{}l}{\emph{Step 5: EIG}} \\
$\mathbf{L}_{\sv}$ (Cholesky factor of $\Sigsv$) & $\np \times \np$ & $\bigO(\np^3)$ & $\np \times \np$ & $\bigO(\np^3)$ \\
$\mathbf{L}_{\sv}^{-1} \mathbf{B}$ & $\np \times \ncands$ & $\bigO(\np^2 \ncands)$ & $\np \times \ncand$ & $\bigO(\np^2 \ncand)$ \\
$\mathbf{q}$ and EIG & $\ncands$ & $\bigO(\np \ncands)$ & $\ncand$ & $\bigO(\np \ncand)$ \\
\addlinespace
\multicolumn{5}{@{}l}{\emph{SV posterior}} \\
$\mathbf{w} := \mathbf{L}^{-1} \Y$ & $t$ & $\bigO(t^2)$ & $\nodd$ & $\bigO(\nodd^2)$ \\
$\musv = \Y_{\X}\trans \mathbf{w}$ & $\np$ & $\bigO(t \np)$ & $\np$ & $\bigO(\nodd \np)$ \\
\addlinespace
\multicolumn{5}{@{}l}{\emph{Hyperparameter fit, per objective evaluation}} \\
MAP objective and gradient & & $\bigO(t^3 + t^2 \np)$ & & $\bigO(\nodd^3 + \nodd^2 \np)$ \\
\midrule
\textbf{Total per iteration, without fit} & & $\bigO(t^3 + t^2 \ncands + \np^4)$ & & $\bigO(\nodd^3 + \nodd^2 \ncand + \np^2 \nq^2)$ \\
\bottomrule
\end{tabular}
\end{table}
\begin{table}[H]
\centering \small
\caption{The same quantities as in \cref{tab:alg_costs}, with the cost of updating them by one observation in an iteration without refit; "same" denotes the same update as for \shapleig{}, with $t$ and $\ncands$ replaced by $\nodd$ and $\ncand$.}
\label{tab:alg_costs_update}
\footnotesize
\setlength{\tabcolsep}{4pt}
\begin{tabular}{@{}l>{\raggedright\arraybackslash}p{0.28\linewidth}>{\raggedright\arraybackslash}p{0.37\linewidth}@{}}
\toprule
Quantity & \shapleig{} & \methodname{} \\
\midrule
\multicolumn{3}{@{}l}{\emph{Step 1: kernel matrices}} \\
$\widetilde{\K} = \K(\X, \X) + \noise \I$ & one row and column appended, $\bigO(t \np)$ & same \\
$\mathbf{L}$ (Cholesky factor of $\widetilde{\K}$) & recomputed, $\bigO(t^3)$ & one row appended (bordered), $\bigO(\nodd^2)$ \\
$\K(\X, \cands)$ & one row appended, $\bigO(\ncands \np)$; one column removed & same \\
\addlinespace
\multicolumn{3}{@{}l}{\emph{Step 2: SV prior covariances}} \\
$\A \KZZ \A\trans$ & reused & same \\
$\A \K(\Zall, \X)$ & one column appended, $\bigO(\np^2)$ & one column appended, $\bigO(\np \nq)$ \\
$\A \K(\Zall, \cands)$ & one column removed & same \\
\addlinespace
\multicolumn{3}{@{}l}{\emph{Step 3: half-solves}} \\
$\Y_{\X} = \mathbf{L}^{-1} (\A \K(\Zall, \X))\trans$ & recomputed, $\bigO(t^2 \np)$ & one row appended, $\bigO(\nodd \np)$ \\
$\Y_{\cands} = \mathbf{L}^{-1} \K(\X, \cands)$ & recomputed, $\bigO(t^2 \ncands)$ & one row appended, $\bigO(\nodd \ncand)$; one column removed \\
\addlinespace
\multicolumn{3}{@{}l}{\emph{Step 4: posterior covariances}} \\
$\mathbf{v}$ (predictive variances) & recomputed, $\bigO(t \ncands)$ & rank-one update of all entries, $\bigO(\ncand)$; one entry removed \\
$\mathbf{B}$ (cross-covariances) & recomputed, $\bigO(t \np \ncands)$ & rank-one update, $\bigO(\np \ncand)$; one column removed \\
$\Sigsv$ (SV posterior covariance) & recomputed, $\bigO(t \np^2)$ & rank-one update, $\bigO(\np^2)$ \\
\addlinespace
\multicolumn{3}{@{}l}{\emph{Step 5: EIG}} \\
$\mathbf{L}_{\sv}$ (Cholesky factor of $\Sigsv$) & recomputed (Cholesky), $\bigO(\np^3)$ & same \\
$\mathbf{L}_{\sv}^{-1} \mathbf{B}$ & recomputed, $\bigO(\np^2 \ncands)$ & same \\
$\mathbf{q}$ and EIG & recomputed, $\bigO(\np \ncands)$ & same \\
\addlinespace
\multicolumn{3}{@{}l}{\emph{SV posterior}} \\
$\mathbf{w} := \mathbf{L}^{-1} \Y$ and $\musv = \Y_{\X}\trans \mathbf{w}$ & not needed between refits & same \\
\midrule
\textbf{Total per iteration} & $\bigO(t^3 + t^2 \ncands + \np^2 \ncands)$ & $\bigO(\nodd^2 + \nodd \ncand + \np^3 + \np^2 \ncand)$ \\
\bottomrule
\end{tabular}
\end{table}
\subsection{Incremental Posterior Covariance Updates} \label{app:alg_updates}
In an iteration without refit, \methodname{} updates the posterior covariances (Step~4 in \cref{app:shapleig}), and thus the EIG, via a bordered Cholesky factorization, the extension of the half-solves, and rank-one updates. This goes beyond \shapleig{}, which only reuses the kernel matrices and the SV prior covariances and recomputes the remaining quantities. In the following, we show in detail how this is achieved.

Consider the iteration at $t$ without refit, i.e., with fixed hyperparameters $\hp$, and the quantities of \cref{tab:alg_costs} for the odd GP on $\Dtodd$, where now $\widetilde{\K} = \Ko(\Xtodd, \Xtodd) + \noise \I \in \Rmat{\nodd}{\nodd}$ denotes the training kernel matrix of the odd GP with its Cholesky factor $\mathbf{L}$, and $\Y_{\X} = \mathbf{L}^{-1} (\A \K(\Zall, \Xtodd))\trans \in \Rmat{\nodd}{\np}$ and $\Y_{\cands} = \mathbf{L}^{-1} \Ko(\Xtodd, \cands) \in \Rmat{\nodd}{\ncand}$ the half-solves. Evaluating the representative $\z^{+} := \z^{(t+1)}$ of the pair selected at iteration $t$ and its complement appends $\z^{+}$ together with its odd target to the odd dataset, which becomes $\D_{t+2,\odd}$ with one more row in $\Xtodd$, and the EIG of all remaining candidates has to be updated for the next iteration at $t+2$. %
\paragraph{Bordered Cholesky Factorization.}
Given the new observation $\z^{+}$, the training kernel matrix of $\D_{t+2,\odd}$ is
\begin{align} \label{eq:bordered_kernel}
    \widetilde{\K}^{+} = \begin{pmatrix} \widetilde{\K} & \mathbf{k} \\ \mathbf{k}\trans & \kappa \end{pmatrix},
    \qquad
    \mathbf{k} := \Ko(\Xtodd, \z^{+}) \in \Rp{\nodd}, \quad \kappa := \ko(\z^{+}, \z^{+}) + \noise,
\end{align}
where $\mathbf{k}$ is the kernel vector of $\z^{+}$ against the existing rows and $\kappa$ its diagonal entry. We again need a Cholesky factorization $\widetilde{\K}^{+} = \mathbf{L}^{+} (\mathbf{L}^{+})\trans$. Since the factor of the leading block is already available, only the last row of the new factor has to be computed (Cholesky factorization by bordering). To see this, consider a factor of the form
\begin{align} \label{eq:bordered_cholesky}
    \mathbf{L}^{+} := \begin{pmatrix} \mathbf{L} & \mathbf{0} \\ \mathbf{l}\trans & d \end{pmatrix}
    \qquad \text{with} \qquad
    \mathbf{L}^{+} (\mathbf{L}^{+})\trans = \begin{pmatrix} \mathbf{L} \mathbf{L}\trans & \mathbf{L} \mathbf{l} \\ \mathbf{l}\trans \mathbf{L}\trans & \mathbf{l}\trans \mathbf{l} + d^2 \end{pmatrix}
\end{align}
for some $\mathbf{l} \in \Rp{\nodd}$ and $d > 0$. Equating the blocks of $\mathbf{L}^{+} (\mathbf{L}^{+})\trans$ with those of $\widetilde{\K}^{+}$, the upper-left block holds by $\widetilde{\K} = \mathbf{L} \mathbf{L}\trans$, and the remaining blocks require $\mathbf{L} \mathbf{l} = \mathbf{k}$ and $\mathbf{l}\trans \mathbf{l} + d^2 = \kappa$, i.e.,
\begin{align} \label{eq:bordered_solution}
    \mathbf{l} := \mathbf{L}^{-1} \mathbf{k}
    \qquad \text{and} \qquad
    d := \sqrt{\kappa - \mathbf{l}\trans \mathbf{l}},
\end{align}
so that $\mathbf{L}^{+}$ is the Cholesky factor of $\widetilde{\K}^{+}$, which is unique as a lower triangular factor with positive diagonal~\citep[Thm.~4.2.7]{GV13}. Here, $d^2 = \kappa - \mathbf{k}\trans \widetilde{\K}^{-1} \mathbf{k}$ is the Schur complement of $\widetilde{\K}$ in $\widetilde{\K}^{+}$, i.e., the posterior predictive variance of the new observation plus $\noise$. Computing $\mathbf{l}$ requires one forward substitution with $\mathbf{L}$ in $\bigO(\nodd^2)$, and $d$ one inner product in $\bigO(\nodd)$. In other words, the previous Cholesky factor is recycled entirely, and only its new row is computed, in $\bigO(\nodd^2)$ instead of $\bigO(\nodd^3)$ for a factorization from scratch.
\paragraph{Extension of the Half-Solves.}
With $\mathbf{L}^{+}$ at hand, the half-solves of the next iteration are $\Y_{\X}^{+} = (\mathbf{L}^{+})^{-1} \mathbf{R}_{\X}^{+}$ and $\Y_{\cands}^{+} = (\mathbf{L}^{+})^{-1} \mathbf{R}_{\cands}^{+}$ with the right-hand sides $\mathbf{R}_{\X}^{+} := (\A \K(\Zall, \X_{t+2,\odd}))\trans$ and $\mathbf{R}_{\cands}^{+} := \Ko(\X_{t+2,\odd}, \cands^{+})$, where $\cands^{+}$ denotes the candidate set without the selected pair. Compared with the right-hand sides $\mathbf{R}_{\X} := (\A \K(\Zall, \Xtodd))\trans$ and $\mathbf{R}_{\cands} := \Ko(\Xtodd, \cands)$ of the current iteration, both gain one row. The new rows are $\boldsymbol{\rho}_{\X}\trans := (\A \K(\Zall, \z^{+}))\trans$, the new column of the SV prior covariance $\A \K(\Zall, \X_{t+2,\odd})$, which costs one GL evaluation, and $\boldsymbol{\rho}_{\cands}\trans := \Ko(\z^{+}, \cands^{+})$, the kernel vector of $\z^{+}$ against the remaining candidates. In addition, $\mathbf{R}_{\cands}$ loses the column of the selected representative, which merely removes the corresponding column of $\Y_{\cands}$. Hence, for either half-solve, we have to solve a triangular system whose matrix and right-hand side differ from the previously solved system only by one appended row. Forward substitution is row-sequential, i.e., the $i$-th row of $\mathbf{L}^{-1} \mathbf{R}$ depends only on the first $i$ rows of $\mathbf{L}$ and $\mathbf{R}$~\citep[Sec.~3.1.1]{GV13}, so that the previous rows remain valid. Explicitly, for a right-hand side $\mathbf{R}^{+} = (\mathbf{R}\trans, \boldsymbol{\rho})\trans$ with $m$ columns and the inverse of the block-triangular $\mathbf{L}^{+}$,
\begin{align} \label{eq:half_solve_extension}
    \Y^{+} := (\mathbf{L}^{+})^{-1} \mathbf{R}^{+}
    = \begin{pmatrix} \mathbf{L}^{-1} & \mathbf{0} \\ -\mathbf{l}\trans \mathbf{L}^{-1} / d & 1/d \end{pmatrix}
    \begin{pmatrix} \mathbf{R} \\ \boldsymbol{\rho}\trans \end{pmatrix}
    = \begin{pmatrix} \Y \\ \big( \boldsymbol{\rho}\trans - \mathbf{l}\trans \Y \big) / d \end{pmatrix}
    \qquad \text{with } \Y = \mathbf{L}^{-1} \mathbf{R}.
\end{align}
Thus, the previous half-solve $\Y$ is recycled entirely, and only the new row is computed, which requires one inner product of $\mathbf{l}$ with each column of $\Y$, in $\bigO(\nodd m)$ instead of $\bigO(\nodd^2 m)$ for a new triangular solve, where $m = \np$ for $\Y_{\X}$ and $m = \ncand$ for $\Y_{\cands}$.
\paragraph{Rank-One Updates of the Posterior Covariances.}
At this point, the Cholesky factor and both half-solves of the next iteration are available. The posterior covariances of Step~4 in \cref{app:shapleig}, i.e., $\Sigsv = \A \KZZ \A\trans - \Y_{\X}\trans \Y_{\X}$, $\mathbf{B} = \A \K(\Zall, \cands) - \Y_{\X}\trans \Y_{\cands}$, and $\mathbf{v} = \big( \ko(\z, \z) \big)_{\z \in \cands} - \diag(\Y_{\cands}\trans \Y_{\cands})$, each consist of a prior term, which is reused or merely loses the column or entry of the selected representative, minus a product of two half-solves. Writing the extended half-solves in terms of their previous rows and their new rows $\mathbf{y}_{\X}\trans \in \Rmat{1}{\np}$ and $\mathbf{y}_{\cands}\trans \in \Rmat{1}{\ncand}$ from \cref{eq:half_solve_extension}, $\Y_{\X}^{+} = (\Y_{\X}\trans, \mathbf{y}_{\X})\trans$ and $\Y_{\cands}^{+} = (\Y_{\cands}\trans, \mathbf{y}_{\cands})\trans$, block multiplication gives
\begin{align} \label{eq:gram_updates}
    (\Y_{\X}^{+})\trans \Y_{\X}^{+} &= \Y_{\X}\trans \Y_{\X} + \mathbf{y}_{\X} \mathbf{y}_{\X}\trans, \notag \\
    (\Y_{\X}^{+})\trans \Y_{\cands}^{+} &= \Y_{\X}\trans \Y_{\cands} + \mathbf{y}_{\X} \mathbf{y}_{\cands}\trans, \notag \\
    \diag\big( (\Y_{\cands}^{+})\trans \Y_{\cands}^{+} \big) &= \diag\big( \Y_{\cands}\trans \Y_{\cands} \big) + \mathbf{y}_{\cands} \circ \mathbf{y}_{\cands},
\end{align}
where $\circ$ denotes the elementwise product; each product changes by a rank-one term, since a product of this form is the sum of the outer products of the rows and exactly one row was added. Absorbing the previous products into the previous posterior covariances yields the updates
\begin{align} \label{eq:rank_one_covariances}
    \Sigsv^{+} &= \Sigsv - \mathbf{y}_{\X} \mathbf{y}_{\X}\trans, \notag \\
    \mathbf{B}^{+} &= \mathbf{B} - \mathbf{y}_{\X} \mathbf{y}_{\cands}\trans, \notag \\
    \mathbf{v}^{+} &= \mathbf{v} - \mathbf{y}_{\cands} \circ \mathbf{y}_{\cands},
\end{align}
where the column of $\mathbf{B}$ and the entry of $\mathbf{v}$ of the selected representative are removed. Hence, the posterior covariances need not be recomputed from the half-solves in $\bigO(\nodd \np^2)$, $\bigO(\nodd \np \ncand)$, and $\bigO(\nodd \ncand)$, but are updated in $\bigO(\np^2)$, $\bigO(\np \ncand)$, and $\bigO(\ncand)$. The EIG is then assembled from the updated quantities as in Step~5 of \cref{app:shapleig}, i.e., the Cholesky factor of $\Sigsv^{+}$, the triangular solve, and the scores are recomputed (\cref{tab:alg_costs_update}).
\paragraph{Numerical Behavior.}
All updates above are exact identities. The extensions of the Cholesky factor and of the half-solves perform exactly the operations that a computation from scratch performs for the new row and leave the existing entries untouched, so that rounding errors do not accumulate over the iterations; both operations are backward stable~\citep[Secs.~3.1.1 and 4.2.6]{GV13}. The rank-one updates compute the same sums as the products of the extended half-solves in a different summation order and thus carry the same type of rounding error bound~\citep[Sec.~2.7.8]{GV13}.
\clearpage
\section{Experimental Details} \label{app:exp}
In the following, we provide further details on the experimental setup of \cref{sec:experiments}.
\subsection{Games} \label{app:exp_games}

\begin{table}[h]
\centering \footnotesize
\setlength{\tabcolsep}{4pt}
\caption{Overview of the games considered in the experiments.}
\label{tab:dataset_summary}
\begin{tabular}{@{}ll>{\raggedright\arraybackslash}p{0.16\linewidth}r>{\raggedright\arraybackslash}p{0.19\linewidth}>{\raggedright\arraybackslash}p{0.15\linewidth}@{}}
\toprule
Dataset & Task & Model & $\np$ & Source & License \\
\midrule
\multicolumn{6}{@{}l}{\emph{Vision and language datasets} \citep{MBFKHH24}} \\
Language (IMDB) & Classification & DistilBERT & 14 & shapiq / HuggingFace & Public Domain \\
ResNet18-SP & Image classification & ResNet18 & 14 & shapiq & Public Domain \\
ViT4by4 & Image classification & ViT-B & 16 & shapiq & Public Domain \\
DINOv2-16 & Image classification & DINOv2-B + linear head & 16 & ImageNet-1k val.\ / HuggingFace & ImageNet Terms of Access \\
\addlinespace
\multicolumn{6}{@{}l}{\emph{Tabular datasets} \citep{SK10,MBFKHH24}} \\
Forest Fires & Regression & TabPFN-3 & 13 & UCI & CC-BY 4.0 \\
AdultCensus & Classification & TabPFN-3 & 14 & OpenML (1590) & CC-BY 4.0 \\
Estate & Regression & TabPFN-3 & 15 & UCI & CC-BY 4.0 \\
Crime & Regression & LightGBM & 101 & SHAP & CC-BY 4.0 \\
\addlinespace
\multicolumn{6}{@{}l}{\emph{TabArena datasets}~\citep{EPTHDSH25}} \\
Seismic Bumps & Classification & TabPFN-3 & 15 & OpenML (46956) & CC-BY 4.0 \\
Churn & Classification & TabPFN-3 & 19 & OpenML (46915) & MIT \\
Credit-G & Classification & TabPFN-3 & 20 & OpenML (46918) & CC-BY 4.0 \\
Bankruptcy & Classification & LightGBM & 64 & OpenML (46950) & CC-BY 4.0 \\
Superconductivity & Regression & LightGBM & 81 & OpenML (46961) & CC-BY 4.0 \\
NATICUSdroid & Classification & LightGBM & 86 & OpenML (46969) & CC-BY 4.0 \\
MIC & Classification & LightGBM & 111 & OpenML (46980) & CC-BY 4.0 \\
QSAR TID11 & Classification & LightGBM & 1024 & OpenML (46953) & Public Domain \\
\bottomrule
\end{tabular}
\end{table}
\Cref{tab:dataset_summary} gives an overview of the games. All of them are local explanation games, i.e., the value function explains the prediction of a machine learning model for a single instance. For all games, the seed selects the explained instance, whereas the explained models and, for the tabular games, the train-test splits are fixed. The table lists the dataset, the task of the model on this dataset, the explained model, the number of players $\np$, i.e., features, image patches, or tokens, as well as the source and the license of the data. Data sources are OpenML~\citep{BCplus25} with task IDs in parentheses, the UCI repository~\citep{KLN}, the SHAP library~\citep{LL17}, the \texttt{shapiq} benchmark suite~\citep{MBFKHH24}, and the ImageNet-1k validation set~\citep{DDSLLL09}. The games fall into two families: precomputed local explanation games of pretrained models, and local explanation games of tree ensembles that are evaluated on the fly.

The first family consists of local explanations of the image classifiers ResNet18~\citep{HZRS16}, ViT~\citep{DBplus21}, and DINOv2~\citep{Oquab.2024}, of the sentiment classifier DistilBERT~\citep{Sanh.2019}, and of predictions of the tabular foundation model TabPFN-3~\citep{grinsztajn_tabpfn-3_2026}. For the image classifiers, the players are image patches or superpixels, and the value of a coalition is the model output when only the patches in the coalition are retained. For DistilBERT, the players are the tokens of a movie review, and the value of a coalition is the model output when the remaining tokens are masked. For TabPFN, the players are the features of the explained instance. The value of a coalition is the prediction when the features outside the coalition are set to $+\infty$ and thereby treated as missing by TabPFN, so that every coalition costs a single forward pass.
The Forest Fires game deviates from this marginalization. It follows the remove-and-contextualize scheme~\citep{RKVFNR24}, in which the features outside the coalition are removed from both the context and the explained instance before the forward pass, instead of being marked as missing. The ResNet18, ViT, and DistilBERT games are taken from the \texttt{shapiq} benchmark suite~\citep{MBFKHH24}; the DINOv2 and TabPFN games were precomputed with \texttt{shapiq}. The ImageNet-1k validation images underlying the DINOv2 game are governed by the ImageNet Terms of Access, which permit non-commercial research and educational use only, whereas the DINOv2 weights (\texttt{facebook/dinov2-base-imagenet1k-1-layer}) are released under Apache-2.0.

The second family consists of local explanations of LightGBM tree ensembles~\citep{KMplus17} trained on tabular datasets, with the features of a single test instance as players. These games are evaluated on the fly, and their exact SVs are obtained via interventional TreeSHAP~\citep{ZBK23}, which makes them suitable for benchmarking with many players. The tree game QSAR TID11 with $\np = 1024$ players is used only in the quadrature study of \cref{app:exp_ablations}.

Since the raw payoffs of the games differ in scale by several orders of magnitude, every game is normalized such that $\vf(\emptyset) = 0$ and $\vf(\players) = 1$, so that the SVs sum to one by efficiency and squared-error metrics are comparable across games. Game instances with $|\vf(\players) - \vf(\emptyset)|$ below a numerical tolerance are degenerate under this normalization and are skipped; the next seed is used instead.
\subsection{Configuration of \methodname{}} \label{app:exp_shapleyog}
We detail the configuration of \methodname{} beyond \cref{sec:exp_setup}. Our implementation builds on the publicly available implementation of \shapleig{}~\citep{RFMBF26}. \methodname{} uses the odd GP of \cref{sec:oddgp}, i.e., a zero-mean GP with the odd Hamming kernel, fitted to the odd targets, which are rescaled to unit variance. Its kernel hyperparameters, i.e., the lengthscales and the output scale, are learned by MAP estimation with the dimension-scaled default lengthscale prior of BoTorch~\citep{BKplus20}, $\ls_j / \np \sim \mathrm{LogNormal}(\sqrt{2} + \tfrac{1}{2} \log \np, \sqrt{3})$, and a lower bound of $10^{-6}$ on $\ls_j / \np$, where the division by $\np$ reflects the parametrization of the categorical kernel of BoTorch, which averages the mismatches over the players. The MAP estimate is computed with L-BFGS-B~\citep{BLNZ95} in the log-coordinates of the hyperparameters, with closed-form gradients~\citep{RW06}, warm starts at the hyperparameters of the previous refit, and up to five restarts in case of failed optimization runs. For numerical stability, we assume additive zero-mean Gaussian noise with fixed variance $10^{-6}$ in the standardized units, which yields quasi-noiseless GPs (\cref{app:gp}).
The hyperparameters are refit when the number of observations reaches a power of two and on the final data (\cref{sec:algorithm}). The SV prior covariances are computed by GL quadrature with the exact node count (\cref{sec:quad}). Between refits, the posterior covariances and the EIG are updated incrementally via the bordered Cholesky factorization, the extension of the half-solves, and the rank-one updates of \cref{app:alg_updates}. Hamming kernel matrices are formed via a single matrix product of the binary designs.

The border trick initial design of \cref{sec:exp_setup} is deterministic and paired by construction. At every refit, the candidate set of $10240$ coalitions ($4096$ in the ablation path of \cref{app:exp_ablations}) is drawn from the not yet evaluated coalitions with the coalition sampler of \texttt{shapiq}, parameterized to sample coalition sizes uniformly and to complete every draw by its complement; for the games with at most $16$ players, all unevaluated coalitions are used if fewer remain. The EIG is scored for one representative of every complement pair in the candidate set, and the selected pair is evaluated as a whole.
\subsection{Baselines} \label{app:exp_baselines}
All non-adaptive baselines are run in their implementations of \texttt{shapiq}~\citep{MBFKHH24} version 1.7.0 with the default settings of this version. Kernel SHAP uses its kernel weights without paired sampling. Leverage SHAP samples coalition pairs without replacement with the leverage-score size weights. Regression MSR uses an XGBoost~\citep{CG16} proxy model and paired sampling. OddSHAP, which is paired by construction, uses its default interaction factor $\eta = 10$. Permutation sampling and SVARM use their own sampling schemes. Every baseline receives exactly the budget at which the estimate is recorded. Since the baselines are not adaptive, their estimates at different budgets stem from independent runs rather than from one sequential run, so that the coalitions evaluated at a smaller budget are in general not a subset of those evaluated at a larger budget.
\subsection{Metrics} \label{app:exp_metrics}
\paragraph{Estimation Accuracy.} For an estimate $\hat{\sv}$ and the exact SVs $\sv$, the normalized mean squared error is
\begin{align*}
    \mathrm{NMSE} := \frac{\| \hat{\sv} - \sv \|_2^2}{\| \sv \|_2^2}.
\end{align*}
Since the NMSE often differs by orders of magnitude across games and seeds, we aggregate it by the geometric mean, i.e., the mean in log space, across seeds and, where results are pooled, across games. The per-game results of \cref{app:res_pergame} additionally report the arithmetic mean. The \emph{average rank} of an estimator at a fixed evaluation budget is obtained by ranking all competing estimators by NMSE on every game instance, i.e., (game, seed) combination, with rank $1$ for the lowest NMSE and tied values receiving the mean of their ranks, and averaging the rank over the instances.
\paragraph{Computational Overhead.} The overhead of an estimator is its wall-clock time excluding the value function evaluations. For \methodname{}, it comprises (i) the hyperparameter fitting, including the final refit at the recorded budget, (ii) the EIG computation from scratch at refit iterations, (iii) the EIG updates at all other iterations, (iv) the bookkeeping, i.e., the candidate resampling at refits and the data updates of the surrogate between refits, and (v) the computation of the SV posterior at the recorded budget. The value function evaluations are not included, since the runner evaluates them outside these steps. The baselines are run in \texttt{shapiq}, whose wall-clock time includes their value function evaluations, which is not negligible for the tree games that are evaluated on the fly. We therefore report as the overhead of a baseline its wall-clock time minus the smallest wall-clock time among all baselines on the same game, seed, and budget. This credits the cheapest baseline with zero overhead and every other baseline with at most its true overhead, so the correction is, if anything, in favour of the baselines. Overheads are aggregated across seeds by the mean or the median, as stated in the respective figure or table. The total computation time for an assumed cost $c$ per value function evaluation is the overhead plus $c$ times the budget. All reported overheads are wall-clock times on the hardware of \cref{app:exp_compute} with four CPU threads per run. Absolute times naturally depend on the hardware and the implementations, but the orders of magnitude and the relations between the methods are indicative of what is attained on a standard computer.
\subsection{Ablations} \label{app:exp_ablations}
\paragraph{Ablation Path from \shapleig{} to \methodname{}.}
The ablation path of \cref{sec:exp_setup} adds the measures of \methodname{} one at a time, starting from \shapleig{} (see \cref{app:shapleig,app:algorithm} for the main differences). The ablation path is run on all $15$ games of the main experiment with $10$ seeds, evaluation budgets up to $2048$, and a candidate set of $4096$ coalitions. The measures are added in the order of \cref{sec:exp_setup}. Paired acquisition is first added under the full GP with the joint EIG of the pair, before the odd GP with odd hyperparameter fitting is switched on, which yields \methodname{}. The \shapleig{} variant of this ablation path is not the published implementation but our implementation of its algorithm. All variants share the implementation of \cref{app:exp_shapleyog}, i.e., the border trick initial design, the hyperparameter optimizer in log-coordinates with closed-form gradients and warm starts, and the kernel assembly via a single matrix product, so that the ablation path isolates the algorithmic measures. The variants without quadrature compute both SV prior covariances via the ESP route with contracted prefix and suffix coefficient tables, as discussed in \cref{app:quad_vs_esp}. The linear term of the ESP route uses all available threads, whereas its quadratic term runs on a single thread, which we empirically found to be faster.
\paragraph{Quadrature Study.}
The GL quadrature of \cref{sec:quad} is exact for $\nq \geq \lceil \np/2 \rceil$ nodes; the quadrature study asks how many are needed in practice. On QSAR TID11 ($\np = 1024$), we run \methodname{} with $\nq \in \{514, 256, 64, 8, 2\}$ nodes for both SV prior covariances, where $514$ nodes are exact. The study uses $10$ seeds, $2050$ initial coalitions, evaluation budgets of $4096$ and $8192$, and a candidate set of $10240$ coalitions, and is otherwise configured as in \cref{app:exp_shapleyog}. At regular checkpoints, each run also recomputes its SV prior covariances and EIG scores with the exact quadrature on its own state, which yields the relative errors in \cref{tab:quadrature} and whether the exact scores select the same coalition. A separate run computes both SV prior covariances with the ESP route of \shapleig{} instead, i.e., via contracted prefix and suffix coefficient tables as discussed in \cref{app:quad_vs_esp}, and is otherwise identical to the exact arm, i.e., it keeps the odd GP, the refit schedule, and the incremental updates of \methodname{}. Since the bivariate ESP tables of the quadratic term require $34$\,GB at this $\np$ and every call takes hours, this run is restricted to one seed and the budget of $4096$ (\cref{app:res_quadrature}).
\subsection{Compute and Reproducibility} \label{app:exp_compute}
All experiments were run on CPU nodes with $32$ cores and $64$\,GB of RAM, using Python~3.12, Torch~2.9.1, GPyTorch~1.14, BoTorch~0.14.0, and \texttt{shapiq}~1.7.0. Every run uses four CPU threads (BLAS, OpenMP, and Torch), so all reported overheads are wall-clock times with four threads; the only exception is the ESP route of the quadratic term in the ablation path and in the full-ESP run of the quadrature study, which runs on a single thread (\cref{app:exp_ablations}). The code, including the configuration files of all experiments and the scripts for preparing the games, will be made publicly available, so that all results are reproducible.

\clearpage
\section{Additional Experimental Results} \label{app:results}
This appendix collects further results of the experiments of \cref{sec:experiments}. \Cref{app:res_pergame} shows the accuracy of every estimator per game in the main experiment (\cref{fig:facets_gmean,fig:facets_mean,tab:pergame_all,tab:pergame_all_mean}). \Cref{app:res_avgrank} complements the average ranks of \methodname{} in \cref{fig:main_results} with those of its strongest competitors (\cref{tab:avgrank_methods}). \Cref{app:res_overhead} breaks the overhead of \methodname{} down into its components for a representative subset of the games of the main experiment (\cref{tab:overhead_breakdown}). \Cref{app:res_ablation} presents the ablation path from \shapleig{} to \methodname{} for every game in isolation (\cref{fig:ladder_all,tab:ablation_speedup,tab:ablation_absolute}). \Cref{app:res_quadrature} presents the detailed results of the quadrature study (\cref{tab:quadrature}).
\subsection{Per-Game Accuracy} \label{app:res_pergame}
\begin{figure}[p]
  \centering
  \includegraphics[width=\textwidth]{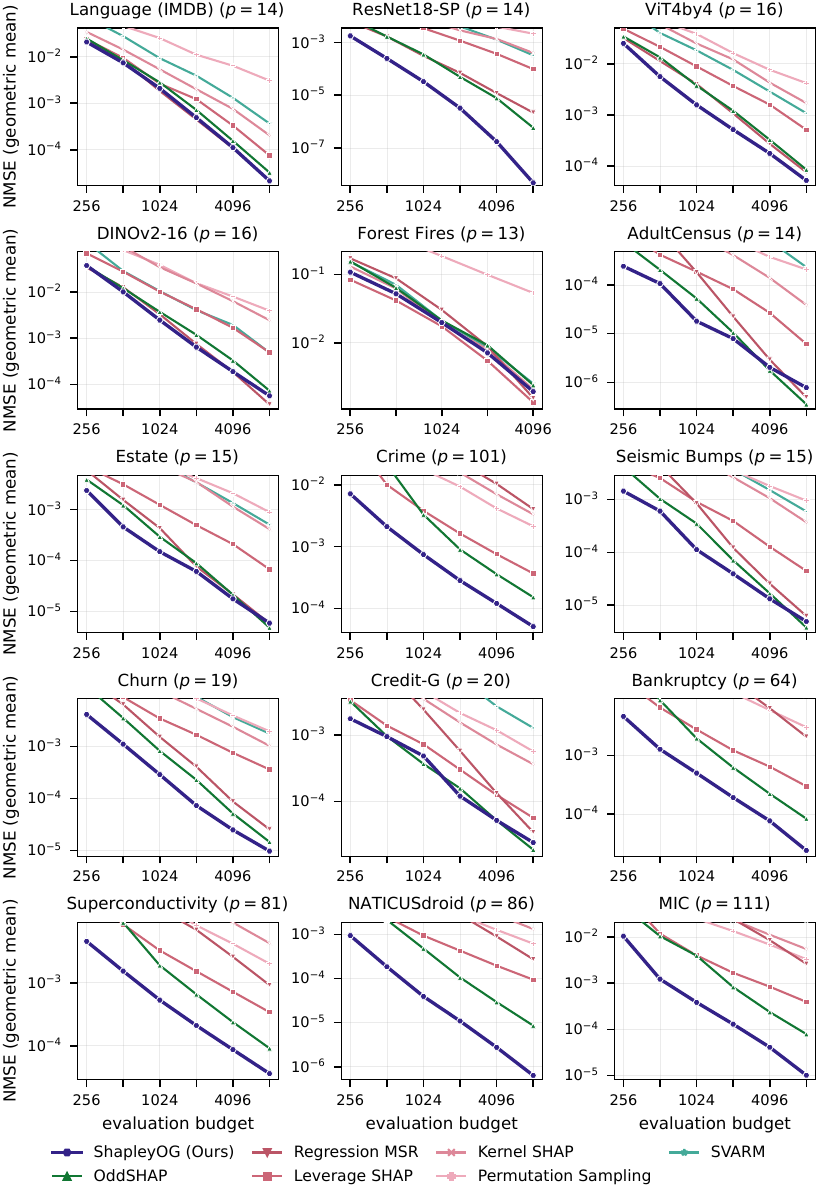}
  \caption{NMSE vs.\ evaluation budget per game, geometric mean over $30$ seeds. The y-axis starts roughly at the NMSE of \methodname{} at $256$ evaluations, so estimators starting above it enter the panel later.}
  \label{fig:facets_gmean}
\end{figure}

\begin{figure}[p]
  \centering
  \includegraphics[width=\textwidth]{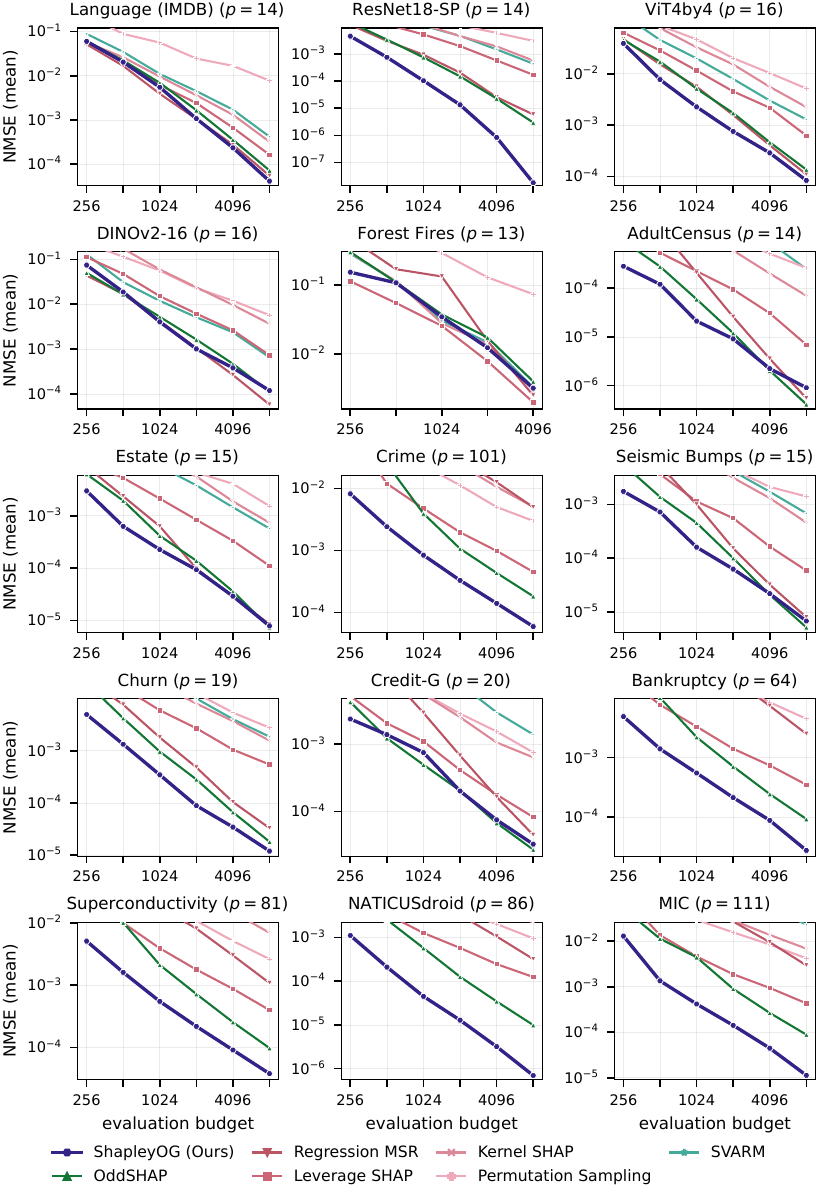}
  \caption{NMSE vs.\ evaluation budget per game, arithmetic mean over $30$ seeds; otherwise as \cref{fig:facets_gmean}.}
  \label{fig:facets_mean}
\end{figure}

In \cref{sec:exp_results}, the NMSE of the main experiment is shown pooled across games. Here, \cref{fig:facets_gmean,fig:facets_mean} plot the NMSE of all estimators for each of the $15$ games individually and at every recorded budget, aggregated over the $30$ seeds by the geometric mean and the arithmetic mean, respectively, and \cref{tab:pergame_all,tab:pergame_all_mean} list the corresponding numbers. Aggregating by the arithmetic instead of the geometric mean leads to the same overall picture.
\begin{table}[p]
  \caption{NMSE per game (geometric mean over $30$ seeds) at every recorded budget $\budget$; the best method per game and budget is in bold; -- marks budgets beyond a game's grid.}
  \label{tab:pergame_all}
  \centering \tiny \setlength{\tabcolsep}{4pt} \renewcommand{\arraystretch}{0.84}
  \begin{tabular}{@{}lrrrrrrr@{}}
\toprule
Game ($p$) & ShapleyOG & OddSHAP & Regression MSR & Leverage SHAP & Kernel SHAP & Permutation & SVARM \\
\midrule
\multicolumn{8}{@{}l}{\emph{$T = 256$}} \\
Language (IMDB) (14) & \textbf{2.1e-2} & 2.5e-2 & 2.2e-2 & 2.4e-2 & 3.4e-2 & 9.6e-2 & 7.1e-2 \\
ResNet18-SP (14) & \textbf{1.8e-3} & 8.2e-3 & 6.4e-3 & 2.1e-2 & 4.5e-2 & 5.8e-2 & 5.5e-2 \\
ViT4by4 (16) & \textbf{2.5e-2} & 3.4e-2 & 3.2e-2 & 4.8e-2 & 1.3e-1 & 1.6e-1 & 1.3e-1 \\
DINOv2-16 (16) & 3.7e-2 & 3.6e-2 & \textbf{3.4e-2} & 6.9e-2 & 2.1e-1 & 1.6e-1 & 1.2e-1 \\
Forest Fires (13) & 1.1e-1 & 1.5e-1 & 1.7e-1 & \textbf{8.3e-2} & 1.3e-1 & 6.8e-1 & 1.5e-1 \\
AdultCensus (14) & \textbf{2.4e-4} & 7.5e-4 & 5.5e-3 & 1.3e-3 & 5.9e-3 & 7.3e-3 & 3.8e-2 \\
Estate (15) & \textbf{2.4e-3} & 3.9e-3 & 5.9e-3 & 7.1e-3 & 4.0e-2 & 2.4e-2 & 6.0e-2 \\
Crime (101) & \textbf{7.1e-3} & 6.6e-2 & 1.3e-1 & 6.1e-2 & 2.0e-1 & 7.8e-2 & 5.4e0 \\
Seismic Bumps (15) & \textbf{1.4e-3} & 4.1e-3 & 1.8e-2 & 4.9e-3 & 4.1e-2 & 2.9e-2 & 6.3e-2 \\
Churn (19) & \textbf{4.1e-3} & 1.4e-2 & 2.0e-2 & 1.7e-2 & 5.9e-2 & 6.9e-2 & 1.3e-1 \\
Credit-G (20) & \textbf{1.8e-3} & 3.2e-3 & 2.3e-2 & 3.3e-3 & 1.8e-2 & 2.3e-2 & 1.0e-1 \\
Bankruptcy (64) & \textbf{4.6e-3} & 4.4e-2 & 1.6e-1 & 1.9e-2 & 5.0e-1 & 9.6e-2 & 1.1e0 \\
Superconductivity (81) & \textbf{4.6e-3} & 2.9e-2 & 5.3e-2 & 3.1e-2 & 2.1e-1 & 6.9e-2 & 3.3e0 \\
NATICUSdroid (86) & \textbf{9.4e-4} & 6.3e-3 & 3.5e-2 & 9.2e-3 & 7.4e-2 & 2.8e-2 & 3.1e0 \\
MIC (111) & \textbf{1.0e-2} & 4.6e-2 & 2.1e-1 & 8.7e-2 & 3.4e-1 & 1.1e-1 & 2.1e0 \\
\midrule
\multicolumn{8}{@{}l}{\emph{$T = 512$}} \\
Language (IMDB) (14) & \textbf{7.4e-3} & 8.4e-3 & 8.2e-3 & 9.4e-3 & 1.4e-2 & 4.6e-2 & 2.8e-2 \\
ResNet18-SP (14) & \textbf{2.5e-4} & 1.7e-3 & 1.7e-3 & 8.1e-3 & 2.1e-2 & 2.7e-2 & 2.1e-2 \\
ViT4by4 (16) & \textbf{5.6e-3} & 1.3e-2 & 1.1e-2 & 2.1e-2 & 5.9e-2 & 7.0e-2 & 4.0e-2 \\
DINOv2-16 (16) & \textbf{1.0e-2} & 1.2e-2 & 1.3e-2 & 2.8e-2 & 9.6e-2 & 8.1e-2 & 2.8e-2 \\
Forest Fires (13) & 5.2e-2 & 6.3e-2 & 8.7e-2 & \textbf{4.1e-2} & 6.2e-2 & 3.4e-1 & 7.0e-2 \\
AdultCensus (14) & \textbf{1.1e-4} & 2.0e-4 & 1.3e-3 & 4.2e-4 & 2.5e-3 & 3.8e-3 & 1.6e-2 \\
Estate (15) & \textbf{4.6e-4} & 1.2e-3 & 1.6e-3 & 3.2e-3 & 1.6e-2 & 1.6e-2 & 1.9e-2 \\
Crime (101) & \textbf{2.1e-3} & 2.1e-2 & 8.2e-2 & 1.0e-2 & 7.7e-2 & 3.4e-2 & 4.2e0 \\
Seismic Bumps (15) & \textbf{6.0e-4} & 1.0e-3 & 4.4e-3 & 2.5e-3 & 1.6e-2 & 1.4e-2 & 2.3e-2 \\
Churn (19) & \textbf{1.1e-3} & 3.5e-3 & 6.4e-3 & 8.9e-3 & 2.2e-2 & 3.6e-2 & 4.8e-2 \\
Credit-G (20) & \textbf{9.5e-4} & 9.6e-4 & 9.3e-3 & 1.4e-3 & 8.0e-3 & 1.1e-2 & 3.8e-2 \\
Bankruptcy (64) & \textbf{1.3e-3} & 8.8e-3 & 9.2e-2 & 6.5e-3 & 2.2e-1 & 4.9e-2 & 5.8e-1 \\
Superconductivity (81) & \textbf{1.5e-3} & 9.0e-3 & 2.9e-2 & 8.5e-3 & 9.1e-2 & 4.0e-2 & 2.1e0 \\
NATICUSdroid (86) & \textbf{1.9e-4} & 2.1e-3 & 1.8e-2 & 2.5e-3 & 3.1e-2 & 1.0e-2 & 2.0e0 \\
MIC (111) & \textbf{1.2e-3} & 1.1e-2 & 1.2e-1 & 1.2e-2 & 1.2e-1 & 5.7e-2 & 1.5e0 \\
\midrule
\multicolumn{8}{@{}l}{\emph{$T = 1024$}} \\
Language (IMDB) (14) & 2.1e-3 & 2.8e-3 & \textbf{1.8e-3} & 2.6e-3 & 5.4e-3 & 2.5e-2 & 9.3e-3 \\
ResNet18-SP (14) & \textbf{3.3e-5} & 3.6e-4 & 3.1e-4 & 3.5e-3 & 7.7e-3 & 1.5e-2 & 9.6e-3 \\
ViT4by4 (16) & \textbf{1.6e-3} & 3.7e-3 & 4.0e-3 & 8.9e-3 & 2.5e-2 & 3.8e-2 & 1.8e-2 \\
DINOv2-16 (16) & \textbf{2.5e-3} & 3.7e-3 & 3.3e-3 & 1.0e-2 & 3.4e-2 & 3.9e-2 & 1.0e-2 \\
Forest Fires (13) & 2.0e-2 & 2.1e-2 & 3.0e-2 & \textbf{1.7e-2} & 1.9e-2 & 1.9e-1 & 2.1e-2 \\
AdultCensus (14) & \textbf{1.8e-5} & 5.3e-5 & 1.8e-4 & 1.9e-4 & 1.0e-3 & 1.8e-3 & 5.7e-3 \\
Estate (15) & \textbf{1.5e-4} & 2.9e-4 & 4.3e-4 & 1.2e-3 & 7.2e-3 & 7.7e-3 & 8.3e-3 \\
Crime (101) & \textbf{7.4e-4} & 3.3e-3 & 4.5e-2 & 3.8e-3 & 3.4e-2 & 1.8e-2 & 1.6e0 \\
Seismic Bumps (15) & \textbf{1.1e-4} & 3.5e-4 & 8.5e-4 & 9.0e-4 & 6.3e-3 & 6.8e-3 & 9.0e-3 \\
Churn (19) & \textbf{2.9e-4} & 8.2e-4 & 1.5e-3 & 3.5e-3 & 1.2e-2 & 1.9e-2 & 2.0e-2 \\
Credit-G (20) & 4.8e-4 & \textbf{3.7e-4} & 2.4e-3 & 7.3e-4 & 3.5e-3 & 4.6e-3 & 1.5e-2 \\
Bankruptcy (64) & \textbf{5.0e-4} & 2.0e-3 & 4.4e-2 & 2.8e-3 & 1.0e-1 & 2.6e-2 & 1.8e-1 \\
Superconductivity (81) & \textbf{5.3e-4} & 1.9e-3 & 1.5e-2 & 3.3e-3 & 4.4e-2 & 1.8e-2 & 6.0e-1 \\
NATICUSdroid (86) & \textbf{3.9e-5} & 4.8e-4 & 8.0e-3 & 9.4e-4 & 1.5e-2 & 6.0e-3 & 6.2e-1 \\
MIC (111) & \textbf{3.8e-4} & 4.0e-3 & 6.2e-2 & 4.0e-3 & 5.2e-2 & 2.5e-2 & 7.2e-1 \\
\midrule
\multicolumn{8}{@{}l}{\emph{$T = 2048$}} \\
Language (IMDB) (14) & 4.9e-4 & 7.2e-4 & \textbf{4.5e-4} & 1.2e-3 & 2.0e-3 & 1.1e-2 & 3.9e-3 \\
ResNet18-SP (14) & \textbf{3.3e-6} & 5.0e-5 & 7.1e-5 & 1.2e-3 & 2.9e-3 & 7.8e-3 & 4.1e-3 \\
ViT4by4 (16) & \textbf{5.1e-4} & 1.2e-3 & 1.1e-3 & 3.7e-3 & 1.2e-2 & 1.6e-2 & 7.6e-3 \\
DINOv2-16 (16) & \textbf{6.5e-4} & 1.2e-3 & 7.6e-4 & 4.2e-3 & 1.5e-2 & 1.6e-2 & 4.1e-3 \\
Forest Fires (13) & 7.1e-3 & 9.2e-3 & 8.9e-3 & \textbf{5.5e-3} & 8.0e-3 & 9.9e-2 & 8.5e-3 \\
AdultCensus (14) & \textbf{7.9e-6} & 1.0e-5 & 2.2e-5 & 8.1e-5 & 4.1e-4 & 8.2e-4 & 2.2e-3 \\
Estate (15) & \textbf{6.1e-5} & 8.9e-5 & 7.6e-5 & 5.0e-4 & 3.5e-3 & 4.2e-3 & 3.4e-3 \\
Crime (101) & \textbf{2.8e-4} & 9.0e-4 & 2.3e-2 & 1.6e-3 & 1.6e-2 & 9.3e-3 & 2.4e-1 \\
Seismic Bumps (15) & \textbf{3.9e-5} & 7.1e-5 & 1.2e-4 & 3.9e-4 & 2.6e-3 & 3.6e-3 & 3.7e-3 \\
Churn (19) & \textbf{7.4e-5} & 2.3e-4 & 4.1e-4 & 1.7e-3 & 5.3e-3 & 8.4e-3 & 9.0e-3 \\
Credit-G (20) & \textbf{1.2e-4} & 1.6e-4 & 5.8e-4 & 3.0e-4 & 1.5e-3 & 2.1e-3 & 7.2e-3 \\
Bankruptcy (64) & \textbf{1.9e-4} & 6.2e-4 & 1.8e-2 & 1.2e-3 & 4.3e-2 & 1.2e-2 & 6.0e-2 \\
Superconductivity (81) & \textbf{2.1e-4} & 6.6e-4 & 6.9e-3 & 1.5e-3 & 1.8e-2 & 8.2e-3 & 1.0e-1 \\
NATICUSdroid (86) & \textbf{1.1e-5} & 1.1e-4 & 2.8e-3 & 4.3e-4 & 6.0e-3 & 2.7e-3 & 1.1e-1 \\
MIC (111) & \textbf{1.3e-4} & 8.2e-4 & 2.5e-2 & 1.7e-3 & 2.4e-2 & 1.4e-2 & 1.9e-1 \\
\midrule
\multicolumn{8}{@{}l}{\emph{$T = 4096$}} \\
Language (IMDB) (14) & 1.1e-4 & 1.6e-4 & \textbf{1.1e-4} & 3.3e-4 & 7.6e-4 & 6.4e-3 & 1.3e-3 \\
ResNet18-SP (14) & \textbf{1.7e-7} & 7.8e-6 & 1.2e-5 & 3.8e-4 & 1.4e-3 & 4.0e-3 & 1.3e-3 \\
ViT4by4 (16) & \textbf{1.8e-4} & 3.3e-4 & 2.7e-4 & 1.6e-3 & 4.3e-3 & 7.6e-3 & 2.9e-3 \\
DINOv2-16 (16) & 1.9e-4 & 3.3e-4 & \textbf{1.9e-4} & 1.7e-3 & 6.3e-3 & 8.0e-3 & 1.9e-3 \\
Forest Fires (13) & 1.9e-3 & 2.4e-3 & 1.5e-3 & \textbf{1.3e-3} & 2.1e-3 & 5.3e-2 & 2.5e-3 \\
AdultCensus (14) & 2.0e-6 & \textbf{1.7e-6} & 2.9e-6 & 2.7e-5 & 1.4e-4 & 3.7e-4 & 8.0e-4 \\
Estate (15) & \textbf{1.8e-5} & 2.1e-5 & 2.2e-5 & 2.1e-4 & 1.1e-3 & 2.1e-3 & 1.4e-3 \\
Crime (101) & \textbf{1.2e-4} & 3.6e-4 & 1.0e-2 & 7.5e-4 & 7.0e-3 & 4.1e-3 & 4.9e-2 \\
Seismic Bumps (15) & \textbf{1.3e-5} & 1.7e-5 & 2.6e-5 & 1.3e-4 & 1.1e-3 & 1.8e-3 & 1.5e-3 \\
Churn (19) & \textbf{2.5e-5} & 5.2e-5 & 8.8e-5 & 7.6e-4 & 2.4e-3 & 3.9e-3 & 3.7e-3 \\
Credit-G (20) & 5.2e-5 & \textbf{5.2e-5} & 1.3e-4 & 1.2e-4 & 7.2e-4 & 1.2e-3 & 2.7e-3 \\
Bankruptcy (64) & \textbf{7.7e-5} & 2.2e-4 & 6.4e-3 & 6.4e-4 & 2.2e-2 & 5.9e-3 & 2.5e-2 \\
Superconductivity (81) & \textbf{8.7e-5} & 2.4e-4 & 2.6e-3 & 7.2e-4 & 9.0e-3 & 4.2e-3 & 3.1e-2 \\
NATICUSdroid (86) & \textbf{2.8e-6} & 2.9e-5 & 9.0e-4 & 2.0e-4 & 2.8e-3 & 1.3e-3 & 3.6e-2 \\
MIC (111) & \textbf{4.1e-5} & 2.4e-4 & 8.6e-3 & 8.4e-4 & 1.1e-2 & 6.9e-3 & 5.4e-2 \\
\midrule
\multicolumn{8}{@{}l}{\emph{$T = 8192$}} \\
Language (IMDB) (14) & \textbf{2.1e-5} & 3.2e-5 & 2.1e-5 & 7.6e-5 & 2.0e-4 & 3.1e-3 & 3.7e-4 \\
ResNet18-SP (14) & \textbf{4.7e-9} & 5.9e-7 & 2.2e-6 & 9.9e-5 & 4.1e-4 & 2.2e-3 & 3.3e-4 \\
ViT4by4 (16) & \textbf{5.2e-5} & 8.5e-5 & 7.5e-5 & 5.1e-4 & 1.7e-3 & 4.2e-3 & 1.1e-3 \\
DINOv2-16 (16) & 5.7e-5 & 7.4e-5 & \textbf{3.7e-5} & 5.0e-4 & 2.5e-3 & 3.9e-3 & 4.9e-4 \\
Forest Fires (13) & -- & -- & -- & -- & -- & -- & -- \\
AdultCensus (14) & 7.8e-7 & \textbf{3.5e-7} & 4.9e-7 & 6.1e-6 & 3.9e-5 & 2.1e-4 & 2.3e-4 \\
Estate (15) & 5.9e-6 & \textbf{4.8e-6} & 5.8e-6 & 6.7e-5 & 4.1e-4 & 8.9e-4 & 5.1e-4 \\
Crime (101) & \textbf{5.1e-5} & 1.5e-4 & 4.0e-3 & 3.7e-4 & 3.3e-3 & 2.1e-3 & 2.0e-2 \\
Seismic Bumps (15) & 4.9e-6 & \textbf{3.8e-6} & 6.3e-6 & 4.5e-5 & 3.7e-4 & 9.5e-4 & 5.9e-4 \\
Churn (19) & \textbf{9.8e-6} & 1.5e-5 & 2.6e-5 & 3.6e-4 & 1.0e-3 & 2.0e-3 & 1.8e-3 \\
Credit-G (20) & 2.4e-5 & \textbf{1.9e-5} & 3.4e-5 & 5.7e-5 & 3.6e-4 & 5.7e-4 & 1.3e-3 \\
Bankruptcy (64) & \textbf{2.4e-5} & 8.4e-5 & 2.1e-3 & 3.0e-4 & 1.1e-2 & 3.0e-3 & 1.1e-2 \\
Superconductivity (81) & \textbf{3.6e-5} & 9.1e-5 & 9.2e-4 & 3.4e-4 & 4.3e-3 & 2.0e-3 & 1.4e-2 \\
NATICUSdroid (86) & \textbf{6.3e-7} & 8.6e-6 & 2.7e-4 & 9.4e-5 & 1.3e-3 & 6.2e-4 & 1.6e-2 \\
MIC (111) & \textbf{1.0e-5} & 7.9e-5 & 2.6e-3 & 3.9e-4 & 5.5e-3 & 3.3e-3 & 2.3e-2 \\
\bottomrule
\end{tabular}

\end{table}
\begin{table}[p]
  \caption{NMSE per game (arithmetic mean over $30$ seeds) at every recorded budget $\budget$; otherwise as \cref{tab:pergame_all}.}
  \label{tab:pergame_all_mean}
  \centering \tiny \setlength{\tabcolsep}{4pt} \renewcommand{\arraystretch}{0.84}
  \begin{tabular}{@{}lrrrrrrr@{}}
\toprule
Game ($p$) & ShapleyOG & OddSHAP & Regression MSR & Leverage SHAP & Kernel SHAP & Permutation & SVARM \\
\midrule
\multicolumn{8}{@{}l}{\emph{$T = 256$}} \\
Language (IMDB) (14) & 6.0e-2 & 6.0e-2 & \textbf{5.2e-2} & 5.6e-2 & 5.7e-2 & 2.3e-1 & 8.9e-2 \\
ResNet18-SP (14) & \textbf{4.5e-3} & 1.4e-2 & 1.2e-2 & 3.2e-2 & 6.7e-2 & 1.0e-1 & 6.1e-2 \\
ViT4by4 (16) & \textbf{3.9e-2} & 4.6e-2 & 4.7e-2 & 6.3e-2 & 1.7e-1 & 2.0e-1 & 1.4e-1 \\
DINOv2-16 (16) & 7.5e-2 & 5.0e-2 & \textbf{4.3e-2} & 1.1e-1 & 3.4e-1 & 2.3e-1 & 1.3e-1 \\
Forest Fires (13) & 1.5e-1 & 3.0e-1 & 4.8e-1 & \textbf{1.1e-1} & 3.0e-1 & 2.2e0 & 2.9e-1 \\
AdultCensus (14) & \textbf{2.8e-4} & 9.4e-4 & 7.0e-3 & 1.7e-3 & 7.4e-3 & 9.9e-3 & 4.5e-2 \\
Estate (15) & \textbf{3.0e-3} & 6.2e-3 & 8.7e-3 & 1.0e-2 & 9.1e-2 & 3.3e-2 & 6.7e-2 \\
Crime (101) & \textbf{8.2e-3} & 8.1e-2 & 1.5e-1 & 8.6e-2 & 3.0e-1 & 9.7e-2 & 7.0e0 \\
Seismic Bumps (15) & \textbf{1.8e-3} & 5.5e-3 & 2.1e-2 & 6.2e-3 & 4.7e-2 & 3.7e-2 & 7.2e-2 \\
Churn (19) & \textbf{5.0e-3} & 2.0e-2 & 2.4e-2 & 3.0e-2 & 1.0e-1 & 1.0e-1 & 1.4e-1 \\
Credit-G (20) & \textbf{2.4e-3} & 4.3e-3 & 2.7e-2 & 5.2e-3 & 3.0e-2 & 3.1e-2 & 1.1e-1 \\
Bankruptcy (64) & \textbf{4.9e-3} & 5.0e-2 & 1.8e-1 & 2.4e-2 & 6.1e-1 & 1.1e-1 & 1.5e0 \\
Superconductivity (81) & \textbf{5.1e-3} & 3.2e-2 & 6.4e-2 & 3.9e-2 & 3.2e-1 & 8.3e-2 & 4.7e0 \\
NATICUSdroid (86) & \textbf{1.1e-3} & 7.6e-3 & 4.4e-2 & 1.2e-2 & 1.1e-1 & 4.3e-2 & 4.2e0 \\
MIC (111) & \textbf{1.3e-2} & 5.2e-2 & 2.4e-1 & 1.0e-1 & 4.1e-1 & 1.3e-1 & 3.1e0 \\
\midrule
\multicolumn{8}{@{}l}{\emph{$T = 512$}} \\
Language (IMDB) (14) & 2.0e-2 & 2.2e-2 & \textbf{1.7e-2} & 2.1e-2 & 2.6e-2 & 8.9e-2 & 3.5e-2 \\
ResNet18-SP (14) & \textbf{7.6e-4} & 3.4e-3 & 3.0e-3 & 1.2e-2 & 3.3e-2 & 4.9e-2 & 2.4e-2 \\
ViT4by4 (16) & \textbf{7.7e-3} & 1.7e-2 & 1.5e-2 & 2.8e-2 & 7.7e-2 & 8.7e-2 & 4.6e-2 \\
DINOv2-16 (16) & 1.9e-2 & \textbf{1.7e-2} & 1.7e-2 & 4.7e-2 & 1.7e-1 & 1.1e-1 & 3.1e-2 \\
Forest Fires (13) & 1.1e-1 & 1.1e-1 & 1.7e-1 & \textbf{5.5e-2} & 1.1e-1 & 7.1e-1 & 1.1e-1 \\
AdultCensus (14) & \textbf{1.2e-4} & 2.8e-4 & 1.7e-3 & 5.4e-4 & 3.3e-3 & 4.9e-3 & 1.7e-2 \\
Estate (15) & \textbf{6.3e-4} & 1.9e-3 & 2.4e-3 & 5.3e-3 & 3.0e-2 & 2.6e-2 & 2.1e-2 \\
Crime (101) & \textbf{2.4e-3} & 2.5e-2 & 9.9e-2 & 1.2e-2 & 1.2e-1 & 4.2e-2 & 5.5e0 \\
Seismic Bumps (15) & \textbf{7.2e-4} & 1.4e-3 & 5.2e-3 & 3.7e-3 & 1.9e-2 & 1.7e-2 & 2.6e-2 \\
Churn (19) & \textbf{1.3e-3} & 4.2e-3 & 7.7e-3 & 1.6e-2 & 3.9e-2 & 5.0e-2 & 5.0e-2 \\
Credit-G (20) & 1.4e-3 & \textbf{1.2e-3} & 1.2e-2 & 2.1e-3 & 1.5e-2 & 1.4e-2 & 4.0e-2 \\
Bankruptcy (64) & \textbf{1.4e-3} & 1.0e-2 & 1.0e-1 & 7.8e-3 & 2.6e-1 & 7.3e-2 & 7.4e-1 \\
Superconductivity (81) & \textbf{1.6e-3} & 1.0e-2 & 3.6e-2 & 9.9e-3 & 1.4e-1 & 5.0e-2 & 3.0e0 \\
NATICUSdroid (86) & \textbf{2.1e-4} & 2.6e-3 & 2.1e-2 & 3.4e-3 & 4.8e-2 & 1.6e-2 & 2.7e0 \\
MIC (111) & \textbf{1.4e-3} & 1.1e-2 & 1.4e-1 & 1.4e-2 & 1.5e-1 & 7.1e-2 & 2.3e0 \\
\midrule
\multicolumn{8}{@{}l}{\emph{$T = 1024$}} \\
Language (IMDB) (14) & 5.5e-3 & 6.9e-3 & \textbf{3.9e-3} & 6.2e-3 & 9.6e-3 & 5.6e-2 & 1.1e-2 \\
ResNet18-SP (14) & \textbf{1.0e-4} & 7.5e-4 & 9.4e-4 & 5.2e-3 & 1.2e-2 & 2.4e-2 & 1.1e-2 \\
ViT4by4 (16) & \textbf{2.3e-3} & 5.1e-3 & 5.6e-3 & 1.2e-2 & 3.3e-2 & 4.7e-2 & 2.0e-2 \\
DINOv2-16 (16) & \textbf{4.1e-3} & 5.2e-3 & 4.3e-3 & 1.5e-2 & 5.9e-2 & 5.6e-2 & 1.2e-2 \\
Forest Fires (13) & 3.4e-2 & 3.7e-2 & 1.3e-1 & \textbf{2.5e-2} & 2.8e-2 & 2.9e-1 & 3.0e-2 \\
AdultCensus (14) & \textbf{2.1e-5} & 6.0e-5 & 2.1e-4 & 2.3e-4 & 1.5e-3 & 2.5e-3 & 6.1e-3 \\
Estate (15) & \textbf{2.3e-4} & 4.1e-4 & 6.3e-4 & 2.1e-3 & 1.2e-2 & 1.1e-2 & 8.8e-3 \\
Crime (101) & \textbf{8.4e-4} & 3.9e-3 & 5.4e-2 & 4.7e-3 & 5.2e-2 & 2.3e-2 & 2.0e0 \\
Seismic Bumps (15) & \textbf{1.6e-4} & 4.5e-4 & 1.0e-3 & 1.1e-3 & 8.2e-3 & 9.1e-3 & 1.0e-2 \\
Churn (19) & \textbf{3.5e-4} & 9.7e-4 & 1.8e-3 & 6.0e-3 & 1.9e-2 & 3.0e-2 & 2.1e-2 \\
Credit-G (20) & 7.5e-4 & \textbf{5.0e-4} & 2.9e-3 & 1.1e-3 & 5.5e-3 & 5.7e-3 & 1.7e-2 \\
Bankruptcy (64) & \textbf{5.5e-4} & 2.2e-3 & 5.0e-2 & 3.3e-3 & 1.2e-1 & 3.8e-2 & 2.0e-1 \\
Superconductivity (81) & \textbf{5.5e-4} & 2.1e-3 & 1.8e-2 & 3.9e-3 & 7.5e-2 & 2.2e-2 & 7.5e-1 \\
NATICUSdroid (86) & \textbf{4.5e-5} & 5.8e-4 & 9.3e-3 & 1.3e-3 & 2.3e-2 & 9.1e-3 & 7.6e-1 \\
MIC (111) & \textbf{4.2e-4} & 4.5e-3 & 6.8e-2 & 4.6e-3 & 6.4e-2 & 3.0e-2 & 1.0e0 \\
\midrule
\multicolumn{8}{@{}l}{\emph{$T = 2048$}} \\
Language (IMDB) (14) & 1.1e-3 & 1.7e-3 & \textbf{1.1e-3} & 2.4e-3 & 3.6e-3 & 2.5e-2 & 4.5e-3 \\
ResNet18-SP (14) & \textbf{1.3e-5} & 1.5e-4 & 2.1e-4 & 1.9e-3 & 4.6e-3 & 1.1e-2 & 4.6e-3 \\
ViT4by4 (16) & \textbf{7.5e-4} & 1.7e-3 & 1.6e-3 & 4.5e-3 & 1.5e-2 & 2.0e-2 & 7.9e-3 \\
DINOv2-16 (16) & \textbf{1.0e-3} & 1.6e-3 & 1.1e-3 & 6.2e-3 & 2.4e-2 & 2.3e-2 & 5.1e-3 \\
Forest Fires (13) & 1.2e-2 & 1.7e-2 & 1.5e-2 & \textbf{7.8e-3} & 1.4e-2 & 1.3e-1 & 1.5e-2 \\
AdultCensus (14) & \textbf{9.1e-6} & 1.2e-5 & 2.6e-5 & 9.4e-5 & 6.0e-4 & 1.1e-3 & 2.4e-3 \\
Estate (15) & \textbf{9.3e-5} & 1.4e-4 & 9.8e-5 & 8.4e-4 & 5.6e-3 & 6.7e-3 & 3.8e-3 \\
Crime (101) & \textbf{3.3e-4} & 1.1e-3 & 2.8e-2 & 2.0e-3 & 2.4e-2 & 1.1e-2 & 2.7e-1 \\
Seismic Bumps (15) & \textbf{6.3e-5} & 1.0e-4 & 1.5e-4 & 5.6e-4 & 3.1e-3 & 4.5e-3 & 4.1e-3 \\
Churn (19) & \textbf{8.9e-5} & 2.8e-4 & 4.8e-4 & 2.7e-3 & 8.0e-3 & 1.3e-2 & 1.0e-2 \\
Credit-G (20) & \textbf{2.0e-4} & 2.0e-4 & 6.8e-4 & 4.1e-4 & 2.5e-3 & 2.9e-3 & 7.8e-3 \\
Bankruptcy (64) & \textbf{2.1e-4} & 7.1e-4 & 2.1e-2 & 1.4e-3 & 5.2e-2 & 1.8e-2 & 6.1e-2 \\
Superconductivity (81) & \textbf{2.2e-4} & 7.2e-4 & 8.1e-3 & 1.8e-3 & 3.1e-2 & 1.1e-2 & 1.0e-1 \\
NATICUSdroid (86) & \textbf{1.3e-5} & 1.3e-4 & 3.3e-3 & 5.8e-4 & 9.4e-3 & 4.0e-3 & 1.1e-1 \\
MIC (111) & \textbf{1.4e-4} & 9.0e-4 & 2.7e-2 & 1.9e-3 & 2.9e-2 & 1.6e-2 & 2.2e-1 \\
\midrule
\multicolumn{8}{@{}l}{\emph{$T = 4096$}} \\
Language (IMDB) (14) & \textbf{2.3e-4} & 3.6e-4 & 2.8e-4 & 6.7e-4 & 1.3e-3 & 1.7e-2 & 1.7e-3 \\
ResNet18-SP (14) & \textbf{8.4e-7} & 2.3e-5 & 2.6e-5 & 5.8e-4 & 1.9e-3 & 6.0e-3 & 1.5e-3 \\
ViT4by4 (16) & \textbf{2.9e-4} & 4.6e-4 & 4.0e-4 & 2.2e-3 & 5.5e-3 & 1.0e-2 & 3.0e-3 \\
DINOv2-16 (16) & 3.8e-4 & 4.7e-4 & \textbf{2.7e-4} & 2.6e-3 & 9.6e-3 & 1.2e-2 & 2.4e-3 \\
Forest Fires (13) & 3.1e-3 & 3.9e-3 & 2.5e-3 & \textbf{2.0e-3} & 3.0e-3 & 7.4e-2 & 3.2e-3 \\
AdultCensus (14) & 2.2e-6 & \textbf{2.0e-6} & 3.6e-6 & 3.1e-5 & 2.0e-4 & 5.1e-4 & 8.8e-4 \\
Estate (15) & 2.9e-5 & 3.6e-5 & \textbf{2.9e-5} & 3.3e-4 & 1.9e-3 & 4.1e-3 & 1.5e-3 \\
Crime (101) & \textbf{1.4e-4} & 4.3e-4 & 1.3e-2 & 9.8e-4 & 1.1e-2 & 5.0e-3 & 5.0e-2 \\
Seismic Bumps (15) & 2.2e-5 & \textbf{2.1e-5} & 3.2e-5 & 1.6e-4 & 1.3e-3 & 2.1e-3 & 1.7e-3 \\
Churn (19) & \textbf{3.5e-5} & 6.7e-5 & 1.0e-4 & 1.1e-3 & 3.7e-3 & 5.4e-3 & 4.0e-3 \\
Credit-G (20) & 7.5e-5 & \textbf{6.8e-5} & 1.7e-4 & 1.8e-4 & 1.1e-3 & 1.5e-3 & 3.0e-3 \\
Bankruptcy (64) & \textbf{8.7e-5} & 2.4e-4 & 7.5e-3 & 7.4e-4 & 2.6e-2 & 8.6e-3 & 2.5e-2 \\
Superconductivity (81) & \textbf{9.2e-5} & 2.6e-4 & 3.0e-3 & 8.7e-4 & 1.5e-2 & 5.1e-3 & 3.2e-2 \\
NATICUSdroid (86) & \textbf{3.2e-6} & 3.5e-5 & 1.1e-3 & 2.5e-4 & 4.7e-3 & 2.0e-3 & 3.6e-2 \\
MIC (111) & \textbf{4.5e-5} & 2.7e-4 & 9.5e-3 & 9.4e-4 & 1.4e-2 & 8.5e-3 & 5.5e-2 \\
\midrule
\multicolumn{8}{@{}l}{\emph{$T = 8192$}} \\
Language (IMDB) (14) & \textbf{4.1e-5} & 7.2e-5 & 5.5e-5 & 1.6e-4 & 3.2e-4 & 7.8e-3 & 4.2e-4 \\
ResNet18-SP (14) & \textbf{1.8e-8} & 3.0e-6 & 5.9e-6 & 1.7e-4 & 5.7e-4 & 3.0e-3 & 4.4e-4 \\
ViT4by4 (16) & \textbf{8.3e-5} & 1.4e-4 & 1.1e-4 & 6.2e-4 & 2.2e-3 & 5.2e-3 & 1.3e-3 \\
DINOv2-16 (16) & 1.2e-4 & 1.1e-4 & \textbf{5.8e-5} & 7.3e-4 & 3.7e-3 & 5.7e-3 & 6.6e-4 \\
Forest Fires (13) & -- & -- & -- & -- & -- & -- & -- \\
AdultCensus (14) & 9.1e-7 & \textbf{4.1e-7} & 5.5e-7 & 6.9e-6 & 7.0e-5 & 2.6e-4 & 2.5e-4 \\
Estate (15) & 7.8e-6 & \textbf{7.3e-6} & 8.4e-6 & 1.1e-4 & 7.3e-4 & 1.5e-3 & 5.7e-4 \\
Crime (101) & \textbf{5.9e-5} & 1.8e-4 & 5.0e-3 & 4.5e-4 & 5.0e-3 & 3.0e-3 & 2.0e-2 \\
Seismic Bumps (15) & 6.8e-6 & \textbf{5.2e-6} & 8.1e-6 & 5.9e-5 & 4.6e-4 & 1.4e-3 & 6.8e-4 \\
Churn (19) & \textbf{1.2e-5} & 1.8e-5 & 3.3e-5 & 5.5e-4 & 1.6e-3 & 2.8e-3 & 1.9e-3 \\
Credit-G (20) & 3.3e-5 & \textbf{2.7e-5} & 4.5e-5 & 8.3e-5 & 6.4e-4 & 7.6e-4 & 1.4e-3 \\
Bankruptcy (64) & \textbf{2.7e-5} & 9.2e-5 & 2.5e-3 & 3.5e-4 & 1.3e-2 & 4.5e-3 & 1.2e-2 \\
Superconductivity (81) & \textbf{3.8e-5} & 9.8e-5 & 1.1e-3 & 4.0e-4 & 6.9e-3 & 2.6e-3 & 1.5e-2 \\
NATICUSdroid (86) & \textbf{7.0e-7} & 9.9e-6 & 3.2e-4 & 1.3e-4 & 2.2e-3 & 9.6e-4 & 1.6e-2 \\
MIC (111) & \textbf{1.1e-5} & 8.9e-5 & 3.0e-3 & 4.3e-4 & 6.9e-3 & 4.2e-3 & 2.3e-2 \\
\bottomrule
\end{tabular}

\end{table}
\clearpage
\subsection{Average Ranks} \label{app:res_avgrank}
\Cref{tab:avgrank_methods} extends the right panel of \cref{fig:main_results} to the three strongest competitors: for every game and budget, it lists the average rank of \methodname{}, OddSHAP, Regression MSR, and Leverage SHAP among all seven estimators (\cref{app:exp_metrics}).
\begin{table}[htbp]
  \caption{Average rank (over $30$ seeds, $1$ = lowest NMSE) of \methodname{} and its three strongest competitors among all seven estimators, per game and budget $\budget$; \emph{pooled} averages over the $14$ games whose grid reaches $8192$ evaluations; -- marks budgets beyond a game's grid. Gold, silver, bronze: lowest, second lowest, third lowest average rank of all seven estimators.}
  \label{tab:avgrank_methods}
  \centering \scriptsize \setlength{\tabcolsep}{2.5pt}
  \resizebox{\textwidth}{!}{\begin{tabular}{@{}lrrrrrrrrrrrrrrrr@{}}
\toprule
Method & \rotatebox{60}{Language (IMDB) (14)} & \rotatebox{60}{ResNet18-SP (14)} & \rotatebox{60}{ViT4by4 (16)} & \rotatebox{60}{DINOv2-16 (16)} & \rotatebox{60}{Forest Fires (13)} & \rotatebox{60}{AdultCensus (14)} & \rotatebox{60}{Estate (15)} & \rotatebox{60}{Crime (101)} & \rotatebox{60}{Seismic Bumps (15)} & \rotatebox{60}{Churn (19)} & \rotatebox{60}{Credit-G (20)} & \rotatebox{60}{Bankruptcy (64)} & \rotatebox{60}{Superconductivity (81)} & \rotatebox{60}{NATICUSdroid (86)} & \rotatebox{60}{MIC (111)} & \rotatebox{60}{\emph{pooled}} \\
\midrule
\multicolumn{17}{@{}l}{\emph{$T = 256$}} \\
ShapleyOG (Ours) & \cellcolor{gold!60}2.9 & \cellcolor{gold!60}1.1 & \cellcolor{gold!60}2.0 & \cellcolor{bronze!60}2.5 & \cellcolor{silver!60}3.1 & \cellcolor{gold!60}1.1 & \cellcolor{gold!60}1.4 & \cellcolor{gold!60}1.0 & \cellcolor{gold!60}1.0 & \cellcolor{gold!60}1.1 & \cellcolor{gold!60}1.4 & \cellcolor{gold!60}1.0 & \cellcolor{gold!60}1.0 & \cellcolor{gold!60}1.0 & \cellcolor{gold!60}1.0 & \cellcolor{gold!60}1.4 \\
OddSHAP & \cellcolor{bronze!60}3.2 & \cellcolor{bronze!60}2.9 & \cellcolor{bronze!60}2.8 & \cellcolor{silver!60}2.3 & 4.1 & \cellcolor{silver!60}2.2 & \cellcolor{silver!60}2.2 & \cellcolor{bronze!60}3.2 & \cellcolor{silver!60}2.4 & \cellcolor{silver!60}2.7 & \cellcolor{bronze!60}2.4 & \cellcolor{bronze!60}3.0 & \cellcolor{silver!60}2.5 & \cellcolor{silver!60}2.2 & \cellcolor{silver!60}2.2 & \cellcolor{silver!60}2.6 \\
Regression MSR & \cellcolor{silver!60}2.9 & \cellcolor{silver!60}2.4 & \cellcolor{silver!60}2.6 & \cellcolor{gold!60}2.2 & 4.2 & 4.7 & \cellcolor{bronze!60}3.1 & 4.9 & 4.3 & 3.6 & 5.2 & 4.9 & 4.2 & 4.7 & 4.9 & 3.9 \\
Leverage SHAP & 3.2 & 4.3 & 3.5 & 3.7 & \cellcolor{gold!60}2.6 & \cellcolor{bronze!60}2.9 & 3.4 & \cellcolor{silver!60}2.8 & \cellcolor{bronze!60}2.7 & \cellcolor{bronze!60}3.1 & \cellcolor{silver!60}2.4 & \cellcolor{silver!60}2.1 & \cellcolor{bronze!60}3.0 & \cellcolor{bronze!60}3.0 & \cellcolor{bronze!60}3.4 & \cellcolor{bronze!60}3.1 \\
\midrule
\multicolumn{17}{@{}l}{\emph{$T = 512$}} \\
ShapleyOG (Ours) & \cellcolor{silver!60}2.6 & \cellcolor{gold!60}1.0 & \cellcolor{gold!60}1.2 & \cellcolor{gold!60}2.1 & \cellcolor{silver!60}3.1 & \cellcolor{gold!60}1.2 & \cellcolor{gold!60}1.1 & \cellcolor{gold!60}1.0 & \cellcolor{gold!60}1.3 & \cellcolor{gold!60}1.0 & \cellcolor{silver!60}2.0 & \cellcolor{gold!60}1.0 & \cellcolor{gold!60}1.0 & \cellcolor{gold!60}1.0 & \cellcolor{gold!60}1.0 & \cellcolor{gold!60}1.3 \\
OddSHAP & \cellcolor{bronze!60}3.2 & \cellcolor{bronze!60}2.6 & \cellcolor{bronze!60}2.9 & \cellcolor{silver!60}2.5 & 3.7 & \cellcolor{silver!60}2.1 & \cellcolor{silver!60}2.4 & \cellcolor{bronze!60}3.1 & \cellcolor{silver!60}1.9 & \cellcolor{silver!60}2.3 & \cellcolor{gold!60}1.8 & \cellcolor{bronze!60}2.8 & \cellcolor{silver!60}2.5 & \cellcolor{silver!60}2.4 & \cellcolor{silver!60}2.3 & \cellcolor{silver!60}2.5 \\
Regression MSR & \cellcolor{gold!60}2.6 & \cellcolor{silver!60}2.5 & \cellcolor{silver!60}2.5 & \cellcolor{bronze!60}2.6 & 4.4 & 4.3 & \cellcolor{bronze!60}2.8 & 5.4 & 3.9 & \cellcolor{bronze!60}3.4 & 5.0 & 4.9 & 4.4 & 5.0 & 5.4 & 3.9 \\
Leverage SHAP & 3.3 & 4.2 & 4.1 & 4.0 & \cellcolor{gold!60}2.4 & \cellcolor{bronze!60}2.9 & 3.9 & \cellcolor{silver!60}2.1 & \cellcolor{bronze!60}3.2 & 3.8 & \cellcolor{bronze!60}2.3 & \cellcolor{silver!60}2.2 & \cellcolor{bronze!60}2.6 & \cellcolor{bronze!60}2.7 & \cellcolor{bronze!60}2.7 & \cellcolor{bronze!60}3.1 \\
\midrule
\multicolumn{17}{@{}l}{\emph{$T = 1024$}} \\
ShapleyOG (Ours) & \cellcolor{silver!60}2.6 & \cellcolor{gold!60}1.0 & \cellcolor{gold!60}1.2 & \cellcolor{gold!60}1.8 & \cellcolor{bronze!60}3.3 & \cellcolor{gold!60}1.1 & \cellcolor{gold!60}1.3 & \cellcolor{gold!60}1.0 & \cellcolor{gold!60}1.1 & \cellcolor{gold!60}1.0 & \cellcolor{silver!60}2.2 & \cellcolor{gold!60}1.0 & \cellcolor{gold!60}1.0 & \cellcolor{gold!60}1.0 & \cellcolor{gold!60}1.0 & \cellcolor{gold!60}1.3 \\
OddSHAP & 3.2 & \cellcolor{silver!60}2.5 & \cellcolor{silver!60}2.5 & \cellcolor{bronze!60}2.7 & 3.8 & \cellcolor{silver!60}2.0 & \cellcolor{silver!60}2.1 & \cellcolor{silver!60}2.4 & \cellcolor{silver!60}2.1 & \cellcolor{silver!60}2.1 & \cellcolor{gold!60}1.5 & \cellcolor{silver!60}2.2 & \cellcolor{silver!60}2.2 & \cellcolor{silver!60}2.0 & \cellcolor{silver!60}2.5 & \cellcolor{silver!60}2.3 \\
Regression MSR & \cellcolor{gold!60}2.3 & \cellcolor{bronze!60}2.5 & \cellcolor{bronze!60}2.6 & \cellcolor{silver!60}2.3 & 4.3 & 3.6 & \cellcolor{bronze!60}2.7 & 5.5 & 3.4 & \cellcolor{bronze!60}3.2 & 4.5 & 4.8 & 4.5 & 5.0 & 5.6 & 3.8 \\
Leverage SHAP & \cellcolor{bronze!60}3.1 & 4.4 & 4.2 & 4.3 & \cellcolor{gold!60}3.0 & \cellcolor{bronze!60}3.4 & 3.9 & \cellcolor{bronze!60}2.7 & \cellcolor{bronze!60}3.4 & 3.8 & \cellcolor{bronze!60}2.5 & \cellcolor{bronze!60}2.8 & \cellcolor{bronze!60}2.9 & \cellcolor{bronze!60}3.0 & \cellcolor{bronze!60}2.5 & \cellcolor{bronze!60}3.4 \\
\midrule
\multicolumn{17}{@{}l}{\emph{$T = 2048$}} \\
ShapleyOG (Ours) & \cellcolor{silver!60}2.0 & \cellcolor{gold!60}1.0 & \cellcolor{gold!60}1.2 & \cellcolor{silver!60}1.8 & \cellcolor{silver!60}3.4 & \cellcolor{gold!60}1.4 & \cellcolor{gold!60}1.6 & \cellcolor{gold!60}1.0 & \cellcolor{gold!60}1.4 & \cellcolor{gold!60}1.0 & \cellcolor{gold!60}1.6 & \cellcolor{gold!60}1.0 & \cellcolor{gold!60}1.0 & \cellcolor{gold!60}1.0 & \cellcolor{gold!60}1.0 & \cellcolor{gold!60}1.3 \\
OddSHAP & \cellcolor{bronze!60}2.8 & \cellcolor{silver!60}2.4 & \cellcolor{bronze!60}2.5 & \cellcolor{bronze!60}2.7 & 4.0 & \cellcolor{silver!60}1.8 & \cellcolor{bronze!60}2.3 & \cellcolor{silver!60}2.0 & \cellcolor{silver!60}2.1 & \cellcolor{silver!60}2.1 & \cellcolor{silver!60}2.0 & \cellcolor{silver!60}2.1 & \cellcolor{silver!60}2.1 & \cellcolor{silver!60}2.0 & \cellcolor{silver!60}2.1 & \cellcolor{silver!60}2.2 \\
Regression MSR & \cellcolor{gold!60}1.7 & \cellcolor{bronze!60}2.6 & \cellcolor{silver!60}2.4 & \cellcolor{gold!60}1.8 & 3.9 & \cellcolor{bronze!60}2.9 & \cellcolor{silver!60}2.1 & 5.6 & \cellcolor{bronze!60}2.6 & \cellcolor{bronze!60}2.9 & 3.9 & 4.9 & 4.5 & 4.6 & 5.4 & \cellcolor{bronze!60}3.4 \\
Leverage SHAP & 4.3 & 4.4 & 4.3 & 4.5 & \cellcolor{gold!60}2.2 & 3.9 & 4.0 & \cellcolor{bronze!60}3.0 & 4.0 & 4.1 & \cellcolor{bronze!60}2.8 & \cellcolor{bronze!60}2.9 & \cellcolor{bronze!60}2.9 & \cellcolor{bronze!60}3.0 & \cellcolor{bronze!60}2.9 & 3.6 \\
\midrule
\multicolumn{17}{@{}l}{\emph{$T = 4096$}} \\
ShapleyOG (Ours) & \cellcolor{gold!60}2.0 & \cellcolor{gold!60}1.0 & \cellcolor{gold!60}1.4 & \cellcolor{silver!60}1.9 & \cellcolor{bronze!60}3.3 & \cellcolor{silver!60}2.0 & \cellcolor{gold!60}1.7 & \cellcolor{gold!60}1.0 & \cellcolor{gold!60}1.5 & \cellcolor{gold!60}1.1 & \cellcolor{silver!60}1.8 & \cellcolor{gold!60}1.0 & \cellcolor{gold!60}1.0 & \cellcolor{gold!60}1.0 & \cellcolor{gold!60}1.0 & \cellcolor{gold!60}1.4 \\
OddSHAP & \cellcolor{bronze!60}2.5 & \cellcolor{silver!60}2.4 & \cellcolor{bronze!60}2.4 & \cellcolor{bronze!60}2.5 & 4.3 & \cellcolor{gold!60}1.7 & \cellcolor{silver!60}2.1 & \cellcolor{silver!60}2.0 & \cellcolor{silver!60}2.0 & \cellcolor{silver!60}2.0 & \cellcolor{gold!60}1.7 & \cellcolor{silver!60}2.0 & \cellcolor{silver!60}2.0 & \cellcolor{silver!60}2.0 & \cellcolor{silver!60}2.0 & \cellcolor{silver!60}2.1 \\
Regression MSR & \cellcolor{silver!60}2.1 & \cellcolor{bronze!60}2.6 & \cellcolor{silver!60}2.2 & \cellcolor{gold!60}1.7 & \cellcolor{silver!60}2.8 & \cellcolor{bronze!60}2.3 & \cellcolor{bronze!60}2.2 & 5.6 & \cellcolor{bronze!60}2.5 & \cellcolor{bronze!60}2.8 & 3.5 & 4.7 & 4.4 & 4.3 & 4.9 & \cellcolor{bronze!60}3.3 \\
Leverage SHAP & 4.0 & 4.3 & 4.4 & 4.6 & \cellcolor{gold!60}2.3 & 4.0 & 4.1 & \cellcolor{bronze!60}3.0 & 4.0 & 4.0 & \cellcolor{bronze!60}3.1 & \cellcolor{bronze!60}3.0 & \cellcolor{bronze!60}3.0 & \cellcolor{bronze!60}3.0 & \cellcolor{bronze!60}3.0 & 3.7 \\
\midrule
\multicolumn{17}{@{}l}{\emph{$T = 8192$}} \\
ShapleyOG (Ours) & \cellcolor{gold!60}1.7 & \cellcolor{gold!60}1.0 & \cellcolor{gold!60}1.6 & \cellcolor{silver!60}2.3 & -- & \cellcolor{bronze!60}2.6 & \cellcolor{silver!60}2.1 & \cellcolor{gold!60}1.0 & \cellcolor{silver!60}2.0 & \cellcolor{gold!60}1.3 & \cellcolor{silver!60}2.1 & \cellcolor{gold!60}1.0 & \cellcolor{gold!60}1.0 & \cellcolor{gold!60}1.0 & \cellcolor{gold!60}1.0 & \cellcolor{gold!60}1.5 \\
OddSHAP & \cellcolor{bronze!60}2.5 & \cellcolor{silver!60}2.0 & \cellcolor{bronze!60}2.3 & \cellcolor{bronze!60}2.4 & -- & \cellcolor{gold!60}1.3 & \cellcolor{gold!60}1.7 & \cellcolor{silver!60}2.0 & \cellcolor{gold!60}1.5 & \cellcolor{silver!60}2.0 & \cellcolor{gold!60}1.6 & \cellcolor{silver!60}2.0 & \cellcolor{silver!60}2.0 & \cellcolor{silver!60}2.0 & \cellcolor{silver!60}2.0 & \cellcolor{silver!60}2.0 \\
Regression MSR & \cellcolor{silver!60}2.0 & \cellcolor{bronze!60}2.9 & \cellcolor{silver!60}2.2 & \cellcolor{gold!60}1.6 & -- & \cellcolor{silver!60}2.1 & \cellcolor{bronze!60}2.2 & 5.2 & \cellcolor{bronze!60}2.5 & \cellcolor{bronze!60}2.8 & \cellcolor{bronze!60}2.8 & 4.4 & 4.2 & 4.0 & 4.5 & \cellcolor{bronze!60}3.1 \\
Leverage SHAP & 4.0 & 4.2 & 4.2 & 4.4 & -- & 4.0 & 4.0 & \cellcolor{bronze!60}3.0 & 4.1 & 4.1 & 3.5 & \cellcolor{bronze!60}3.0 & \cellcolor{bronze!60}3.0 & \cellcolor{bronze!60}3.1 & \cellcolor{bronze!60}3.0 & 3.7 \\
\bottomrule
\end{tabular}
}
\end{table}
\subsection{Overhead of \methodname{}} \label{app:res_overhead}
\Cref{tab:overhead_breakdown} splits the overhead of \methodname{} reported in \cref{fig:pareto} for three games of different size from the main experiment into the components of \cref{app:exp_metrics}: the hyperparameter fitting, the EIG computation from scratch at refit iterations, the EIG updates at no-refit iterations, the bookkeeping, and the posterior readout. It also reports the mean time of a single EIG computation at a refit iteration and of a single EIG update at a no-refit iteration. Two observations stand out. For the larger games at large budgets, the overhead is dominated by the hyperparameter fitting. Moreover, while a single EIG update is more than an order of magnitude faster than a computation from scratch, the updates dominate the computations from scratch cumulatively, since they are performed in every iteration, whereas refits occur only $\bigO(\log \budget)$ times.
\begin{table}[h]
  \caption{Mean overhead of \methodname{} over $30$ seeds for three games of increasing size, split into its components (\cref{app:exp_metrics}): cumulative times up to the budget $\budget$ in minutes / seconds (share of the total in parentheses), and the mean time in seconds of one EIG computation at a refit iteration and of one EIG update at a no-refit iteration.}
  \label{tab:overhead_breakdown}
  \centering \scriptsize \setlength{\tabcolsep}{2.5pt}
  \resizebox{\textwidth}{!}{\begin{tabular}{@{}llrrrrr@{}}
\toprule
Game ($p$) & Component $\backslash$ $T$ & 512 & 1024 & 2048 & 4096 & 8192 \\
\midrule
DINOv2-16 (16) & hyperparameter fitting (cum.) & 0.0 / 2 (21\%) & 0.1 / 4 (20\%) & 0.1 / 8 (20\%) & 0.4 / 27 (25\%) & 2.1 / 124 (34\%) \\
 & EIG, refit steps (cum.) & 0.0 / 0 (4\%) & 0.0 / 1 (3\%) & 0.0 / 1 (3\%) & 0.0 / 2 (2\%) & 0.1 / 4 (1\%) \\
 & EIG, no-refit steps (cum.) & 0.1 / 3 (27\%) & 0.1 / 7 (36\%) & 0.3 / 18 (44\%) & 0.9 / 54 (50\%) & 3.0 / 181 (50\%) \\
 & bookkeeping (cum.) & 0.1 / 6 (49\%) & 0.1 / 8 (42\%) & 0.2 / 14 (34\%) & 0.4 / 25 (23\%) & 0.9 / 51 (14\%) \\
 & posterior readout at $T$ & 0.0 / 0 (0\%) & 0.0 / 0 (0\%) & 0.0 / 0 (0\%) & 0.0 / 0 (0\%) & 0.0 / 3 (1\%) \\
 & total (cum.) & \textbf{0.2 / 11} & \textbf{0.3 / 20} & \textbf{0.7 / 41} & \textbf{1.8 / 108} & \textbf{6.0 / 363} \\
 & EIG per refit step (s) & 0.10 & 0.13 & 0.18 & 0.29 & 0.54 \\
 & EIG per no-refit step (s) & 0.01 & 0.01 & 0.02 & 0.03 & 0.04 \\
\midrule
Bankruptcy (64) & hyperparameter fitting (cum.) & 0.0 / 2 (18\%) & 0.1 / 7 (23\%) & 0.4 / 22 (32\%) & 1.8 / 109 (52\%) & 9.3 / 558 (66\%) \\
 & EIG, refit steps (cum.) & 0.0 / 1 (8\%) & 0.0 / 2 (6\%) & 0.0 / 3 (4\%) & 0.1 / 4 (2\%) & 0.1 / 7 (1\%) \\
 & EIG, no-refit steps (cum.) & 0.1 / 5 (42\%) & 0.2 / 13 (45\%) & 0.5 / 31 (44\%) & 1.2 / 72 (34\%) & 3.8 / 227 (27\%) \\
 & bookkeeping (cum.) & 0.1 / 4 (32\%) & 0.1 / 7 (26\%) & 0.2 / 14 (20\%) & 0.4 / 26 (12\%) & 0.9 / 54 (6\%) \\
 & posterior readout at $T$ & 0.0 / 0 (0\%) & 0.0 / 0 (0\%) & 0.0 / 0 (0\%) & 0.0 / 0 (0\%) & 0.0 / 2 (0\%) \\
 & total (cum.) & \textbf{0.2 / 13} & \textbf{0.5 / 29} & \textbf{1.2 / 70} & \textbf{3.5 / 212} & \textbf{14.1 / 849} \\
 & EIG per refit step (s) & 0.54 & 0.58 & 0.67 & 0.82 & 1.17 \\
 & EIG per no-refit step (s) & 0.03 & 0.03 & 0.03 & 0.04 & 0.06 \\
\midrule
MIC (111) & hyperparameter fitting (cum.) & 0.0 / 3 (18\%) & 0.1 / 6 (18\%) & 0.4 / 26 (31\%) & 2.0 / 120 (48\%) & 9.3 / 557 (62\%) \\
 & EIG, refit steps (cum.) & 0.0 / 3 (18\%) & 0.1 / 4 (12\%) & 0.1 / 6 (7\%) & 0.1 / 8 (3\%) & 0.2 / 12 (1\%) \\
 & EIG, no-refit steps (cum.) & 0.1 / 6 (37\%) & 0.3 / 16 (47\%) & 0.6 / 38 (46\%) & 1.6 / 94 (38\%) & 4.5 / 271 (30\%) \\
 & bookkeeping (cum.) & 0.1 / 4 (26\%) & 0.1 / 8 (22\%) & 0.2 / 14 (17\%) & 0.5 / 28 (11\%) & 1.1 / 63 (7\%) \\
 & posterior readout at $T$ & 0.0 / 0 (0\%) & 0.0 / 0 (0\%) & 0.0 / 0 (0\%) & 0.0 / 0 (0\%) & 0.0 / 2 (0\%) \\
 & total (cum.) & \textbf{0.3 / 15} & \textbf{0.6 / 34} & \textbf{1.4 / 84} & \textbf{4.2 / 250} & \textbf{15.1 / 905} \\
 & EIG per refit step (s) & 1.38 & 1.41 & 1.48 & 1.66 & 2.04 \\
 & EIG per no-refit step (s) & 0.04 & 0.04 & 0.04 & 0.05 & 0.07 \\
\bottomrule
\end{tabular}
}
\end{table}
\FloatBarrier
\subsection{Ablation Path from \shapleig{} to \methodname{}} \label{app:res_ablation}
\begin{figure}[p]
  \centering
  \includegraphics[width=\textwidth]{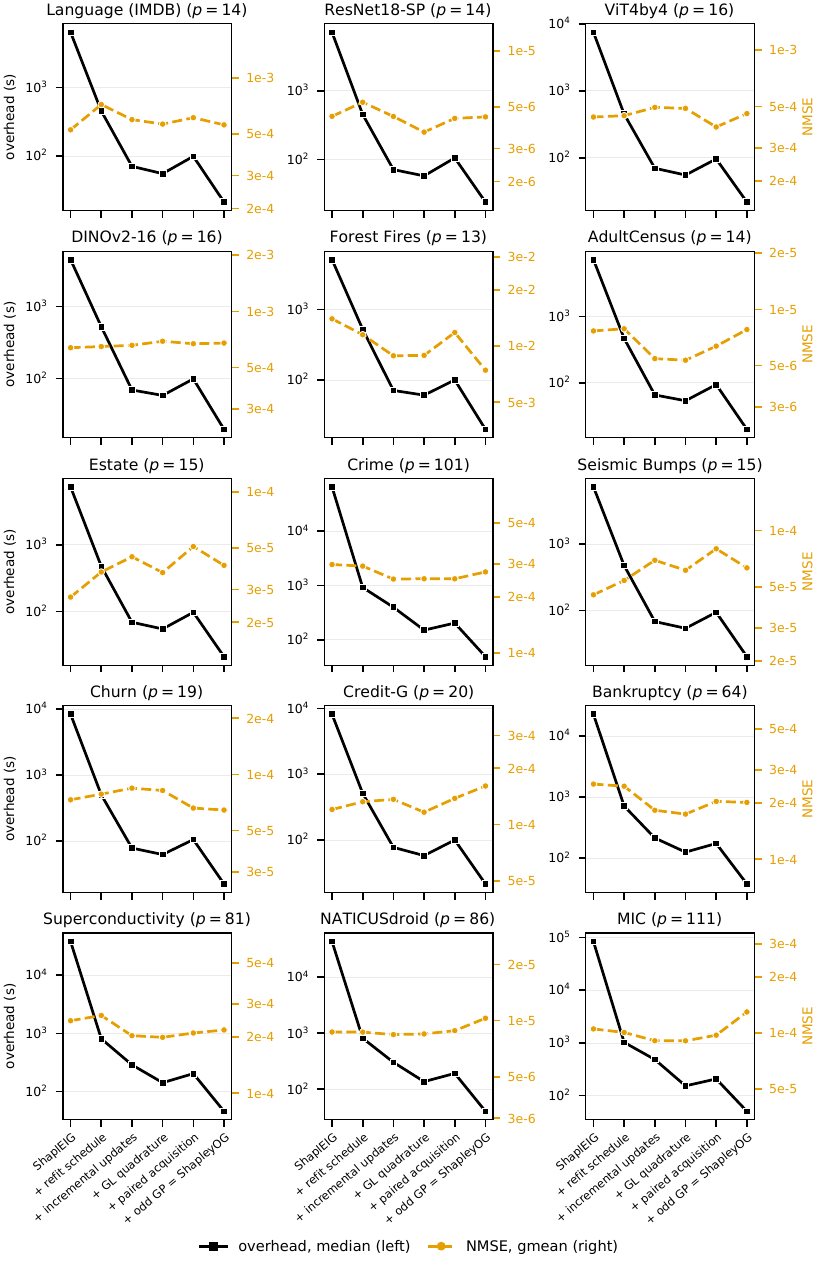}
  \caption{Ablation path from \shapleig{} to \methodname{} at $\budget = 2048$ evaluations for all $15$ games: overhead (median over $10$ seeds, left axis) and NMSE (geometric mean, right axis) when adding one measure at a time.}
  \label{fig:ladder_all}
\end{figure}

This subsection gives the details of the ablation path for all games. \Cref{fig:ladder_all} repeats the ablation path of \cref{fig:pareto} for all $15$ games, \cref{tab:ablation_speedup} quantifies, for every game at $\budget = 2048$ evaluations, the speed-up of each variant over \shapleig{} and its NMSE relative to \shapleig{}, and \cref{tab:ablation_absolute} lists the absolute overheads and NMSEs. In the variant with paired acquisition under the full GP, which adds two observations per iteration, our current implementation does not apply the rank-one updates of the posterior covariances (\cref{app:alg_updates}) but recomputes them in every iteration without refit, so that this variant is even slower than the preceding one and the contribution of paired acquisition alone is understated. \methodname{} is not affected, since under the odd GP every pair adds a single observation. Note that the relative gain of each measure depends on the budget and the number of players; e.g., the GL quadrature pays off more for games with many players (\cref{app:res_quadrature}).
\begin{table}[h]
  \caption{Speed-up of each variant over \shapleig{} (median over $10$ seeds of the paired ratio of overheads) and its NMSE relative to \shapleig{} (geometric mean over seeds of the paired ratio; values above $1$ are worse), per game at $\budget = 2048$ evaluations; the last row is the geometric mean over the games.}
  \label{tab:ablation_speedup}
  \centering \scriptsize \setlength{\tabcolsep}{4pt}
  \begin{tabular}{@{}lrrrrrrrrrr@{}}
\toprule
Game ($p$) & \multicolumn{2}{c}{\shortstack{+ refit\\schedule}} & \multicolumn{2}{c}{\shortstack{+ incremental\\updates}} & \multicolumn{2}{c}{\shortstack{+ GL\\quadrature}} & \multicolumn{2}{c}{\shortstack{+ paired\\acquisition}} & \multicolumn{2}{c}{\shortstack{+ odd GP\\= ShapleyOG}} \\
\cmidrule(lr){2-3} \cmidrule(lr){4-5} \cmidrule(lr){6-7} \cmidrule(lr){8-9} \cmidrule(lr){10-11}
 & speed-up & NMSE & speed-up & NMSE & speed-up & NMSE & speed-up & NMSE & speed-up & NMSE \\
\midrule
Language (IMDB) (14) & 14$\times$ & 1.37 & 95$\times$ & 1.13 & 119$\times$ & 1.07 & 71$\times$ & 1.16 & 306$\times$ & 1.06 \\
ResNet18-SP (14) & 16$\times$ & 1.19 & 97$\times$ & 1.00 & 123$\times$ & 0.82 & 67$\times$ & 0.98 & 290$\times$ & 0.99 \\
ViT4by4 (16) & 17$\times$ & 1.02 & 105$\times$ & 1.13 & 130$\times$ & 1.11 & 79$\times$ & 0.88 & 346$\times$ & 1.04 \\
DINOv2-16 (16) & 8$\times$ & 1.01 & 66$\times$ & 1.03 & 80$\times$ & 1.08 & 46$\times$ & 1.05 & 233$\times$ & 1.06 \\
Forest Fires (13) & 10$\times$ & 0.82 & 73$\times$ & 0.63 & 86$\times$ & 0.64 & 51$\times$ & 0.84 & 256$\times$ & 0.53 \\
AdultCensus (14) & 15$\times$ & 1.03 & 110$\times$ & 0.71 & 134$\times$ & 0.70 & 77$\times$ & 0.83 & 365$\times$ & 1.02 \\
Estate (15) & 16$\times$ & 1.36 & 108$\times$ & 1.64 & 135$\times$ & 1.35 & 77$\times$ & 1.87 & 358$\times$ & 1.48 \\
Crime (101) & 71$\times$ & 0.98 & 164$\times$ & 0.84 & 435$\times$ & 0.84 & 323$\times$ & 0.84 & 1329$\times$ & 0.91 \\
Seismic Bumps (15) & 15$\times$ & 1.19 & 105$\times$ & 1.53 & 129$\times$ & 1.35 & 76$\times$ & 1.76 & 357$\times$ & 1.39 \\
Churn (19) & 17$\times$ & 1.07 & 106$\times$ & 1.15 & 135$\times$ & 1.12 & 80$\times$ & 0.90 & 377$\times$ & 0.88 \\
Credit-G (20) & 16$\times$ & 1.10 & 103$\times$ & 1.13 & 138$\times$ & 0.97 & 80$\times$ & 1.15 & 379$\times$ & 1.34 \\
Bankruptcy (64) & 32$\times$ & 0.97 & 106$\times$ & 0.72 & 183$\times$ & 0.69 & 132$\times$ & 0.81 & 597$\times$ & 0.80 \\
Superconductivity (81) & 48$\times$ & 1.07 & 132$\times$ & 0.83 & 269$\times$ & 0.81 & 189$\times$ & 0.86 & 832$\times$ & 0.89 \\
NATICUSdroid (86) & 53$\times$ & 1.00 & 143$\times$ & 0.97 & 316$\times$ & 0.98 & 222$\times$ & 1.02 & 1058$\times$ & 1.19 \\
MIC (111) & 81$\times$ & 0.96 & 173$\times$ & 0.86 & 545$\times$ & 0.86 & 400$\times$ & 0.93 & 1634$\times$ & 1.23 \\
\midrule
\emph{geometric mean} & 22$\times$ & 1.07 & 109$\times$ & 0.99 & 167$\times$ & 0.94 & 104$\times$ & 1.02 & 475$\times$ & 1.03 \\
\bottomrule
\end{tabular}

\end{table}
\begin{table}[h]
  \caption{Absolute counterpart of \cref{tab:ablation_speedup}: overhead (median over $10$ seeds) and NMSE (geometric mean over seeds) of \shapleig{} and of each variant, per game at $\budget = 2048$ evaluations; the last row is the geometric mean over the games.}
  \label{tab:ablation_absolute}
  \centering \scriptsize \setlength{\tabcolsep}{4pt}
  \setlength{\tabcolsep}{2.5pt}%
\begin{tabular}{@{}lrrrrrrrrrrrr@{}}
\toprule
Game ($p$) & \multicolumn{2}{c}{ShaplEIG} & \multicolumn{2}{c}{\shortstack{+ refit\\schedule}} & \multicolumn{2}{c}{\shortstack{+ incremental\\updates}} & \multicolumn{2}{c}{\shortstack{+ GL\\quadrature}} & \multicolumn{2}{c}{\shortstack{+ paired\\acquisition}} & \multicolumn{2}{c}{\shortstack{+ odd GP\\= ShapleyOG}} \\
\cmidrule(lr){2-3} \cmidrule(lr){4-5} \cmidrule(lr){6-7} \cmidrule(lr){8-9} \cmidrule(lr){10-11} \cmidrule(lr){12-13}
 & overhead & NMSE & overhead & NMSE & overhead & NMSE & overhead & NMSE & overhead & NMSE & overhead & NMSE \\
\midrule
Language (IMDB) (14) & 1.8\,h & 5.3e-4 & 7.6\,min & 7.2e-4 & 1.2\,min & 6.0e-4 & 55\,s & 5.7e-4 & 1.6\,min & 6.1e-4 & 21\,s & 5.6e-4 \\
ResNet18-SP (14) & 2.0\,h & 4.5e-6 & 7.5\,min & 5.3e-6 & 1.2\,min & 4.5e-6 & 58\,s & 3.7e-6 & 1.8\,min & 4.4e-6 & 24\,s & 4.4e-6 \\
ViT4by4 (16) & 2.1\,h & 4.4e-4 & 7.5\,min & 4.5e-4 & 1.2\,min & 4.9e-4 & 55\,s & 4.9e-4 & 1.6\,min & 3.9e-4 & 22\,s & 4.6e-4 \\
DINOv2-16 (16) & 1.2\,h & 6.4e-4 & 8.7\,min & 6.5e-4 & 1.2\,min & 6.6e-4 & 58\,s & 6.9e-4 & 1.6\,min & 6.7e-4 & 20\,s & 6.8e-4 \\
Forest Fires (13) & 1.4\,h & 1.4e-2 & 8.7\,min & 1.2e-2 & 1.2\,min & 8.9e-3 & 1.0\,min & 8.9e-3 & 1.7\,min & 1.2e-2 & 20\,s & 7.4e-3 \\
AdultCensus (14) & 2.0\,h & 7.7e-6 & 7.8\,min & 7.9e-6 & 1.1\,min & 5.4e-6 & 53\,s & 5.3e-6 & 1.6\,min & 6.4e-6 & 20\,s & 7.8e-6 \\
Estate (15) & 2.1\,h & 2.7e-5 & 7.9\,min & 3.7e-5 & 1.1\,min & 4.5e-5 & 54\,s & 3.7e-5 & 1.6\,min & 5.1e-5 & 21\,s & 4.0e-5 \\
Crime (101) & 18.1\,h & 3.0e-4 & 15.3\,min & 2.9e-4 & 6.6\,min & 2.5e-4 & 2.5\,min & 2.5e-4 & 3.4\,min & 2.5e-4 & 49\,s & 2.7e-4 \\
Seismic Bumps (15) & 2.0\,h & 4.5e-5 & 7.9\,min & 5.4e-5 & 1.1\,min & 6.9e-5 & 54\,s & 6.1e-5 & 1.6\,min & 8.0e-5 & 20\,s & 6.3e-5 \\
Churn (19) & 2.3\,h & 7.3e-5 & 8.4\,min & 7.8e-5 & 1.3\,min & 8.4e-5 & 1.0\,min & 8.2e-5 & 1.8\,min & 6.6e-5 & 22\,s & 6.4e-5 \\
Credit-G (20) & 2.3\,h & 1.2e-4 & 8.5\,min & 1.3e-4 & 1.3\,min & 1.4e-4 & 58\,s & 1.2e-4 & 1.7\,min & 1.4e-4 & 21\,s & 1.6e-4 \\
Bankruptcy (64) & 6.3\,h & 2.5e-4 & 11.9\,min & 2.5e-4 & 3.6\,min & 1.8e-4 & 2.1\,min & 1.7e-4 & 2.9\,min & 2.0e-4 & 38\,s & 2.0e-4 \\
Superconductivity (81) & 10.6\,h & 2.4e-4 & 13.5\,min & 2.6e-4 & 4.8\,min & 2.0e-4 & 2.4\,min & 2.0e-4 & 3.4\,min & 2.1e-4 & 46\,s & 2.2e-4 \\
NATICUSdroid (86) & 12.0\,h & 8.7e-6 & 13.6\,min & 8.7e-6 & 5.1\,min & 8.4e-6 & 2.3\,min & 8.5e-6 & 3.2\,min & 8.8e-6 & 41\,s & 1.0e-5 \\
MIC (111) & 23.1\,h & 1.1e-4 & 17.0\,min & 1.0e-4 & 8.1\,min & 9.1e-5 & 2.6\,min & 9.1e-5 & 3.5\,min & 9.7e-5 & 51\,s & 1.3e-4 \\
\midrule
\emph{geometric mean} & 3.6\,h & 1.2e-4 & 9.7\,min & 1.3e-4 & 2.0\,min & 1.2e-4 & 1.3\,min & 1.1e-4 & 2.1\,min & 1.2e-4 & 27\,s & 1.2e-4 \\
\bottomrule
\end{tabular}

\end{table}
\FloatBarrier
\subsection{Quadrature Study} \label{app:res_quadrature}
\Cref{tab:quadrature} presents the detailed results of the quadrature study of \cref{sec:quad} (setup in \cref{app:exp_ablations}). For every number of nodes $\nq$, it lists the median relative Frobenius error of the SV prior covariances $\A\K(\Zall,\cands)$ and $\A\K(\Zall,\Zall)\A\trans$ and of the EIG scores against the exact quadrature at the checkpoints, the share of checkpoints at which the exact scores select the same coalition, the NMSE at $4096$ and $8192$ evaluations (geometric mean over seeds), and the median time per run of the covariance computations and of the whole overhead. The table shows that down to $8$ nodes, the SV prior covariances agree with the exact quadrature up to machine precision and every checkpoint selects the same coalition, while the time of the covariance computations drops from $12$ minutes to $14$ seconds per run and the overhead from $54$ to $43$ minutes. The ESP route of \shapleig{}, run for the first seed, timed out after $24$ hours during its first computation of $\A\K(\Zall,\Zall)\A\trans$, having processed $352$ of the $1024$ players, i.e., it did not even complete the SV prior covariances once.
\begin{table}[htbp]
  \caption{Quadrature study on QSAR TID11 ($\np = 1024$, $10$ seeds) for different numbers of nodes $\nq$ and for the ESP route of \shapleig{} (one seed, budget $4096$).}
  \label{tab:quadrature}
  \centering \scriptsize \setlength{\tabcolsep}{4pt}
  \begin{tabular}{@{}rrrrrrrrr@{}}
\toprule
& \multicolumn{3}{c}{relative error vs.\ exact} & & \multicolumn{2}{c}{NMSE at $\budget$} & \multicolumn{2}{c}{time per run} \\
\cmidrule(lr){2-4} \cmidrule(lr){6-7} \cmidrule(l){8-9}
$\nq$ & $\A\K(\Zall,\cands)$ & $\A\K(\Zall,\Zall)\A^\top$ & EIG & same argmax & $4096$ & $8192$ & covariances & overhead \\
\midrule
$514$ (exact) & -- & -- & -- & -- & 6.2e-4 & 2.9e-4 & 11.7\,min & 53.5\,min \\
$256$ & 4.7e-15 & 7.0e-15 & 9.1e-9 & 100\% & 6.2e-4 & 2.9e-4 & 4.8\,min & 47.1\,min \\
$64$ & 8.4e-15 & 1.2e-14 & 1.6e-8 & 100\% & 6.2e-4 & 2.9e-4 & 1.1\,min & 43.9\,min \\
$8$ & 1.2e-14 & 6.7e-14 & 6.8e-8 & 100\% & 6.2e-4 & 2.9e-4 & 14\,s & 43.0\,min \\
$2$ & 3.6e-4 & 4.1e-4 & 2.5e-2 & 14\% & 7.5e-4 & 3.2e-4 & 8\,s & 42.1\,min \\
\midrule
ESP (timeout) & -- & -- & -- & -- & -- & -- & $> 24$\,h & $> 24$\,h \\
\bottomrule
\end{tabular}

\end{table}
\FloatBarrier
\end{document}